\documentclass[review,3p,12pt,times,sort,compress]{elsarticle}

\usepackage{multirow}
\usepackage{amsmath}
\usepackage{amssymb}
\usepackage{enumitem}
\usepackage{wrapfig}
\usepackage{subfig}
\usepackage{url}

\usepackage{multicol}
\usepackage{array}
\usepackage{amsmath,amsfonts}
\usepackage{textcomp}
\usepackage{stfloats}
\usepackage{url}
\usepackage{verbatim}
\usepackage{graphicx}
\usepackage{booktabs}
\usepackage{tabularx}
\usepackage{longtable}
\usepackage{tabu}
\usepackage{float}
\usepackage{tabularray}
\usepackage{makecell}
\usepackage{placeins}
\usepackage{hyperref}
\usepackage{color}
\usepackage{pdfpages}
\newcommand{\hc}[1]{{\color{black} #1}}
\newcommand{\ml}[1]{{\color{black} #1}}
\newcommand{\xj}[1]{{\color{black} #1}}

\journal{Neurocomputing}

\begin{document}

\begin{frontmatter}



\title{A Survey of Artificial Intelligence in Gait-Based Neurodegenerative Disease Diagnosis}

\renewcommand{\thefootnote}{\fnsymbol{footnote}}
\footnotetext[2]{We provide a public resource repository to track and facilitate developments in this emerging field: \href{https://github.com/minlinzeng/AI4NDD-Survey}{https://github.com/minlinzeng/AI4NDD-Survey}.}

\author[label1,label2]{Haocong Rao\fnref{equal}}
\ead{haocong001@ntu.edu.sg}
\author[label1,label2]{Minlin Zeng\fnref{equal}}
\ead{minlin001@ntu.edu.sg}
\author[label1,label2]{Xuejiao Zhao}
\ead{xjzhao@ntu.edu.sg}
\author[label1,label2]{Chunyan Miao\corref{cor1}}
\ead{ascymiao@ntu.edu.sg}
\affiliation[label1]{organization={Joint NTU-UBC Research Centre of Excellence in Active Living for the Elderly (LILY), Nanyang Technological University},
country={Singapore}}
\affiliation[label2]{organization={College of Computing and Data Science, Nanyang Technological University},
addressline={50 Nanyang Avenue}, 
postcode={639798}, 
country={Singapore}}
\cortext[cor1]{Corresponding author}
\fntext[equal]{The two authors contribute equally to this work.}

\begin{abstract}
Recent years have witnessed an increasing global population affected by neurodegenerative diseases (NDs), which traditionally require extensive healthcare resources and human effort for medical diagnosis and monitoring. As a crucial disease-related motor symptom, human gait can be exploited to characterize different NDs. The current advances in artificial intelligence (AI) models enable automatic gait analysis for NDs identification and classification, opening a new avenue to facilitate faster and more cost-effective diagnosis of NDs.
    In this paper, we provide a comprehensive survey on recent progress of machine learning and deep learning based AI techniques applied to diagnosis of five typical NDs through gait. We provide an overview of the process of AI-assisted NDs diagnosis, and present a systematic taxonomy of existing gait data and AI models. 
  Meanwhile, a novel quality evaluation criterion is proposed to quantitatively assess the quality of existing studies.
    \hc{Through an extensive review and analysis of 169 studies, we present recent technical advancements, discuss existing challenges, potential solutions, and future directions in this field.} Finally, we envision the prospective utilization of 3D skeleton data for human gait representation and the development of more efficient AI models for NDs diagnosis\footnotemark[2].
\end{abstract}




\end{frontmatter}


\section{Introduction}

\begin{figure}[t]
    \centering
     \begin{minipage}{0.48\textwidth}
     \centering
     \scalebox{0.35}{
    \includegraphics[]{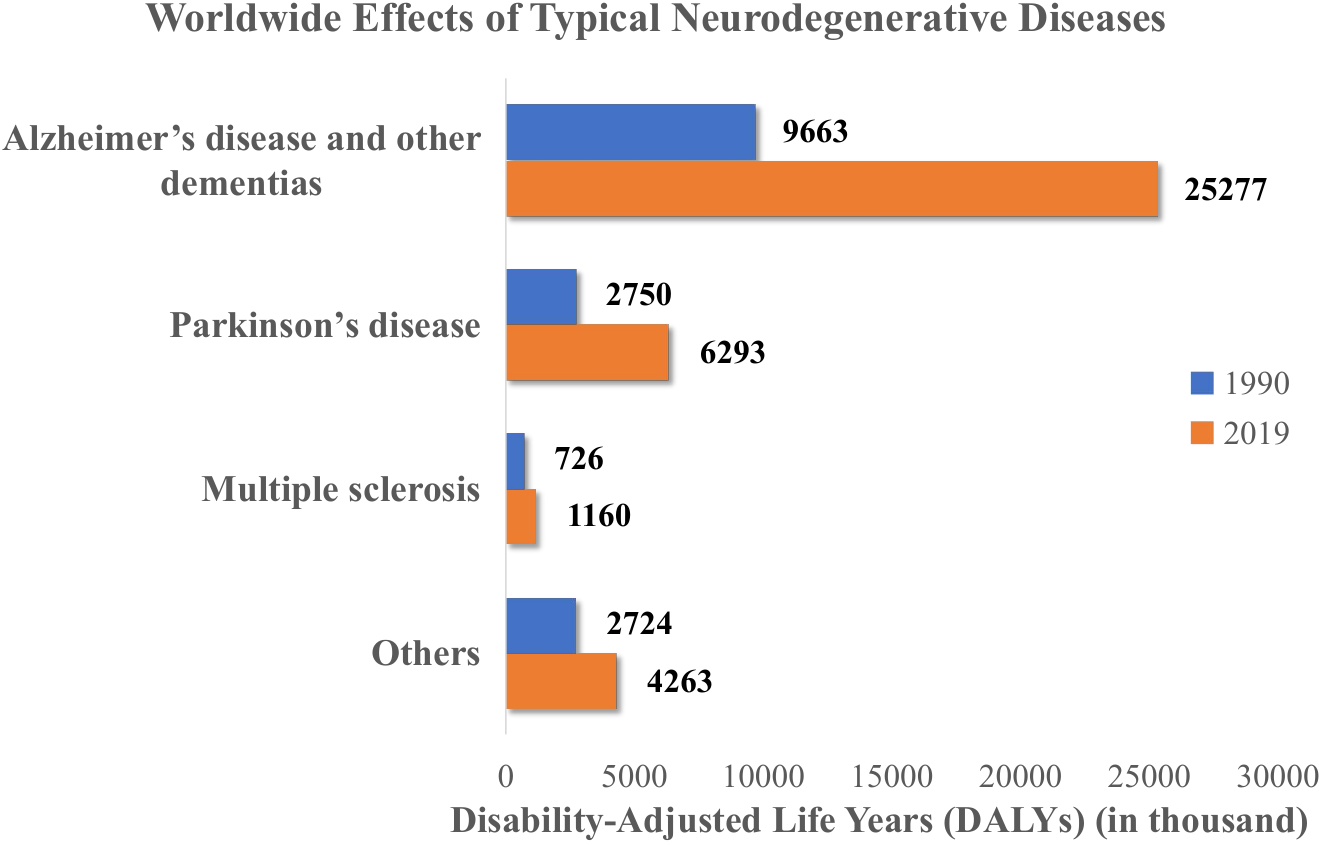}
    }
    \caption{Worldwide effects (1990 versus 2019) of two most representative neurodegenerative diseases (AD and PD), measured in “Disability-Adjusted Life Years (DALYs)” \cite{ding2022global}.}
    \label{overview_pop}
    \end{minipage}
    \quad \quad
    \begin{minipage}{0.45\textwidth}
     \centering
     \scalebox{0.7}{
     \includegraphics[]{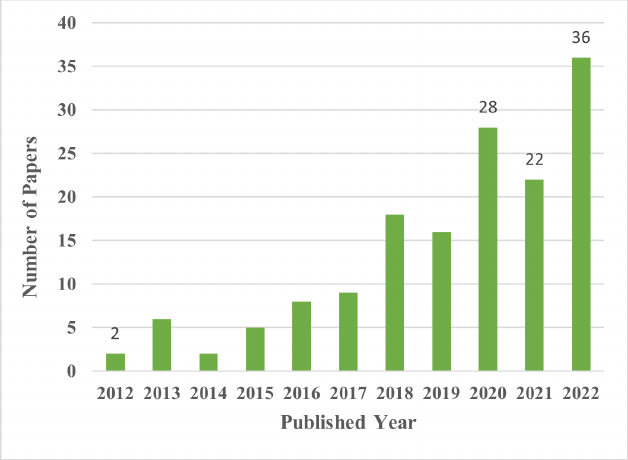}
     }
     \caption{The number of studies (from 2012 to 2022) relevant to gait-based neurodegenerative disease diagnosis using AI.}
     \label{paper_year}
     \end{minipage}
\end{figure}

Neurodegenerative Diseases (NDs) such as Alzheimer's disease (AD) and Parkinson’s disease (PD) are among the most widespread and devastating disorders affecting millions of people worldwide \cite{scheltens2021alzheimer}. According to the Alzheimer's Disease Association \cite{AD_statistic} and the Parkinson’s Foundation Report \cite{PD_statistic} in 2023, nearly 6.2 million people and 1 million people are diagnosed with AD and PD in the United States, respectively. While in China, there are over 9.8 million AD patients \cite{ren2022china}, and it is estimated that Chinese PD patients will increase to 4.9 million by 2030, accounting for a half of the worldwide PD population \cite{dorsey2007projected}. From 1990 to 2019, the effected global population of AD, PD, and other neurodegenerative diseases ($e.g.$, amyotrophic lateral sclerosis) has significantly increased (illustrated in Fig. \ref{overview_pop}), leading to a growing burden on global healthcare system. On the one hand, these diseases could result in pathological gaits and chronic pain in body joints, tissues, and nerves \cite{de2016pain,negre2008chronic}, which severely reduce the flexibility, stability, mobility, and other functional capabilities of a human body. On the other hand, they also impose large psychological and financial burdens to patients and their family \cite{beiske2009pain,ovaska2021literature,averill2007psychological}.
 The treatment cost for these diseases has reached approximately USD 130 billion per year and further rise has been estimated \cite{adejare2016drug,zahra2020global}. 
 Therefore, it is essential to conduct an earlier and more reliable diagnosis of these diseases, so as to provide a timely proper treatment to mitigate the burden of societal and healthcare resources.

\begin{figure}[t]
    \centering
     \begin{minipage}{0.45\textwidth}
     \centering
     \scalebox{0.35}{
    \includegraphics[]{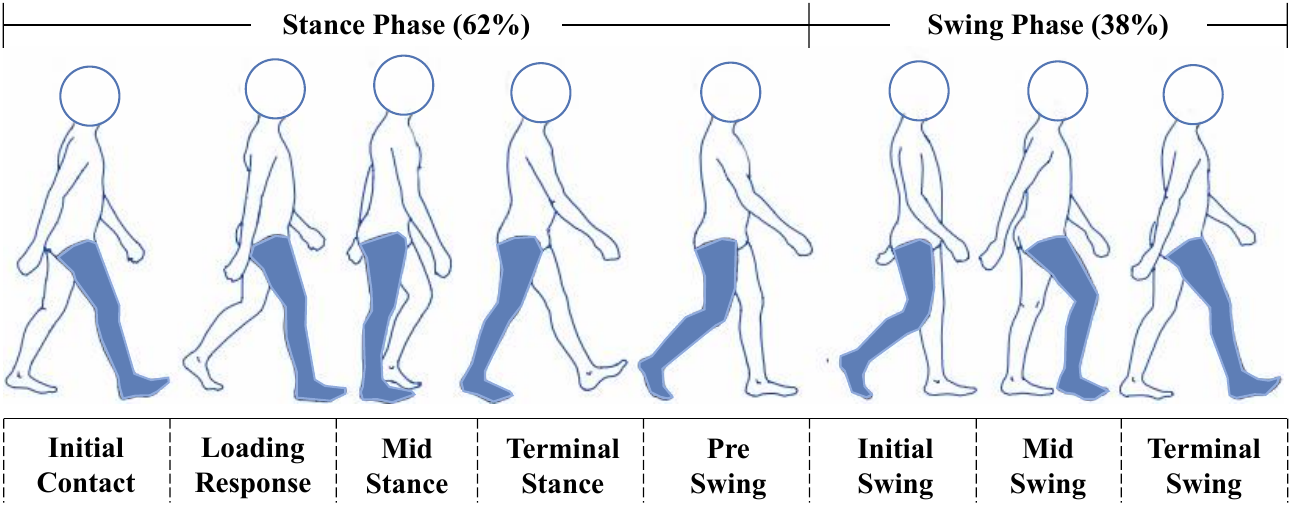}
    }
    \caption{Different phases of normal human gait with coordinated movements of legs and arms.}
    \label{gait_cycle}
    \end{minipage}
    \quad \quad
    \begin{minipage}{0.45\textwidth}
    \centering
    \scalebox{0.132}{
    \includegraphics[]{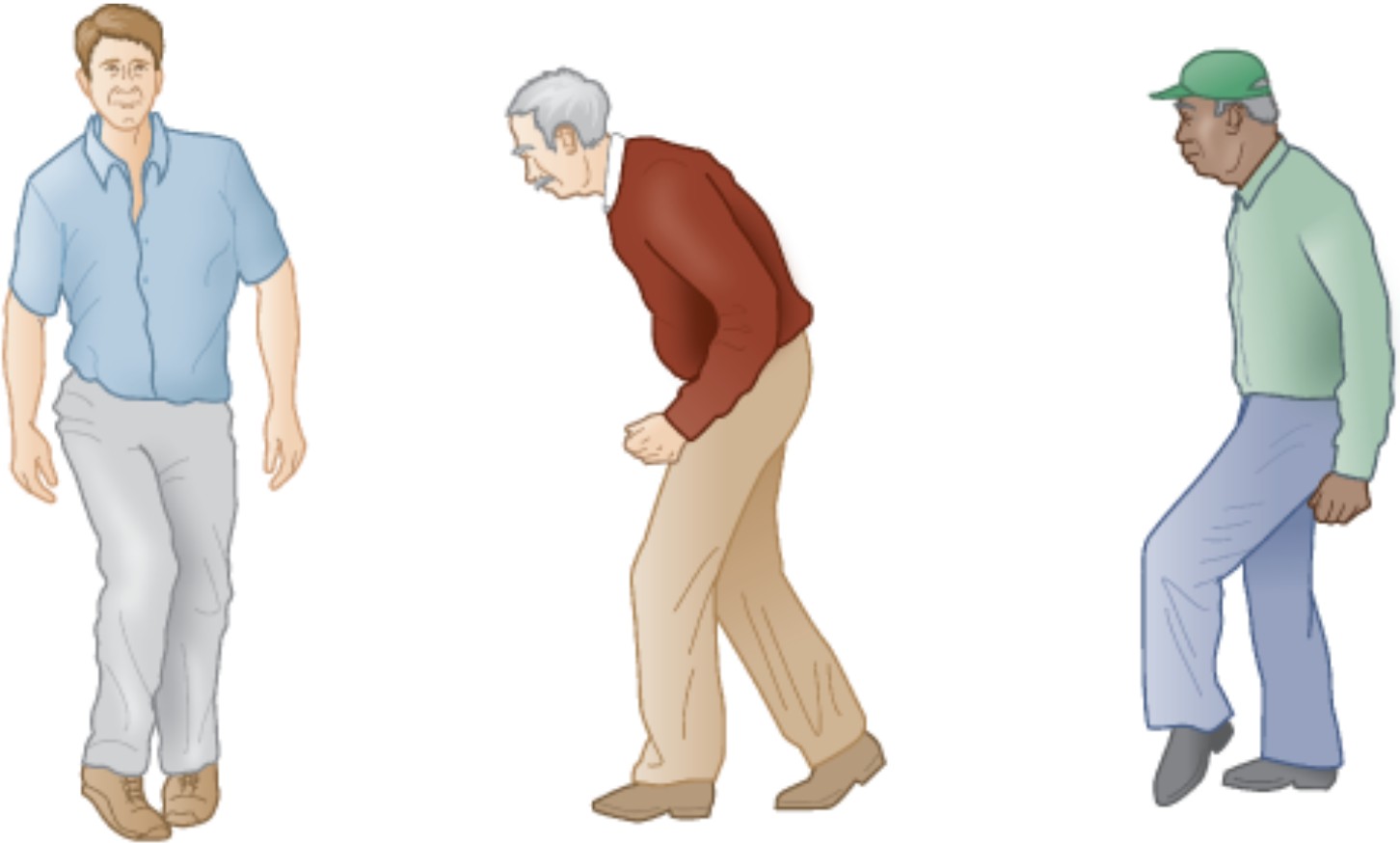}
    }
     \caption{Examples for hemiplegic gait (left), Parkinsonian gait (middle), neuropathic gait (right) \cite{https://doi.org/10.1002/dac.4348}.}
     \label{gait_examples}
    \end{minipage}
\end{figure}

As one of the most essential motor symptoms associated with pathology, locomotion anomalies, especially observed in \textit{human gait}, can reflect the incidence and progression of different NDs \cite{kirchner2014detrended,sejdic2013comprehensive,tang2013locomotion}. Taking AD, PD and amyotrophic lateral sclerosis (ALS) as an example, these diseases often cause patients to exhibit marked alterations and abnormalities in their walking patterns \cite{kobsar2017wearable,rutherford2018knee,spasojevic2015vision,armand2016gait,sanders2010gait,allet2008gait,myers2016gait,myers2013vascular,wurdeman2012patients,szymczak2018gait}.
In general, compared with a normal gait (see Fig. \ref{gait_cycle}), an impaired gait may display changes in speed, cadence, and limb moments within a gait cycle  \cite{favre2014age,creaby2012gait,spasojevic2015vision}.
As shown in Fig. \ref{gait_examples}, the gait with a widened base, unsteadiness and irregularity of steps, and lateral veering (termed \textit{Parkinsonian gait}) could suggest PD \cite{spasojevic2015vision,ebersbach2013clinical}. In the case of advanced ALS, the patients may possess foot drop, where
one foot flops down when lifting the leg, necessitating a
higher lifting of the knee to prevent the toe from dragging on the ground (termed \textit{neuropathic gait}) \cite{hausdorff2000dynamic,isaacs2007amyotrophic,wijesekera2009amyotrophic}. 

In recent years, driven by the widespread availability of body positioning technologies and economical portable devices to monitor human gait, gait-based NDs diagnosis ($i.e.,$ NDs classification, identification or prediction) has propelled a surge of attention across academic, industrial, and medical communities \cite{aich2020design,khera2022age,fraiwan2021computer,kour2023sensor,di2020gait,pardoel2019wearable,gupta2023new,vienne2017inertial,das2022recent,salchow2022emerging,loh2021application,ayaz2023automated,zolfaghari2023sensor}.
To perform NDs diagnosis from diverse gait data, the Artificial Intelligence (AI) technologies, especially Machine Learning (ML) and Deep Learning (DL) models, have been extensively explored and applied \cite{kour2022vision,khera2022age,fraiwan2021computer,kour2019computer,kour2023sensor,di2020gait,seifallahi2024detection,li2023using,chiarello2022effect,varma2021continuous,ramirez2021dual}. 
A common practice is to extract gait parameters ($e.g.,$ stride length, stance duration, gait cycle time) from \textit{sensor-based} data such as vertical ground reaction force (vGRF) or \textit{vision-based} data such as gait images, and leverage ML algorithms ($e.g.,$ support vector machine (SVM)) or deep neural networks to automatically recognize gait patterns associated with NDs to aid in their diagnosis \cite{mannini2016machine,wahid2015classification,abdulhay2018gait}. 
Compared with traditional diagnosis that requires manual observations of clinical motor symptoms ($e.g.,$ rigidity),
these AI-based methods can analyze the abnormality in gait to automatically identify NDs related patterns, which shows immense potential in assisting healthcare professionals to diagnose and treat these diseases more efficiently \cite{kour2023sensor,di2020gait}. 

In this paper, we systematically review recent advancements of utilizing AI to help diagnose NDs from human gait. In particular, our work focuses on the diagnosis of \textit{five} most prevalent and typical NDs (PD, AD, ALS, Huntington's Disease (HD), Multiple Sclerosis (MS)) and \textit{all} existing AI models that can help diagnose these diseases ($i.e.,$ identify, classify or predict NDs) from human gait data. A total of \hc{2074} potentially relevant articles, including 14 survey papers, are found from three databases (\textit{PubMed}, \textit{Web of Science (WoS)}, \textit{Google Scholar}). We first screen these articles by titles and abstracts based on the pre-defined inclusion criteria, and then conduct a full-text assessment to exclude irrelevant papers. Finally, \textit{\hc{169}} of \hc{2074} papers are selected to be included in our survey.

\subsection{Objectives and Contributions}
The objectives of this survey can be encapsulated in the following six key questions (Q): 
\hc{
\begin{itemize}
    \item Q1. How is human gait associated with NDs?\item Q2. What is the process of AI-assisted NDs diagnosis from gait data? 
    \item Q3. What are various gait data types, gait collecting approaches, and AI model types for NDs diagnosis? 
    \item Q4. What are different AI models used in the gait-based diagnosis of different NDs, and how are their performance ($e.g.,$ diagnostic accuracy) and quality?
   \item Q5. What are the most representative datasets and state-of-the-art AI models in gait-based NDs diagnosis? 
   \item Q6. What are the current main challenges and promising future directions of this field? 
   \item Q7. Is it feasible to exploit 3D skeleton data to represent human gait and build AI models for NDs diagnosis? 
\end{itemize}
}
The contributions of this work are as follows. We provide a systematic literature review of existing AI technologies for gait-based diagnosis of five most widespread NDs (PD, AD, ALS, HD, MS). We formulate the general process of AI-assisted gait-based NDs diagnosis, and present a systematic taxonomy of used gait data and AI models. Our survey demonstrates how human gait features can be a crucial indicator for diagnosing various NDs, and highlights the significant role of current AI models in facilitating the automation of this diagnostic process. Moreover, we analyze the statistics of existing studies to reveal the current development and future trend of this area, and comprehensively discuss the key challenges in utilizing AI for gait-based NDs diagnosis, such as scarcity and imbalance of gait data, lack of multiple data modalities and sources, and unreliable AI model designs ($e.g.,$ limited generalizability, efficiency, and interpretability). We elaborate on these issues and delve further into their potential solutions, related technologies, and future research directions. \hc{In addition, our work also introduces a systematic quality evaluation criterion, focusing on AI novelty, comparison comprehensiveness, and sample sufficiency, to facilitate the assessment and comparison of different AI-based studies.}
To the best of our knowledge, this is the most elaborated survey on this topic, and it also provides a practical roadmap for future investigations of AI technologies for NDs diagnosis.
As our research vision, \textit{3D skeleton data}, which is an emerging generic and scalable data modality, is for the first time explored to characterize gait patterns associated with NDs. We empirically present a 3D skeleton based framework for NDs diagnosis, paving the way for more efficient gait representations and AI model learning in this field.

\subsection{Organization}
Aligning with the above objectives, we organize the rest of this survey as follows. Sec. \ref{Preliminary} provides the definitions of fundamental concepts related to gait, NDs, and gait-based NDs diagnosis (Q1). It also presents an overview for AI-assisted gait-based NDs diagnosis process (Q2), and a systematic taxonomy of gait data types, gait collection technologies, and AI model types (Q3). \hc{ Sec. \ref{Results} and \ref{sec_overview_studies} elaborate on the literature screening result with a multi-faceted statistics analysis, and provide a content summary of included studies with a systematic quality evaluation (Q4). Sec. \ref{sec_comparison} compares the differences between our survey and existing surveys on similar topics. 
    Sec. \ref{sec_representative_datasets} provides an introduction for representative public datasets and state-of-the-art methods in the gait-based diagnosis of different NDs (Q5).
 In Sec. \ref{sec_challenges_and_directions}, we discuss the main challenges, potential solutions, and several promising directions in this area (Q6), while proposing our research vision on exploiting 3D skeleton data for NDs diagnosis (Q7).} Finally, we conclude this survey in Sec. \ref{conclusion}.

\section{Preliminaries}\label{Preliminary}
In this section, we first provide the definitions of frequently-used basic concepts related to our topic (see Sec. \ref{sec_basic_concepts}) and introduce abnormal gait types that are correlated to NDs (see Sec. 
\ref{sec_common_gaits}).  
Then, we provide an overview of the AI-assisted diagnostic of NDs based on human gait, and display the workflow of AI learning in this process (see Sec. \ref{sec_overview_process}). A systematic taxonomy of gait data types, gait collecting technologies, and AI model types used in this area is also provided in Sec. \ref{sec_gait_taxonomy} and \ref{sec_AI_taxonomy}.

\subsection{Basic Concepts}
\label{sec_basic_concepts}
\begin{enumerate}
    \item  \textbf{Gait}: It is defined as a mechanism of locomotion, which contains \textit{rhythmic} and \textit{coordinated} movements of different limbs to enable the forward progression of the body \cite{sethi2022comprehensive,owusu2007ai} (as shown in Fig. \ref{gait_cycle}). 
    The gait of a person typically consists of the motion of lower extremities and their correlated posture in upper limbs or torso, which normally requires a synthesized coordination of nerves, skeletons, and muscles \cite{Nutt268}.
    Owing to this reason, the abnormality or impairment of gait can serve as an effective indicator of potential body pains or diseases caused by physical injury, aging or related disorders \cite{kour2021survey,loh2021application}.

\item \textbf{Gait Assessment}: It is a broad observational assessment of the patients' gait patterns, which is typically performed by doctors or experts. This process mainly focuses on assessing patient's walking behaviors and analyzing their abnormalities in motion coordination and rhythms. The gait assessment data including sensor, vision, and their combined data are recorded and collected by professional medical instruments in hospitals or laboratories, and some other forms of data such as gait self-reports, medical history, and psychological examination to help better analyze the reason behind abnormal gait. 

  \item \textbf{Neurodegenerative Diseases}: 
    The diseases with progressive loss of nerve cells, neuron structure, or/and their functions in the brain or peripheral nervous system are collectively termed NDs. 
    The focus of this study is on the five most common NDs, including AD, PD, ALS, HD, and MS.

\item \textbf{Diagnosing Diseases ($e.g.,$ NDs) from Gait}: It is a process of identifying, classifying or predicting a certain disease according to the assessment of gait. For traditional diagnostic methods, this process is executed based purely on physician's medical experience ($e.g.,$ using golden standards), assessment, and decision. 
For AI-based methods, we can input all gait-related data into the model to automatically learn, assess, and identify the abnormality of gait to predict or classify diseases.

\item \textbf{Disease Management}: General disease management methods consists of therapies using a combination of medications, physical therapies, psychological therapies, interventional procedures or surgeries. For NDs, the management is often disease-specific with focus on either the disease pathogenesis or symptoms experienced \cite{lamptey2022review}.
\end{enumerate}

\subsection{Pathological Gaits}
\label{sec_common_gaits}
\subsubsection{Neurodegenerative Disease Related Gaits}
\begin{itemize}

\item \textbf{Neuropathic gait} is characterized by foot drop, where one foot flops down when lifting the leg, necessitating a higher lifting of the knee to prevent the toe from dragging on the ground during walking (see Fig. \ref{gait_examples}). Neuropathic gait may be a symptom of ALS), multiple sclerosis (MS) or peripheral neuropathy.
‌
\item \textbf{Parkinsonian gait} can be identified by a forward stoop with the back and neck bent. This results in patients with Parkinson's disease tending to take smaller steps.

\item \textbf{Choreiform gait} is characterized by the irregular, jerky, involuntary movements in all extremities. This gait can be seen in certain basal ganglia disorders including HD, Sydenham's chorea, and other forms of chorea, athetosis or dystonia.

\end{itemize}

\subsubsection{Other Disease-Related Gaits}

Apart from the above NDs-related gaits, there are many other pathological gaits reflecting different body injuries or diseases (detailed in appendices), such as myopathic gait ($e.g.$, for muscular dystrophy, muscle disease, spinal muscle weakness), ataxic gait ($e.g.$, for alcohol intoxication, brain injury), hemiplegic gait ($e.g.$, for stroke), diplegic gait ($e.g.$, for cerebral palsy, stroke, head trauma), sensory gait ($e.g.$, for tabes dorsalis, diabetes).

\begin{figure*}[t]
  \centering
  \scalebox{0.64}{
  \includegraphics[]{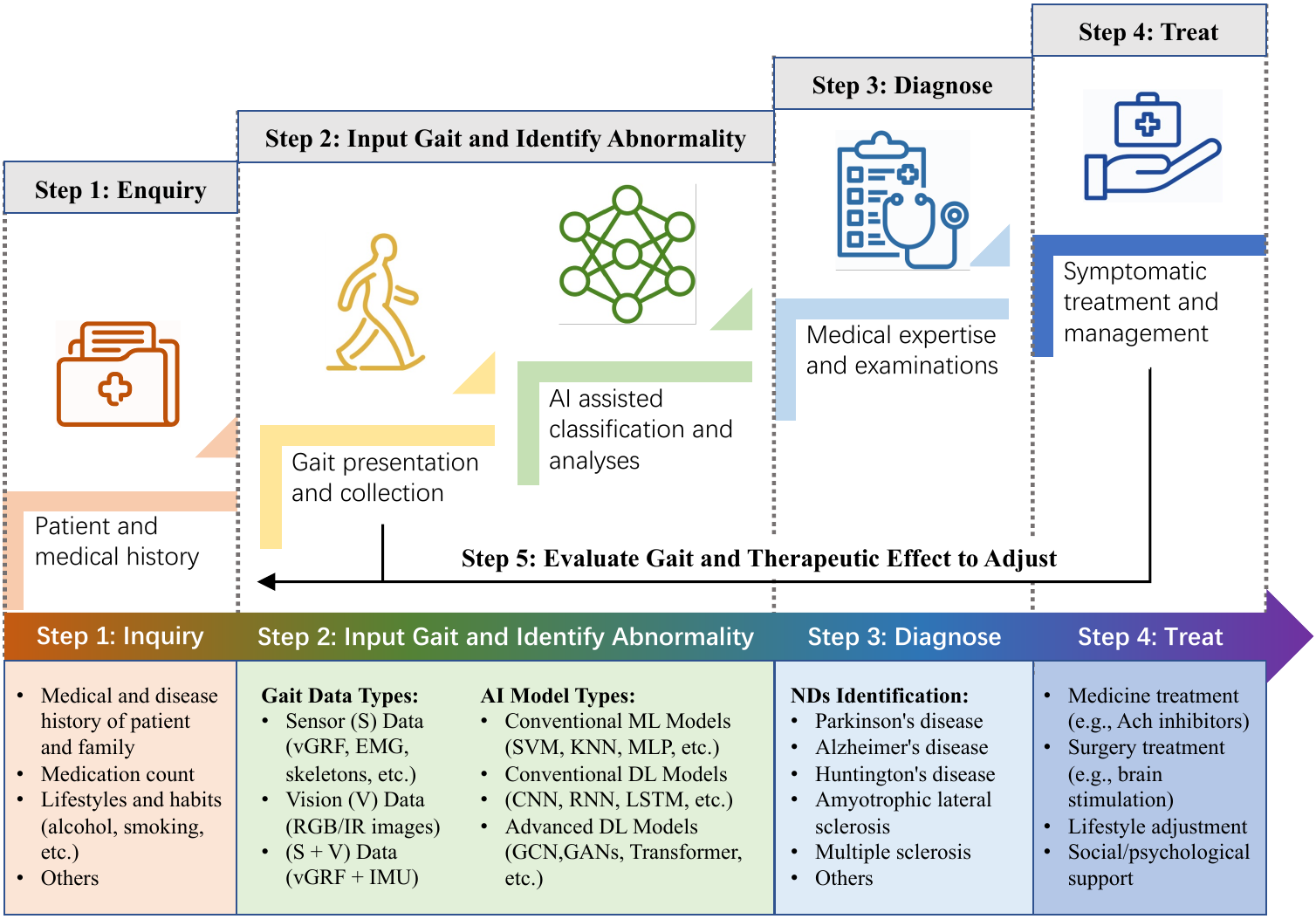}
  }
  \caption{Overview for the process of AI-assisted gait-based neurodegenerative disease diagnosis.}
  \label{process_overview}
\end{figure*}

\begin{figure}[t]
  \centering
    \begin{minipage}{0.45\textwidth}
     \centering
     \scalebox{0.43}{
  \includegraphics[]{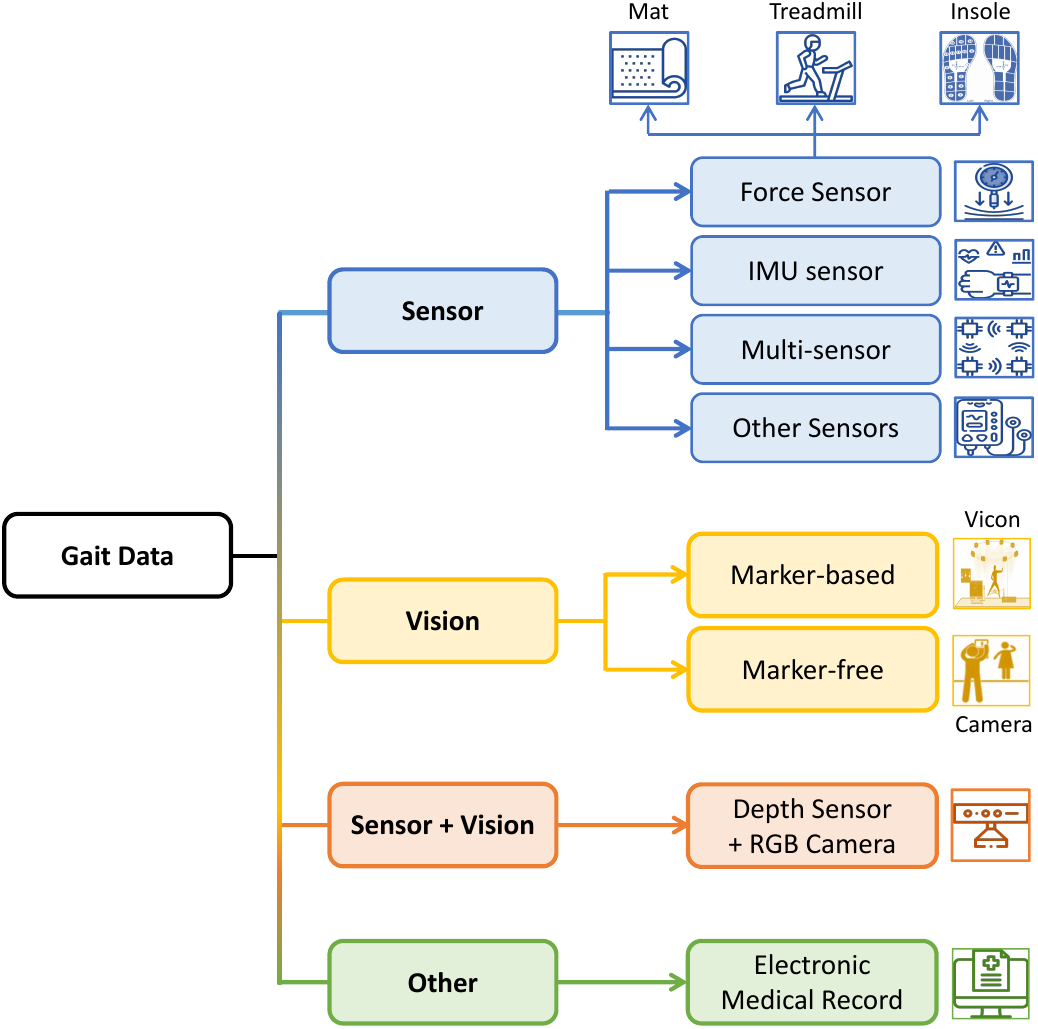}}
  \caption{The taxonomy of gait data types used for AI-assisted NDs diagnosis.}
  \label{gait_data_type}
  \end{minipage}
  \quad \quad
\begin{minipage}{0.45\textwidth}
 \centering
 \scalebox{0.18}{
  \includegraphics[]{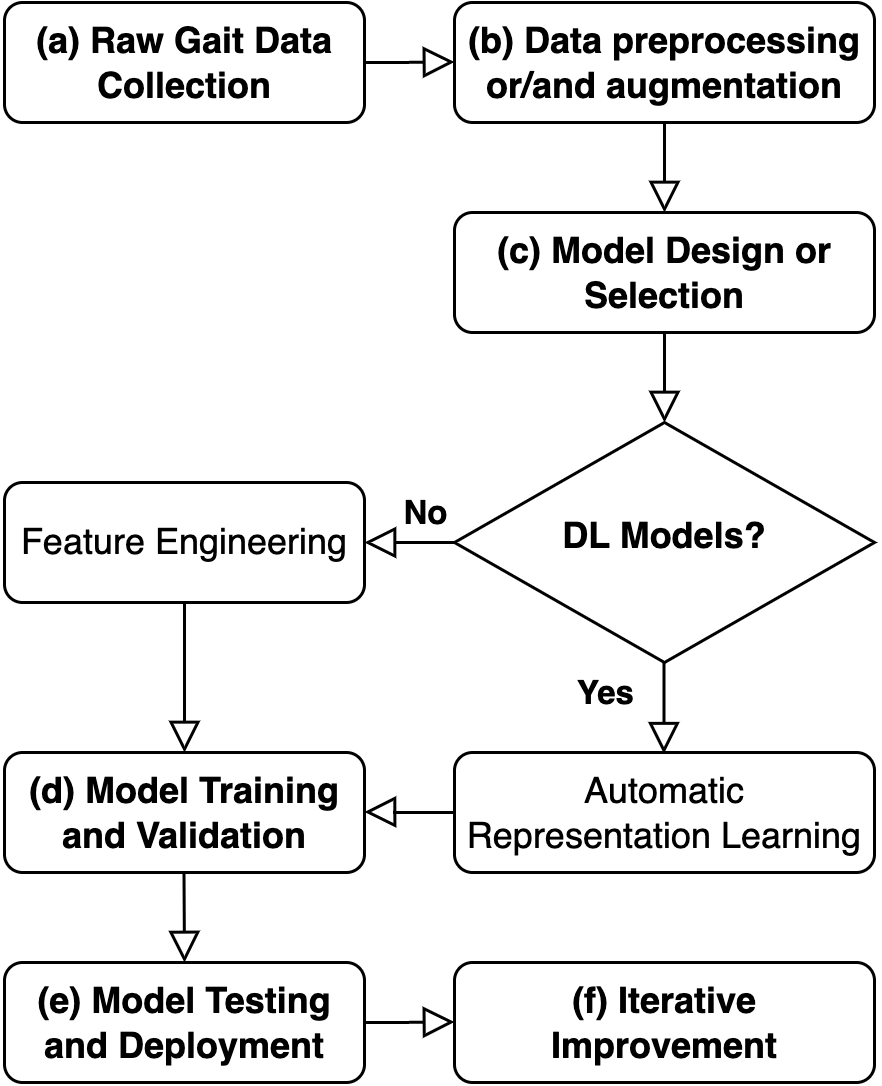}
  }
  \caption{Overview of AI model learning workflow.}
  \label{AI_learning_workflow}
  \end{minipage}
\end{figure}

\subsection{Overview of AI-assisted Gait-Based Neurodegenerative Disease Diagnosis}
\label{sec_overview_process}
The general process for AI-assisted NDs diagnosis based on human gait is presented in Fig. \ref{process_overview}. It can be summarized as five steps: (1) Enquiry; (2) Input gait and identify abnormality; (3) Diagnose; (4) Treat; (5) Evaluate therapeutic effect for adjustment, which can be illustrated as follows:

\begin{enumerate}
    \item \textbf{Enquiry}: Doctors enquiry the medical and disease history of a patient as diagnostic priors or clues for next-step disease risk assessment, disease diagnosis, and medication management.
    \item \textbf{Input Gait and Identify Abnormality}: The gait data ($e.g.,$ walking videos) of the patient are collected and inputted into a pre-trained AI model specifically designed for NDs diagnosis. In this process, the AI model assists doctors to automatically analyze the abnormal gait patterns to predict the most likely NDs (if present) in the patient. 
    \item  \textbf{Diagnose}: Doctors and medical experts synergize their domain expertise and results of other physical or neurological examinations to clinically diagnose the disease.
    \item  \textbf{Treat}: Based on the diagnostic result, a symptomatic treatment is conducted on the patient to halt disease progression, reduce pain, and improve life quality.
    \item \textbf{Evaluate Gait and Therapeutic Effect to Adjust}: The therapeutic effect is recorded into the medical history in step (1), and doctors perform the next-round assessment with step (2) and (3) to optimize the symptomatic treatment and disease management in step (4).
\end{enumerate}

It is noteworthy that exploiting AI models for gait analyses and classification ($i.e.$, Step 2) is the main part of the gait-based NDs diagnosis process.
It includes the workflow of designing, training, and validating AI models using pre-collected gait data. As shown in Fig. \ref{AI_learning_workflow}, the AI learning workflow can be divided into (a) raw gait data collection, (b) data preprocessing or/and augmentation, (c) model design or selection, (d) model training and validation, (e) model testing and deployment, and (f) iterative improvement ($e.g.,$ model fine-tuning using more data). For conventional ML models, feature engineering is usually required after step (c) to manually extract discriminative feature such as gait descriptors. 
For DL models, the design of models ($e.g.$, neural network architectures) (step (c)) and the sufficiency of training data (step (a) and (b)) are often the two most essential determinants of model performance. Since more high-quality training data can better guide the model optimization, extend the learnable feature space, and improve the generalization ability ($i.e.,$ higher performance on new testing data) of the model, it is feasible to collect more gait data while using effective data augmentation strategies ($i.e.$, generate more augmented training samples) to iteratively boost the model performance (detailed in Sec. \ref{sec_challenges_and_directions}).

\ml{

\subsection{Gait Data Taxonomy}\label{sec_gait_taxonomy}

We present the taxonomy of existing gait data types in Fig. \ref{gait_data_type} by categorizing them into four main classes based on their collection modalities. We elaborate on the common techniques used in each modality, along with their corresponding merits and demerits. 

    \begin{itemize}[leftmargin=*]
    \item \textbf{Sensor Modality}: Advancement in sensor technologies allows its application in various tasks, such as gait analysis. The sensors explored among the collected papers offer various advantages and could be classified into four types, including force, inertial, hybrid, and others \cite{alam2017vertical, hu2022machine, kaur2020predicting, perumal2016gait, das2022recent}.
        \begin{enumerate}
        \item \textbf{Force Sensor (FS)}: There are mainly two types of force or pressure capturing approaches. The first is to use precise non-wearable floor-mounted devices ($e.g$, force mats, treadmill) to measure vGRF, moments, or plantar pressure profiles in a laboratory setting \cite{das2022recent}. However, since those devices are often bulky, operating-costly, and expensive, it is not often clinically applicable. Consequently, researchers have developed some small wearable devices ($e.g.$, insole) whose application in the field has risen significantly in the past decade to combat previous limitations, however at the cost of less accuracy and reliability \cite{howell2012insole, howell2013kinetic}.
        \item \textbf{Inertial Sensor}: Also termed as Inertial Measurement Unit (IMU), is often attached to limb segments ($e.g$, foot) \cite{carvajal2022classification} or embedded in wearable devices ($e.g.$, smartwatch) to capture dynamic motion signal ($e.g.$, 3D-acceleration) of the subject, which will then be interpreted with various algorithms for the intended tasks, such as gait event detection \cite{arora2014high, klucken2013unbiased}. 
        \item \textbf{Multi-Sensor}: This category combines various types of sensor for a more comprehensive data collection. For instance, some studies \cite{perez2020identification, mileti2018measuring, li2022abnormal} utilize both IMU sensors and Force Sensitive Resistors (FSRs), while others \cite{guo2019method} employs Electroencephalogram (EEG) sensors and FSRs. These combinations harvest the strengths of each sensor type to enhance the reliability and accuracy of intended tasks.
        \item \textbf{Other Sensors}: Sensors included in this category are less exploited among collected papers and serve as a supplementary to other sensors. For instance, EEG and electromyography (EMG) sensors can be used to measure electrical activity signals of brain and muscle, respectively, to analyze their activity patterns that may correlate with specific gait abnormalities \cite{zhao2023wearable, guo2019method, you2020alzheimer, kour2023sensor, kugler2013automatic}. The rationale behind this is that NDs patients suffer from the degeneration of nerve cells, hence they manifest different brain and muscle electrical pattern from healthy subjects.
        \end{enumerate}

    \item \textbf{Vision Modality}: The vision modality employs optoelectronic Motion capture (Mocap) systems, encompassing various types of cameras ($e.g.$, Infrared (IR) Camera \cite{park2021classification}) to capture the locomotion of subjects during walking. It can be used to overcome the limitations of the traditional clinical examination (see Sec. \ref{sec_basic_concepts}). Depending on the precision requirement for the data, this modality could be further divided into two sub-categories, namely marker-based and marker-free systems \cite{kour2019computer}.
        \begin{enumerate}
        \item \textbf{Marker-based}: These approaches are developed at an early stage when the cameras are not precise enough for motion capturing. Therefore, instead of solely relying on ($e.g.$, Vicon) reflective markers to help tracking the exact location of the body landmarks during movement. Data captured by these approaches are verified in many studies with the highest precision. Therefore, these approaches are considered as the 'gold standard' in the field. However, the body-attached markers have the potential of altering gait. Additionally, this approach incurs high costs in terms of time, manpower, and expenses due to the complex setup \cite{filtjens2021modelling, aich2018machine, ferreira2022machine}.
        \item \textbf{Marker-free}: Due to the limitations of the marker-based approach and the development of the camera technology, the marker-free approach comes into appearance. This approach derives corresponding features from patients' motion videos without using any markers in the data collection process. Another highlight is the emerging application of smartphones in this field, which, despite being in its infancy, offers new insights due to its advantages like compactness, portability, and affordability. However, it still faces challenges, such as lower accuracy, in tracking movements of body parts compared to the gold standard ($i.e.,$ marker-based approaches).
        \end{enumerate}
    
    \item  \textbf{Sensor+Vision Modality}: This modality is also referred to as multi-modal modality. In comparison with single-modal approaches, this combined approach unleashes a bigger potential, such as providing more comprehensive features, capturing more subtle changes in gait, and etc. \cite{you2020alzheimer, zhao2021multimodal, chatzaki2022can}. 
    
    Microsoft Kinect, in particular, is the most applied system in this category due to its utilization of both depth sensor and RGB camera, which achieves near optimal data capturing precision while not sacrificing the advantages provided by the marker-free vision modality. More specifically, using conventional cameras to estimate skeleton coordinates can only produce coarse results. However, by combining  depth sensors, the skeleton joints' coordinates can be precisely captured. This has been verified in many studies by comparing with the gold standard approaches. Nevertheless, current Kinect system's precision reduces as the subject walks further away from the camera, resulting in a width and length-limited data collection space \cite{buongiorno2019low, tupa2015motion, reyes2019lstm, munoz2022machine}. 
    
    Besides combining RGB camera and depth sensor, there are also other ways to integrate sensor and vision modalities. For example, Tahir \textit{et al.} \cite{tahir2012parkinson} fused a marker-based IR camera and a force-sensitive platform, while Chatzaki \textit{et al.} \cite{chatzaki2022can} utilized a conventional camera, IMU, and a force platform. Additionally, Zhao \textit{et al.} \cite{zhao2021multimodal} merged some vGRF and IMU sensors with Kinect.
    
    \item \textbf{Other Modalities}: Trabassi \textit{et al.} use Electronic Medical Record (EMR) which contains various patient-related features and medical history data that are indicative of early PD symptoms or risk factors as well as the first diagnosis of gait or tremor disorders to create prediction models \cite{yuan2021accelerating}.
\end{itemize}

 \begin{figure*}
   \centering
     \subfloat[AI Model Types]{
 	 \begin{minipage}[b]{0.23\textwidth}
       \centering
       \scalebox{0.6}{
 	   \includegraphics[]{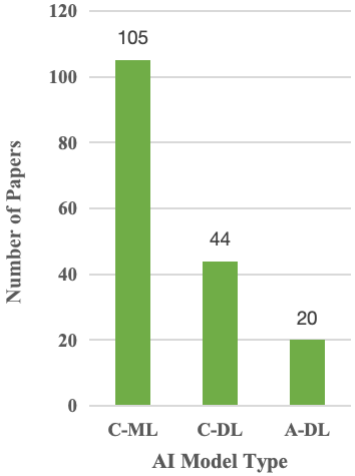}
     }
 	  \end{minipage}
 	  \label{statistics_b}
 	}
   \subfloat[NDs Types]{
 	 \begin{minipage}[b]{0.24\textwidth}
       \centering
       \scalebox{0.55}{
 	   \includegraphics[]{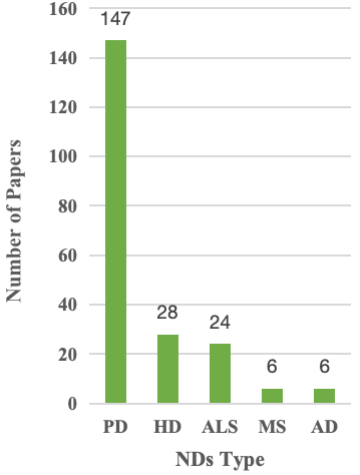}
     }
 	  \end{minipage}
 	  \label{statistics_c}
 	}
    \subfloat[Sample Sizes]{
 	 \begin{minipage}[b]{0.23\textwidth}
       \centering
       \scalebox{0.53}{
 	   \includegraphics[]{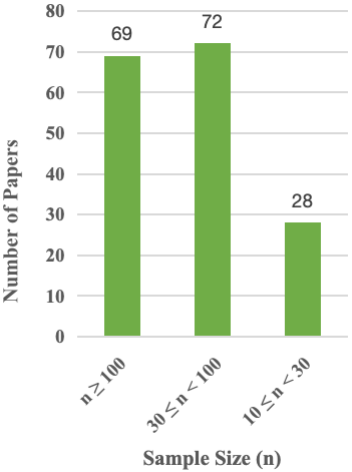}
     }
 	  \end{minipage}
 	  \label{statistics_d}
 	}
   \subfloat[Gait Data Types]{
    \begin{minipage}[b]{0.26\textwidth}
          \centering
   		 \scalebox{0.53}{
 	   \includegraphics[]{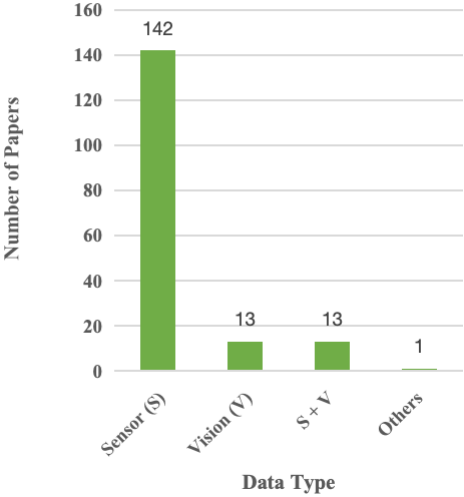}
     }
       \end{minipage}
 	  \label{statistics_f}
     }
     
     \subfloat[Sensor Types]{
    \begin{minipage}[b]{0.23\textwidth}
          \centering
   		 \scalebox{0.48}{
 	   \includegraphics[]{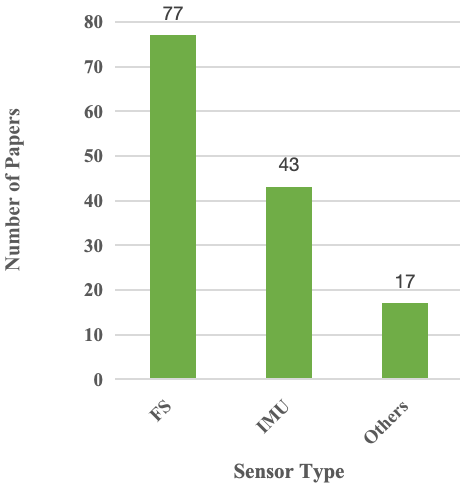}
     }
       \end{minipage}
 	  \label{statistics_g}
     }
     \ \
     \subfloat[Countries]{
 	 \begin{minipage}[b]{0.73\textwidth}
       \centering
       \scalebox{0.6}{
 	   \includegraphics[]{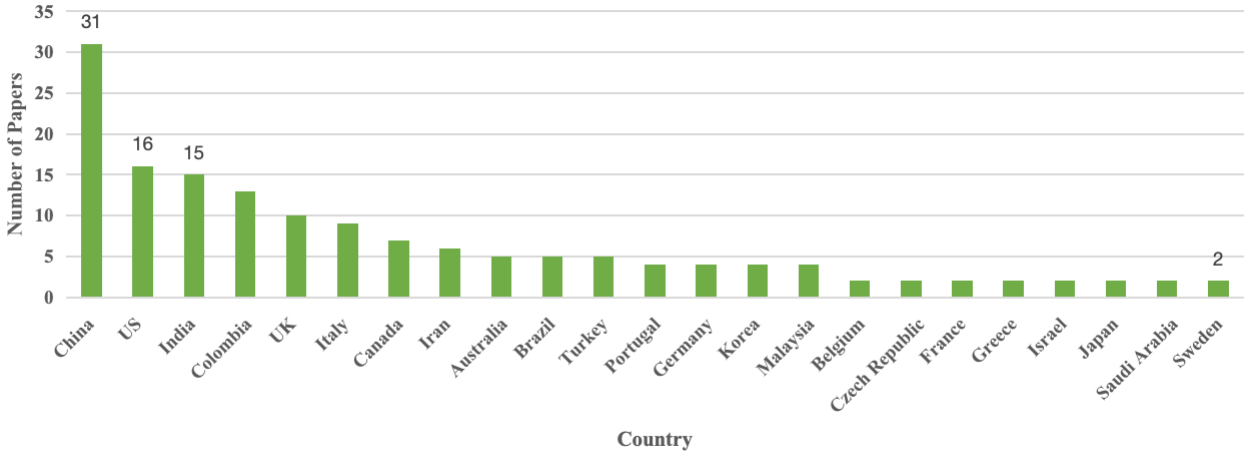}
     }
 	  \end{minipage}
 	  \label{statistics_e}
 	}
   \caption{\hc{Statistics of the included papers with different (a) AI model types, (b) NDs types, (c) experimental sample sizes, (d) gait data types, (e) sensor types, and (f) countries. We show the lowest and top three quantities in (f). Papers of different years are shown in Fig. \ref{paper_year}. Detailed analyses are provided in Sec. \ref{sec_statistics}.}}
   \label{paper_statistics}
 \end{figure*}

}
\subsection{AI Model Taxonomy}
\label{sec_AI_taxonomy}
The AI models used in existing studies can be mainly classified into three types based on the their feature learning manners and model architectures:
\begin{itemize}
    \item \textbf{Conventional Machine Learning (C-ML) Models}: C-ML models primarily use manual feature extraction and data-driven statistical techniques for pattern recognition. Examples include Decision Trees (DT) \cite{myles2004introduction}, Support Vector Machines (SVM) \cite{hearst1998support}, Linear Regression (LR) \cite{montgomery2021introduction}, Naive Bayes (NB) \cite{webb2010naive}, K-Nearest Neighbors (KNN) \cite{peterson2009k}, Random Forest (RF) \cite{breiman2001random}, etc. Artificial neural networks with very few layers such as 3-layer multilayer perceptrons (MLP) \cite{gardner1998artificial} are also viewed as C-ML models. It is noteworthy that deep learning is commonly viewed as an important branch of machine learning, thus we use “conventional machine learning” to denote \textit{classic} machine learning models apart from deep learning models.
    \item \textbf{Conventional Deep Learning (C-DL) Models}: Defined as neural networks with deep layers (typically more than 3 layers) or commonly-used deep learning backbones for automatic feature extraction and representation learning from data. Examples include Convolutional Neural Networks (CNN) \cite{gu2018recent}, Recurrent Neural Networks (RNN) \cite{medsker2001recurrent}, vanilla long short-term memory (LSTM) \cite{hochreiter1997long}, ResNet \cite{he2016deep}, etc.
    \item \textbf{Advanced Deep Learning (A-DL) Models}: Incorporates more complex or novel architectures, advanced training techniques, and elements of other AI fields like reinforcement learning into classic deep networks. Examples include graph convolutional networks (GCN) \cite{kipf2016semi}, Transformers \cite{vaswani2017attention,,fang2023hierarchical}, Generative Adversarial Networks (GANs) \cite{goodfellow2014generative}, etc.
\end{itemize}

\section{Survey Statistics and Analyses}\label{Results}
In this section, we showcase the results of our literature searching and screening, with a statistical analysis of selected papers to shed light on the current development and future trend of this field.

\subsection{Study Selection}\label{Screening}
We adopt the Preferred Reporting Items for Systematic Reviews and Meta-Analysis (PRISMA) \cite{page2021prisma} to conduct our survey, and the flow diagram of PRISMA is provided in the appendices. We summarize the literature selection process as follows:
During the paper identification stage, we retrieve \hc{845} records from PubMed, 895 records from WoS, and 596 records from Google Scholar. By comparing the collected records across these three databases, a total of \hc{2074} records remain after 262 duplicate records are removed. In the paper screening stage, after screening the titles and abstracts of the remaining \hc{2074} papers according to the inclusion criteria, \hc{248} articles are further selected for the full-text assessment. Finally, after an in-depth review of the full texts of these selected papers, 103 articles are excluded and \hc{145} regular articles are included in this survey for further information extraction and content summarization. The list of excluded papers from the full-text screening stage, along with the reasons for their exclusion, is detailed in our appendices. Moreover, 14 eligible survey papers \cite{kour2022vision,khera2022age,fraiwan2021computer,kour2019computer,kour2023sensor,di2020gait,pardoel2019wearable,gupta2023new,figueiredo2018automatic,vienne2017inertial,das2022recent,salchow2022emerging,loh2021application,ayaz2023automated} are found in the screening stage. From these survey papers, we identify another 99 potentially-matched articles via screening their titles and abstracts. After the full-text assessment, we include 24 regular articles from those. In summary, we include a total of 164 eligible articles, comprising \hc{145} from the academic databases and 24 from survey papers, in our survey.

\subsection{Statistics of Selected Papers}
\label{sec_statistics}

To provide an intuitive overview for all \hc{169} selected papers, we present the distribution of papers from different aspects: year of publication (see Fig. \ref{paper_year}), country of publication, type of NDs studied, type of gait data, type of AI models, and size of samples (see Fig. \ref{paper_statistics}). 

\subsubsection{Distribution of Publication Years}\label{published_year_analysis} As presented in Fig. \ref{paper_year}, from 2012 to 2022, the AI-assisted NDs diagnosis from human gait has attracted growing attention in the research community. Especially in the last three years, the annual number of related research papers has exceeded 21, reaching a peak of 36 in 2022. 

\xj{\subsubsection{Timeline of Technical Advancements}

Figure \ref{pd_timeline} illustrates the timeline of technical advancements in gait-based PD diagnosis. The timeline, spanning from 2012 to 2023, highlights significant advancements in both the models and data types utilized for gait-based PD diagnosis. Throughout this period, SVM, as exemplified by models in \cite{tahir2012parkinson} (ID = P-1) and \cite{dotov2023coordination} (ID = P-122) , consistently demonstrated their efficacy and reliability in C-ML approaches for diagnosing PD from 2012 to 2023.

In 2016, a landmark innovation emerged with the introduction of neural networks in \cite{wang2016recognizing} (ID = P-15), marking the first use of DL for PD diagnosis. By 2023, the field saw the development of a more sophisticated DL model, static-dynamic temporal networks, which integrated one-dimensional and two-dimensional convolutional networks along with attention mechanisms. This advanced model, featured in \cite{dong2023static} (ID = P-123), achieved a remarkable diagnostic accuracy of 96.7\%.

Sensor-based data, including EMG and IMUs, gained prominence as effective tools for enhancing diagnostic precision from 2012 to 2023, as evidenced by their application in \cite{kugler2013automatic} (ID = P-2) and \cite{goh2022gait} (ID = P-112). Vision-based approaches also evolved, with models like \cite{reyes2019lstm} (ID = P-43)  and \cite{zhang2023wm} (ID = P-121) utilizing Kinect and smartphone cameras, respectively, to extract spatial and temporal gait features from video data.

In 2022, multimodal methodologies gained traction. For instance, \cite{kour2022vision} (ID = P-113) combined sensor data from DSLR cameras and passive markers to integrate complementary inputs, significantly improving the robustness of PD diagnosis. The trend toward combining sensor and vision data culminated in \cite{zhao2021multimodal} (ID = P-106), where advanced DL models leveraged inputs from Kinect cameras, vGRF, and IMUs for a comprehensive analysis of gait dynamics.

This timeline reflects the continuous evolution of gait-based PD diagnostic methods, progressing from traditional machine learning techniques to cutting-edge deep learning models and multimodal approaches, paving the way for more precise and holistic diagnostic capabilities. We also provide a more comprehensive table on the timeline of gait-based PD diagnosis and the combinatorial diagnosis of different NDs in our appendices.

\begin{figure*}[t]
    \centering
    \scalebox{0.55}{
    \includegraphics{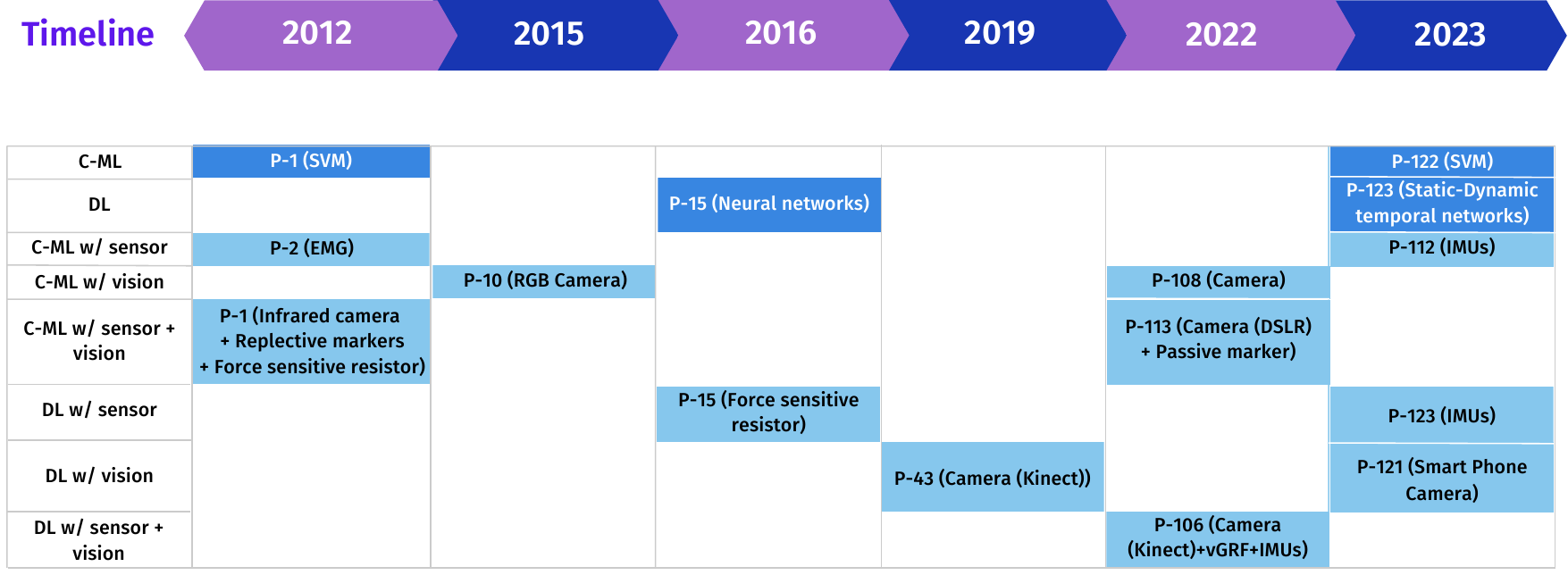}
    }
    \caption{\hc{Timeline for technical advancement of gait-based PD diagnosis.}}
    \label{pd_timeline}
\end{figure*}
}

\subsubsection{Distribution of Countries}\label{country_analysis} The results in Fig. \ref{statistics_e} suggest that China, the United States (US), and India are the top 3 contributing countries to this research field, with \hc{31 papers ($18.3\%$), 15 papers ($8.9\%$), and  15 papers ($8.9\%$), }respectively. \hc{This approximately matches the real-world situation where  NDs patients of these countries constitute the three largest groups globally, thus a broader research resources and efforts are put into this important topic}. It is also observed that, within the top 10 countries with the most publications, the number of developed countries (US, UK, Italy, Canada, Australia) participated in this research field is equal to that of the developing countries (China, India, Colombia, Iran, Brazil). However, among these 10 countries, \hc{the total publications from developing countries constitutes $59.8\%$ (70 publications), whereas that from developed countries accounts for only $40.2\%$ (47 publications)}.

\subsubsection{Distribution of ND Types}\label{ND_type_analysis} Fig. \ref{statistics_c} shows the numbers of papers focused on different NDs types. Note that the studies containing diagnosis of multiple NDs (\hc{21} articles) are also included into each NDs type ($e.g.$, a study simultaneously diagnosing PD, HD, and ALS are counted as one PD/HD/ALS study). 
\hc{The statistics show that PD receives the most research focus among all the NDs, including a total of 147 articles ($69.7\%$). The second most studied ND's type is HD with 28 papers ($13.3\%$), while there are 24 ($11.4\%$), 6 ($2.8\%$), 6 ($2.8\%$) papers focused on ALS, MS, and AD respectively. }Interestingly, although AD is the most populated diseases among all NDs (see Fig. \ref{overview_pop}), the study of its diagnosis based on gait has yet to receive much attention, possibly because other NDs patients, such as PD patients, may exhibit more evident abnormal gaits that can be exploited to perform more reliable diagnosis.
Promisingly, more and more studies \cite{mannini2016machine,daliri2012automatic,xia2015classification,zhao2021multimodal,beyrami2020robust,xia2016novel,patil2019clinical,creagh2020smartphone} have demonstrated the feasibility of leveraging gait data to identify HD, AD, ALS, and MS, therefore a deeper exploration of these research directions should be conducted.

\subsubsection{Distribution of Gait Data Types}\label{data_type_analysis} As reported in Fig. \ref{statistics_f}, \hc{142 studies ($84.0\%$)} utilize sensors including Inertial Measurement Units and force/pressure sensitive resistors to collect vertical Ground Reaction force data. These sensors can capture detailed and dynamic information about body movements and forces involved in walking to analyze gait cycles and parameters. There are \hc{13 studies ($7.7\%$)} focused on vision-based devices such as RGB cameras and infrared cameras to extract gait patterns from videos. A same number of studies (13 articles) combine both sensor-based and vision-based devices to collect multi-modal gait data for learning. 

\hc{Fig. \ref{statistics_g} further shows the distribution of sensor types used in 142 sensor-based studies: The force sensors ($e.g.,$ force/pressure sensitive resistors) and IMUs are the two most widely-used sensor types, occupying $55.6\%$ (79 studies) and $31.0\%$ (44 studies) respectively}. For other sensors (19 studies), there are 5 studies combining both FS and IMUs as gait data sources, while 5 studies utilize gait analysis system (GAS) ($e.g.,$ mats, walkways, treadmills). The low-cost and non-invasive motion capture device, Kinect \cite{seifallahi2022alzheimer,munoz2022machine}, is also used as both a sensor and a vision-based device for gait data collection \hc{(3 studies)}. 
We provide a further discussion on exploiting sensor-based ($e.g.,$ Kinect-based) 3D skeleton data for NDs diagnosis in Sec. \ref{sec_research_vision}.

\subsubsection{Distribution of AI Models}\label{model_type_analysis} 
\hc{As shown in Fig. \ref{statistics_b}, in the past decade, only 20 articles ($11.8\%$) propose novel A-DL models for NDs diagnosis from gait, while most of the studies use conventional AI methods such as C-ML and C-DL with 105 ($62.1\%$) and 44 articles ($26.0\%$) respectively}. However, most conventional AI models possess limited performance on large-scale medical data, thereby underscoring the challenge and urgency of developing more advanced AI models in this field. Such development requires more efforts and cooperation from AI scientists and healthcare professionals in the future.

\subsubsection{Distribution of Sample Sizes}\label{sample_size_analysis}
\hc{As shown in Fig. \ref{statistics_d}, 141 papers ($83.4\%$) conduct their experiments with more than 30 samples ($i.e.,$ number of patients plus number of controls), and the number of papers that possess small-size samples ($10\leq n \leq 30$) only occupies $16.6\%$ (28 papers)}. This suggests that most AI models require a relatively larger size of samples and gait data to obtain satisfactory performance, otherwise it might be insufficient for obtaining reliable diagnosis. This also aligns with the recent advancements of medical deep learning based AI models that necessitate massive medical labeled data.

\section{Overview of Existing Studies}
\label{sec_overview_studies}
In this section, we systematically review existing studies related to AI-assisted NDs diagnosis based on human gait. First, a quality evaluation criterion based on the novelty of AI models, comprehensiveness of method comparison, and sufficiency of experimental samples, is devised to quantitatively measure the quality of each study.
Then, we categorize all studies by different NDs types (PD, HD, AD, ALS, MS) with a data summary for them (shown in Table \ref{PD_1}, \ref{PD_2}, \ref{PD_3}, \ref{PD_4}, \ref{combine_table}, and \ref{AD_ALS_HD_MS_table}). 
Last, according to the quality scores, we provide a discussion on representative studies.

\subsection{Quality Evaluation Criterion}
\label{quality_criterion}
To comprehensively measure the quality of each study, we empirically grade the novelty of AI models, comprehensiveness of method comparison, and sufficiency of experimental samples at three levels with the score 1, 2, or 3. \hc{For novelty (defined as “\textit{risk-evaluated novelty}”) scores, we not only measure how advance the AI architecture is, but also empirically consider the risk of model over-fitting in real-world application scenarios ($i.e.,$ learning on different sizes of data). For example, C-ML models can be advantageous with higher interpretability and reliability on small datasets, and it may even outperform deep learning methods that could over-fit in some cases \cite{srivastava2014dropout}.} We sum \hc{different kinds scores} and divide them by the maximum score to get a normalized quality score $S_{Q}$ ranging from 0 to 1. $S_{Q}$ is computed for each study, which are defined as follows:
\begin{itemize}
    \item Score ($S_{n}$) for \textit{\hc{Risk-Evaluated} Novelty of AI Models} 
    \begin{enumerate}
    \item \hc{We first compute the basic score to assess the novelty level of the AI architecture:}
    \begin{itemize}
        \item \hc{Basic Score $\overline{S}_{n}=1$}: The main proposed method belongs to or contains only C-ML models.
        \item \hc{Basic Score $\overline{S}_{n}=2$}: The main proposed method belongs to or contains C-DL models.
        \item \hc{Basic Score $\overline{S}_{n}=3$}: The main proposed method belongs to or contains A-DL models.
    \end{itemize}
    \item \hc{We then compute the risk score of over-fitting based on the data size ($i.e.,$ size of subjects):}
        \begin{itemize}
            \item \hc{Risk Score $\Hat{S}_{n}=1$: The deep learning models (C-DL and A-DL) are learned on the small-size dataset with less than 30 subjects.}
            \item \hc{Risk Score $\Hat{S}_{n}=0$:The deep learning models (C-DL and A-DL) are learned on the large-size dataset with more than 100 subjects, or medium-size dataset with more than 30 and less than 100 subjects.}
            \item \hc{Risk Score $\Hat{S}_{n}=-1$: The C-ML models are learned on the small-size data with less than 30 subjects, or medium-size dataset with more than 30 and less than 100 subjects.}
        \end{itemize}
    \item \hc{Finally, the risk-evaluated novelty score is computed by subtracting the risk score from the basic score, $i.e.$, $S_{n}=\overline{S}_{n}-\Hat{S}_{n}$.}
    \end{enumerate}
    \item Score ($S_{c}$) for \textit{Comprehensiveness of Method Comparison} 
        \begin{enumerate}
        \item Score $S_{c}=1$: The study only compares different settings of the proposed method, without comparing with any other existing methods or studies.
        \item Score $S_{c}=2$: The study compares the proposed method with one published method or study.
        \item Score $S_{c}=3$: The study compares the proposed method with more than one published methods or studies.
    \end{enumerate}
    \item Score ($S_{s}$) for \textit{Sufficiency of Experimental Samples} 
        \begin{enumerate}
        \item Score $S_{s}=1$: The size of experimental samples ($i.e.$, number of patients plus number of controls) is less than 30.
        \item Score $S_{s}=2$: The size of experimental samples is equal to or more than 30, and less than 100.
        \item Score $S_{s}=3$: The size of experimental samples is equal to or more than 100.
    \end{enumerate}
\end{itemize}
To normalize the final quality score $S_{Q}$ to the value range from 0 to 1, we sum the scores from all three aspects and divide them by the maximum score:
\begin{equation}
    S_{Q}=\frac{S_{n} + S_{c} + S_{s}}{9}
\end{equation}
The quality score $S_{Q}$ is computed for each study and presented in the tables. We empirically set $S_{Q}>0.6$ as the threshold for selecting representative high-quality studies.

\begin{table}[t]
\centering
\caption{Number of studies related to different NDs (PD, HD, MS, AD, ALS) and different model types (C-ML, C-DL, A-DL). “Combinatorial” denotes studies that simultaneously contain diagnosis of multiple NDs including PD, ALS, and HD. \hc{The numbers of studies under different quality score thresholds $S_{Q}>0.5, 0.6, 0.7$ are also presented.}}
\label{num_types}
\scalebox{0.65}{
\renewcommand\arraystretch{1.1}{
\setlength{\tabcolsep}{8mm}{
\hc{
\begin{tabular}{l|ccc|c|c|c|c}
\hline
\textbf{Type} & \textbf{C-ML} & \textbf{C-DL} & \textbf{A-DL} & \textbf{Total} & \textbf{$\boldsymbol{S_{Q}>0.5}$} & \textbf{$\boldsymbol{S_{Q}>0.6}$} & \textbf{$\boldsymbol{S_{Q}>0.7}$} \\ \hline
\textbf{PD} & 78 & 34 & 14 & \textbf{126} & 112 & 56 & 43 \\ \hline
\textbf{\begin{tabular}[c]{@{}l@{}}Combinatorial\\ (PD, ALS, HD)\end{tabular}} & 11 & 7 & 3 & \textbf{21} & 21 & 12 & 10 \\ \hline
\textbf{HD} & 6 & 1 & 0 & \textbf{7} & 6 & 2 & 1 \\ \hline
\textbf{MS} & 4 & 1 & 1 & \textbf{6} & 2 & 2 & 1 \\ \hline
\textbf{AD} & 3 & 1 & 2 & \textbf{6} & 6 & 4 & 2 \\ \hline
\textbf{ALS} & 3 & 0 & 0 & \textbf{3} & 1 & 1 & 0 \\ \hline
\textbf{Total} & \textbf{105} & \textbf{44} & \textbf{20} & \textbf{169} & \textbf{148} & \textbf{77} & \textbf{57} \\ \hline
\end{tabular}
}
}
}
}
\end{table}

 \begin{figure*}
   \centering
     \subfloat[PD Studies with Different Thresholds]{
    \begin{minipage}[b]{0.48\textwidth}
          \centering
   		 \scalebox{0.7}{
 	   \includegraphics[]{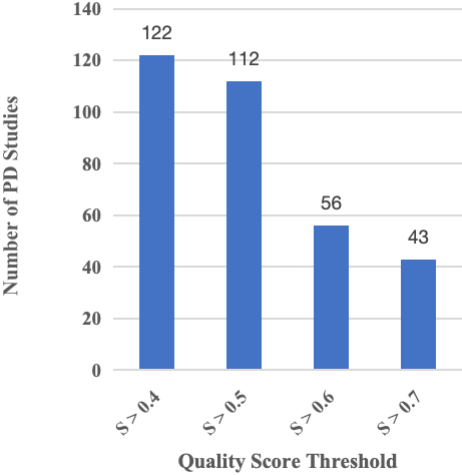}
     }
       \end{minipage}
     }
     \ \
     \subfloat[Combinatorial Studies with Different Thresholds]{
 	 \begin{minipage}[b]{0.48\textwidth}
       \centering
       \scalebox{0.7}{
 	   \includegraphics[]{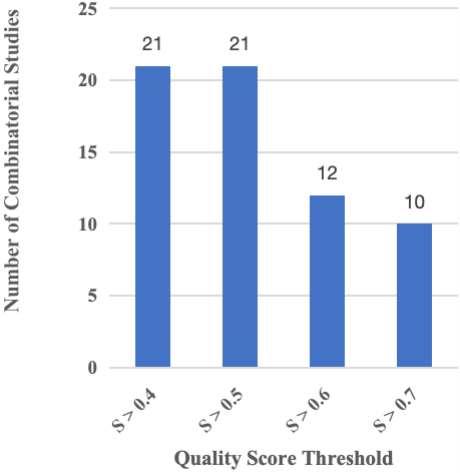}
     }
 	  \end{minipage}
 	  \label{statistics_e}
 	}
   \caption{\hc{Comparison of numbers of selected PD studies (Fig. (a)) and combinatorial studies (Fig. (b)) when setting different quality score thresholds ($S_{Q}$=0.4, 0.5, 0.6, 0.7).}}
   \label{quality_score_thresholds}
 \end{figure*}

\subsection{Parkinson's Disease (PD)}\label{sec_PD}

As shown in Table \ref{PD_1}, \ref{PD_2}, \ref{PD_3}, \ref{PD_4} and \ref{combine_table}, 
there are 126 studies that focus on PD diagnosis based on human gait. It is worth mentioning that there are 21 studies, each of which simultaneously includes the diagnosis of PD and other NDs, as shown in Table \ref{combine_table} (referred to as \textit{“Combinatorial Diagnosis of Different NDs”}). We separately discuss these studies in Sec. \ref{sec_Combine}. Based on our survey, \hc{there are 78, 34, and 14 PD studies using C-ML, C-DL, and A-DL models\ml{,} respectively.}
It should be noted that there are a large number of PD studies (up to 126 studies). Since the studies with high-quality AI models and diagnostic results are the focus of our survey, we first categorize all studies by three AI model types (C-ML, C-DL, A-DL), and then select representative studies with quality score $S_{Q}>0.6$ for content summary and analysis. \hc{It is worth noting that we empirically set the quality score threshold to 0.6 because it can select more representative ($e.g.,$ top 40\% to 50\%) high-quality studies than setting $S_{Q}>0.5$ (88.9\% studies), as shown in Fig. \ref{quality_score_thresholds}.
}

\subsubsection{C-ML Models for PD Diagnosis}
Khorasani et al. \cite{khorasani2014hmm} (Table \ref{PD_1}, ID = P-7) utilized Hidden Markov Models (HMM) with Gaussian Mixtures to classify gait data and differentiate PD patients from healthy subjects. Using stride interval as the distinguishing feature, the HMM method achieved an accuracy rate of 90.3\%. 
In their study, the authors, Tupa et al. \cite{tupa2015motion} (ID = P-10). presented a methodology using Microsoft Kinect sensors to detect gait attributes and recognize gait of PD patients. The leg length, stride length, and gait velocity were extracted to achieve high-accuracy classification using neural networks (NN). This study suggested the potential of using Kinect sensors for gait analysis and disorder recognition.
Zeng et al. \cite{zeng2016parkinson} (ID = P-14) proposed a radial basis function neural networks (RBF-NN) based method for classifying PD patients and healthy individuals using gait characteristics derived from vertical ground reaction forces, achieving an over 90\% accuracy.
Ertuugrul et al. \cite{ertuugrul2016detection} (ID = P-16) focused on a technique called Shifted One-Dimensional Local Binary Patterns (Shifted 1D-LBP) to track local changes in time series data. Combining with machine learning model, this work demonstrated Shifted 1D-LBP's potential in real-time NDs detection applications.
The study of Alam et al. \cite{alam2017vertical} (ID = P-20) utilized the vGRF data to extract gait features and feeded them into the SVM classifier to obtain a good accuracy result of 93.6\%.
Joshi et al. \cite{joshi2017automatic} (ID = P-21) explored the use of wavelet analysis combined with SVM for the identification of PD based on spatio-temporal gait variables, achieving a classification accuracy of 90.32\%. The study demonstrated the potential of using wavelet analysis as an efficient method for classifying PD and healthy subjects.
Rovini et al. \cite{rovini2018comparative} (ID = P-30) utilized a wearable inertial device to collect motor data from healthy subjects, individuals with idiopathic hyposmia (IH), and PD patients during lower limb assessments, and developed a machine learning-based system with SVM, RF, NB to classify the three groups. 
In their work , Aich et al. \cite{aich2018validation} (ID = P-32) quantified gait parameters objectively using wearable accelerometers, and compare them with a motion capture system to automatically discriminate patients with PD using machine learning models such as SVM, KNN, and DT. The proposed approach achieves an accuracy of approximately 89.1\% and suggests that the wearable accelerometer-based system is suitable for assessing and monitoring PD and Freezing of gait \ml{(FoG)} in real-life scenarios.
Elden et al. \cite{elden2018computer} (ID = P-38) emphasized the benefit of utilizing a hybrid of linear and non-linear features extracted from human gait time series data to improve the NDs classification system which employed SVM with a radial basis kernel function.
Khourly et al. \cite{khoury2019data} (ID = P-42) leveraged different classic machine learning methods (KNN, DT, RF, NB, SVM, K-Means) to identify PD from vGRFs collected from gait cycles. The structure of PD recognition process from data acquisition to performance evaluation was also systematically presented in this study.
Andrei et al. \cite{andrei2019parkinson} (ID = P-47) devised an automated system with force sensors to extract gait parameters and then used these parameters as input to the SVM model to diagnosis PD. 
Zeng et al. \cite{zeng2019classification} (ID = P-50) proposed a novel method for classifying gait patterns between patients with PD and healthy controls. This method reconstructed the phase space of vGRF and decomposed the gait dynamics using empirical mode decomposition, which were fed to NN for classification with accuracy of 98.8\%.
The effectiveness of Extreme Gradient Boosting (XGBoost) and Artificial Neural Network (ANN) models for monitoring the gait of PD patients has been explored, achieving an accuracy of 41.0\% (Hughes et al. \cite{hughes2019models}).
The study by Som et al. \cite{som2020unsupervised} (ID = P-60) explored the use of single-sensor accelerometer data from healthy subjects to extract features relevant to multi-sensor accelerometer gait data for PD classification. They used a pre-trained convolutional autoencoder as the source model and a simple multi-layer perceptron as the target model to classify PD and achieved a 68.92\% accuracy. 
A machine learning based gait classification system with different models, such as SVM and RT, was presented to assist in diagnosing PD form vGRF (Balaji et al. \cite{balaji2020supervised} (ID = P-65)).
Balaji et al. also \cite{balaji2021data} (ID = P-71) extracted spatio-temporal gait features such as stride time and stance time from vGRF and exploits different classic machine learning models (KNN, NB, SVM) to classify health control and PD subjects with an accuracy of 88.5-98.8\%.
In the work by Ghaderyan et al. \cite{ghaderyan2021inter} (ID = P-81), a time-varying singular value decomposition method was designed to extract the most useful and informative part of the vGRF signal data, and then sparse non-negative least squares (NNLS) approach was leveraged to classify PD and HCs, achieving 97.2\% accuracy.
Mehra et al. \cite{mehra2022design} (ID = P-87) utilized RF to identify PD patients from vGRF and achieves up to 98.8\% accuracy. 
The SVM model was used in \cite{wang2022gait} (ID = P-111) to classify PD and HC gait from vGRF data, with an accuracy of 98.2\%.
\hc{Johnson et al. \cite{johnson2024wearable} (ID = P-125, $S_{Q}=0.88$) addressed early-stage PD monitoring and diagnostic challenges using consumer-grade wearable devices and sensor data from the WATCH-PD study. Data were gathered over a year from early-stage PD patients (n = 82) and age-matched controls (n = 50). Leveraging disease- and behavior-relevant feature engineering with multivariate machine learning, the authors developed a PD screening tool based on a random forest model. Achieving 92.3\% accuracy, 90.0\% sensitivity, and 100\% specificity (AUC = 0.92), this approach underscores the potential of remotely monitored wearable devices for population-wide PD screening and monitoring.}

\subsubsection{C-DL Models for PD Diagnosis}
Acsurouglu et al. \cite{acsurouglu2018parkinson} (ID = P-37) proposed a Locally Weighted Random Forest (LWRF) regression model to predict the PD symptoms using time- and frequency-domain features derived from vGRF signals collected using foot worn sensors.
Zhao et al. \cite{zhao2018hybrid} (ID = P-39) developed a hybrid model combining Long Short-Term Memory and Convolutional Neural Network for PD's detection and severity rating. The model captured the spatio-temporal characteristics of gait dynamics using vGRF signals and offered a multi-category classification approach for PD severity, beyond the binary detection typically provided by other ML methods.
The study by Reyes et al. \cite{reyes2019lstm} (ID = P-43) leveraged convolutional LSTM to identify PD patients from Kinect-based skeleton data and achieved up to 83\% accuracy. It also suggested that the data preprocessing ($e.g.$, cropped data) could help boost the classification performance.
El et al. \cite{el2020deep} (ID = P-52) devised a deep neural network (DNN) based model to identify Parkinsonian gait based on vGRF data and achieved 98.7\% accuracy. This was also the first study to predict the Unified Parkinson's Disease Rating Scale (UPDRS) severity ($i.e.$, Parkinsion severity rate) of a subject and achieved a 85.3\% accuracy.
Xia et al. \cite{xia2019dual} (ID = P-58) utilized both CNN and attention-enhanced LSTM networks for left and right gait analysis, which helps achieved highly competitive PD classification accuracy (98.03\%). 
Moon et al. \cite{moon2020classification} (ID = P-63) used balance and gait variables obtained with wearable inertial motion sensors to differentiate PD patients using machine learning models (NN, SVM, KNN, DT, RF). 
Gong et al. \cite{gong2020novel} (ID = P-64) applied a mask R-CNN model for extracting human silhouettes from video frames, and then obtained gait energy images (GEIs) from human silhouettes as features for PD gait classification. 
Butt et al. \cite{butt2020biomechanical} (ID = P-66) presented a novel Bidirectional Long Short-Term Memory (Bi-LSTM) model to classify PD from HC using motion data, acquired during performing 13 tasks derived from the Movement Disorder Society-Unified Parkinson's Disease Rating Scale III (MDS-UPDRS III), with wearable sensors.
Zhang et al. \cite{zhang2020deep} (ID = P-67) reported a smart-phone based PD diagnosis method by utilizing real-world accelerometer records and achieved an AUC score of 0.86. This method demonstrated the superior accuracy for at-home screening of PD and other NDD~(Neurodegenerative Disease) conditions affecting movement while acknowledges limitations such as potential over-enrollment of tremor-dominant PD subjects in their experiment.
Yurdakul et al. \cite{yurdakul2020detection} (ID = P-68) used a novel technique called Neighborhood Representation Local Binary Pattern (NR-LBP) and ANN to model the relationship between statistical features extracted from vGRF data and PD symptoms.
Balaji et al. \cite{balaji2021automatic} (ID = P-69) utilized LSTM to differentiate the PD patients and healthy subjects based on the vGRF and gait time series, which performed better than other state-of-the-art methods with binary-classification accuracy of 98.6\%.
The Bi-LSTM model was leveraged in the work by Pham et al. \cite{pham2021classification} (ID = P-70)
to classify PD gait using vGRF signals from single sensors. This study revealed that combining time-frequency and time-space features of physiological signals were more effective and could boost the classification accuracy of deep-learning models.
Force domain, peak domain, and abnormality domain features (22 features) were combined to identify PD patients based on CNN and XGBoost, with an accuracy of 95.68\%-97.32\% (Ma et al. \cite{ma2023explainable} (ID = P-74)).
Yuan et al. \cite{yuan2021accelerating} (ID = P-76) characterized the pre-diagnostic period of PD and develops a ML model (\textit{i.e.,} LR) and DL model (\textit{i.e.,} RNN) for PD diagnosis. 
The authors focused on identifying changes in gait patterns associated with PD by analyzing the spatio-temporal data of vGRF. They proposed a model based on CNN that evaluated the changes in vGRF and identify the Parkinsonian gait from healthy gait with an accuracy of 97.0\% (Ahamed et al. \cite{ahamed2021spatiotemporal} (ID = P-84)).
Carvajal et al. \cite{carvajal2022classification} (ID = P-86) combined CNN and gated recurrent units (GRUs) to perform classification between PD and health  subjects from IMU signals, achieving a 83.7-92.7\% accuracy. The results suggested that the combination of temporal and spectral information was more effective in classifying gait patterns of PD patients.
Brand et al. \cite{brand2022gait} (ID = P-104) devised an anomaly detection algorithm to classify PD and healthy controls based on wrist-sensor recordings (tri-axial accelerometer data) and achieved a 83.3\% accuracy.
Baby et al. \cite{baby2017parkinsons} (ID = P-115) suggested a real-time PD identification method using wavelet transform-based features extracted from vGRF data. Experimental results using NN with resilient back-propagation algorithm achieved an average efficiency of 86.75\%.
2D CNN was leveraged by Divyashree et al. \cite{divyashree2022ai} (ID = P-116) to diagnose PD at an early stage from vGRF signal data and achieved 98\% accuracy.
Sun et al. \cite{sun2023novel} (ID = P-120) proposed a plantar pressure analysis method to identify the gait dynamics of PD patients using a constant radial basis function neural network (RBFNN).

\subsubsection{A-DL Models for PD Diagnosis}
In the work of Wang et al.\cite{wang2016recognizing} (ID = P-15), a novel hybrid model based on BP neural networks was proposed to classify PD and HC, which outperformed four baseline methods on a public dataset with an average accuracy of 87.9\%.
Jane \cite{jane2016q} (ID = P-17) introduced a Q-backpropagated Time Delay Neural Network (Q-BTDNN), which harvested the temporal gait data collected from wearable devices, combined with the Q-backpropagation (Q-BP) algorithm for predicting gait disturbances in PD.
The study of Pham et al. \cite{pham2017tensor} (ID = P-31) proposed a computational method based on tensor decomposition for analyzing gait dynamics in PD from multi-sensor time series data, achieving an accuracy of 86.2\%. 
Alharthi et al. \cite{alharthi2020gait} (ID = P-75) used deep CNN to classify gait deterioration in PD patients based on spatio-temporal vGRF signals and achieves 95.5\% accuracy in average. The study suggested that the heel strike and body balance are the most indicative gait elements in PD classification.
Nguyen et al. \cite{nguyen2021classification} (ID = P-82) developed an NN model to classify PD and HC based on the vGRF dataset collected from a public dataset and achieved an accuracy result of 84.8\%.
Salmi et al. \cite{salimi2023type} (ID = P-85) proposed an interval type-2 fuzzy neural network with novel quasi-Levenberg-Marquardt (qLM) learning approach to identify PD patients from the gait cycle based vGRF signals. Based on the 10 clinical features extracted from 14 gait features, the model achieved a 97.6\% accuracy with a smaller size of parameters and a good interpretability ($i.e.$, rules based on fuzzy logic). 
In the work by Yang et al. \cite{yang2022pd} (ID = P-93), the authors proposed a novel architecture PD-ResNet based on ResNet \cite{he2016deep} and used it to differentiate PD and HC. Their result outperformed many existing state-of-the-art models with 95.5\% accuracy. 
\hc{
A hybrid deep learning model combining CNN and locally weighted RF (LWRF) was devised to perform PD detection tasks from time and frequency features of vGRF, achieving a 99.5\% accuracy (Acsurouglu et al. \cite{acsurouglu2022deep} (ID = P-97).}
\hc{
Zhao et al. \cite{zhao2021multimodal} (ID = P-106) proposed a correlative memory neural network (CorrMNN) scheme to extract spatio-temporal features to discriminate NDs' gait from normal gait based on vGRF, and achieved up to 100\% accuracy.} 
A video-based SlowFast GCN network was proposed by Zeng et al. \cite{zeng2022slowfast} (ID = P-109) to leverage 2D pose body-joint graph for diagnosing gait disorders such as PD. 
Zhang et al. \cite{zhang2023wm} (ID = P-121) proposed a novel model named WM-STGCN that leveraged a weighted adjacency matrix and multi-scale temporal convolution in a spatio-temporal graph convolution network to classify PD gaits from normal gaits and achieved up to 87.1\% accuracy on 2D skeleton data. This method outperformed many existing methods including LSTM, KNN, DT, AdaBoost, and spatial temporal graph convolutional networks (ST-GCN) model by Yan et al. \cite{yan2018spatial}.
Dong et al.\cite{dong2023static} (ID = P-123) devised a static-dynamic temporal networks that consisted of one-dimensional, two-dimensional convolutional networks, and attention mechanisms to classify PD patients based on foot force (vGRF time series) and achieved a 96.7\% accuracy.
\hc{The necessity for objective and efficient gait assessment in PD using the Movement Disorder Society Unified Parkinson’s Disease Rating Scale~(MDS-UPDRS) is proposed in~\cite{tian2024cross} (ID = P-126, $S_{Q}=1$).To overcome the limitations of existing PD gait skeleton datasets, they collect a sizable dataset including 102 PD patients, 30 healthy older adults, and 16 young adults using multi-view Azure Kinect sensors. In addition, this paper develops a cross-spatiotemporal graph convolution network (CST-GCN) that captures complex gait features by modeling cross-spacetime dependencies between joints to assess age-related effects on PD gait.}

\begin{table*}[]
\centering
\caption{Existing studies (P-1 to P-126) of AI-assisted PD diagnosis based on gait. We summarize input data, main AI model(s), experimental sample size, and accuracy (range) of each study. Quality scores ($S_{Q}$) that evaluate advancement of AI models, comprehensiveness of comparison, and sufficiency of samples are computed (see Sec. \ref{quality_criterion}).}
\label{PD_1}
\scalebox{0.565}{
\renewcommand\arraystretch{1.1}{
\setlength{\tabcolsep}{1.75mm}{
\hc{
\begin{tabular}{cclccllccc}
\hline
\textbf{ID} & \textbf{Study} & \multicolumn{1}{c}{\textbf{Year}} & \textbf{NDs Types} & \textbf{Data Type} & \textbf{Device/Data} & \textbf{AI Model} & \textbf{Sample Size} & \textbf{Accuracy (\%)} & \textbf{Quality ($S_{Q}$)} \\ \hline
P-1 & \cite{tahir2012parkinson} & 2012 & PD & Senosr+Vision & \begin{tabular}[c]{@{}l@{}}Infrared camera+Replective markers\\ +Force sensitive resistor\end{tabular} & NN, SVM & 32 & 98.2 & 0.45 \\ \hline
P-2 & \cite{kugler2013automatic} & 2013 & PD & Sensor & EMG & SVM & 10 & 90.0 & 0.33 \\ \hline
P-3 & \cite{klucken2013unbiased} & 2013 & PD & Sensor & IMUs & AdaBoost & 173 & 81.0 & 0.56 \\ \hline
P-4 & \cite{manap2013parkinsonian} & 2013 & PD & Senosr+Vision & \begin{tabular}[c]{@{}l@{}}Infrared camera+Replective markers\\ +Force sensitive resistor\end{tabular} & DT & 32 & 100.0 & 0.45 \\ \hline
P-5 & \cite{zhang2013pathological} & 2013 & PD & Sensor & vGRF & SVM & 62 & 81.5-83.4 & 0.56 \\ \hline
P-6 & \cite{tripoliti2013automatic} & 2013 & PD & Sensor & IMUs & RF, DT, NB & 16 & 96.1 & 0.33 \\ \hline
P-7 & \cite{khorasani2014hmm} & 2014 & PD & Sensor & Force sensitive resistor & LS-SVM, HMM & 31 & 90.3 & 0.67 \\ \hline
P-8 & \cite{arora2014high} & 2014 & PD & Sensor & IMUs & RF & 20 & 98.0 & 0.33 \\ \hline
P-9 & \cite{wahid2015classification} & 2015 & PD & Sensor & vGRF & KFD, BA, KNN, SVM, RF & 49 & 82.0-92.6, & 0.44 \\ \hline
P-10 & \cite{tupa2015motion} & 2015 & PD & Vision & RGB Camera & NN & 51 & 97.2 & 0.67 \\ \hline
P-11 & \cite{alkhatib2015gait} & 2015 & PD & Sensor & vGRF & KNN & 166 & 83.0 & 0.56 \\ \hline
P-12 & \cite{prochazka2015bayesian} & 2015 & PD & Senosr+Vision & Kinect (Camera+Depth sensor) & \begin{tabular}[c]{@{}l@{}}Bayesian probability \\ classification\end{tabular} & 51 & 94.1 & 0.44 \\ \hline
P-13 & \cite{perumal2016gait} & 2016 & PD & Sensor & IMUs+Force sensitive resistor & LDA, SVM, ANN & 166 & 83.3-90.0 & 0.56 \\ \hline
P-14 & \cite{zeng2016parkinson} & 2016 & PD & Sensor & vGRF & \begin{tabular}[c]{@{}l@{}}Deterministic learning, \\ RBF-NN\end{tabular} & 166 & 91.6-99.4 & 0.78 \\ \hline
P-15 & \cite{wang2016recognizing} & 2016 & PD & Sensor & Force sensitive resistor & Neural networks & 165 & 87.9 & 1.00 \\ \hline
P-16 & \cite{ertuugrul2016detection} & 2016 & PD & Sensor & vGRF & \begin{tabular}[c]{@{}l@{}}BN, NB, MLP, \\ LR, RF, FT, etc\end{tabular} & 166 & 87.6-88.9 & 0.78 \\ \hline
P-17 & \cite{jane2016q} & 2016 & PD & Sensor & Force sensitive resistor & Q-BTDNN & 166 & 90.9-92.2 & 1.00 \\ \hline
P-18 & \cite{cuzzolin2017metric} & 2017 & PD & Sensor & IMUs & KNN & 580 & 80.8-90.2 & 0.56 \\ \hline
P-19 & \cite{abujrida2017smartphone} & 2017 & PD & Sensor & IMUs (Smartphone) & RF & 50 & 87.0 & 0.44 \\ \hline
P-20 & \cite{alam2017vertical} & 2017 & PD & Sensor & IMUs+vGRF & SVM, KNN, RF, DT & 47 & 93.6 & 0.67 \\ \hline
P-21 & \cite{joshi2017automatic} & 2017 & PD & Sensor & vGRF & SVM & 31 & 90.3-100 & 0.67 \\ \hline
P-22 & \cite{djuric2017selection} & 2017 & PD & Sensor & Walkway system (GAITRite) & RF, RBF-SVM & 80 & 85.0 & 0.44 \\ \hline
P-23 & \cite{pham2017tensor} & 2017 & PD & Sensor & vGRF & Tensor decomposition & 165 & 100.0 & 0.56 \\ \hline
P-24 & \cite{wu2017measuring} & 2017 & PD & Sensor & vGRF & GLRA, SVM & 58 & 82.8-84.5 & 0.44 \\ \hline
P-25 & \cite{fernandes2018artificial} & 2018 & PD & Sensor & IMUs & MLP, DBNs & 45 & 93.5-94.5 & 0.56 \\ \hline
P-26 & \cite{urcuqui2018exploring} & 2018 & PD & Sensor & Kinect & LR, NB, RF, DT & 60 & 61.0-82.0 & 0.44 \\ \hline
P-27 & \cite{abdulhay2018gait} & 2018 & PD & Sensor & vGRF & SVM & 166 & 94.8 & 0.56 \\ \hline
P-28 & \cite{caramia2018imu} & 2018 & PD & Sensor & IMUs & NB, LDA, k-NN, DT, SVM & 50 & 68.9-75.6 & 0.44 \\ \hline
P-29 & \cite{mileti2018measuring} & 2018 & PD & Sensor & IMUs+Force sensitive resistor & \begin{tabular}[c]{@{}l@{}}Continuous Hidden  \\ Markov Model (cHMM)\end{tabular} & 26 & 0.98 (AUC) & 0.44 \\ \hline
P-30 & \cite{rovini2018comparative} & 2018 & PD & Sensor & IMUs & SVM, RF, NB & 90 & 78.0-96.7 & 0.67 \\ \hline
P-31 & \cite{pham2017tensor} & 2018 & PD & Sensor & vGRF & Tensor model & 165 & 86.2 & 0.78 \\ \hline
P-32 & \cite{aich2018validation} & 2018 & PD & Sensor & IMUs & SVM, KNN, DT, NB & 51 & 89.1 & 0.67 \\ \hline
P-33 & \cite{mittra2018classification} & 2018 & PD & Sensor & vGRF & LR, DT, RF, SVM, KNN & 166 & 93.1 & 0.56 \\ \hline
\end{tabular}
}
}
}
}
\end{table*}

\begin{table*}[]
\centering
\caption{Continued table of Table \ref{PD_1} for AI-assisted PD diagnosis based on gait.}
\label{PD_2}
\scalebox{0.565}{
\renewcommand\arraystretch{1.1}{
\setlength{\tabcolsep}{1.9mm}{
\hc{
\begin{tabular}{lclccllccc}
\hline
\multicolumn{1}{c}{\textbf{ID}} & \textbf{Study} & \multicolumn{1}{c}{\textbf{Year}} & \textbf{NDs Types} & \textbf{Data Type} & \textbf{Device/Data} & \textbf{AI Model} & \textbf{Sample Size} & \textbf{Accuracy (\%)} & \textbf{Quality ($S_{Q}$)} \\ \hline
P-34 & \cite{chen2018detecting} & 2018 & PD & Sensor & vGRF & SVM, PSO & 135 & 87.1-95.7 & 0.56 \\ \hline
P-35 & \cite{aich2018machine} & 2018 & PD & Vision & Camera+Reflective markers & \begin{tabular}[c]{@{}l@{}}Minimum redundancy \\ maximum relevance \\ (MRMR), PCA\end{tabular} & 40 & 98.5 & 0.44 \\ \hline
P-36 & \cite{krajushkina2018gait} & 2018 & PD & Sensor & Kinect & Fisher's Score (FS) & 40 & 75.0-85.0 & 0.33 \\ \hline
P-37 & \cite{acsurouglu2018parkinson} & 2018 & PD & Sensor & vGRF & \begin{tabular}[c]{@{}l@{}}Locally Weighting  Random \\ Forest (LWRF)\end{tabular} & 165 & 99.0 & 0.89 \\ \hline
P-38 & \cite{elden2018computer} & 2018 & PD & Sensor & vGRF & RBF-SVM & 64 & 90.6 & 0.67 \\ \hline
P-39 & \cite{zhao2018hybrid} & 2018 & PD & Sensor & vGRF & LSTM, CNN & 166 & 93.9-98.6 & 0.89 \\ \hline
P-40 & \cite{buongiorno2019low} & 2019 & PD & Vision & Camera (Kinect) & SVM, ANN & 30 & 89.4 & 0.44 \\ \hline
\\ \hline
P-41 & \cite{rehman2019comparison} & 2019 & PD & Sensor & IMUs & SVM, RF & 196 & 0.805-0.878 (AUC) & 0.56 \\ \hline
P-42 & \cite{khoury2019data} & 2019 & PD & Sensor & vGRF & \begin{tabular}[c]{@{}l@{}}KNN, DT, RF, NB, \\ SVM,  K-Means, GMM\end{tabular} & 165 & 69.0-90.0 & 0.78 \\ \hline
P-43 & \cite{reyes2019lstm} & 2019 & PD & Vision & Camera (Kinect) & Conv LSTM & 182 & 64.0-83.0 & 0.67 \\ \hline
P-44 & \cite{DBLP:conf/hicss/RastegariAA19} & 2019 & PD & Sensor & IMUs & \begin{tabular}[c]{@{}l@{}}SVM, RF, AdaBoost, \\ Bagging, NB\end{tabular} & 30 & 83.6-96.7 & 0.56 \\ \hline
P-45 & \cite{guo2019method} & 2019 & PD & Sensor & EEG & RF, SVM, KNN & 41 & 85.7 & 0.56 \\ \hline
P-46 & \cite{vasquez2018multimodal} & 2019 & PD & Sensor & IMUs & CNN & 84 & 97.6 & 0.56 \\ \hline
\multicolumn{1}{c}{P-47} & \cite{andrei2019parkinso} & 2019 & PD & Sensor & Force sensitive resistor & SVM & 166 & 88.9-100.0 & 0.78 \\ \hline
P-48 & \cite{rehman2019selecting} & 2019 & PD & Sensor & Walkway system (GAITRite) & RF, SVM, LR & 303 & 97.0 & 0.56 \\ \hline
P-49 & \cite{ricciardi2019using} & 2019 & PD & Sensor+Vision & \begin{tabular}[c]{@{}l@{}}Optoelectronic cameras+\\ Dynamometric platform\end{tabular} & RF, GBM & 46, & 86.4 & 0.56 \\ \hline
P-50 & \cite{zeng2019classification} & 2019 & PD & Sensor & vGRF & NN & 166 & 98.8 & 0.78 \\ \hline
P-51 & \cite{alharbi2020genetic} & 2020 & PD & Sensor & Force sensitive resistor & \begin{tabular}[c]{@{}l@{}}Extreme learning machine\\ (ELM) neural network\end{tabular} & 31 & 93.5 & 0.56 \\ \hline
P-52 & \cite{el2020deep} & 2020 & PD & Sensor & vGRF & DNN & 165 & 98.7 & 0.89 \\ \hline
P-53 & \cite{perez2020identification} & 2020 & PD & Sensor & IMUs+Force sensitive resistor & SVM & 10 & 97.4-98.8 & 0.44 \\ \hline
P-54 & \cite{alkhatib2020machine} & 2020 & PD & Sensor & vGRF & Discriminant analysis & 47 & 95.0 & 0.56 \\ \hline
P-55 & \cite{abujrida2020machine} & 2020 & PD & Sensor & IMUs (Smart Phone) & RF & 456 & 85.5-95.0 & 0.56 \\ \hline
P-56 & \cite{veeraragavan2020parkinson} & 2020 & PD & Sensor & vGRF & NN & 166 & 97.4 & 0.89 \\ \hline
P-57 & \cite{hughes2019models} & 2020 & PD & Sensor & Force sensitive resistor & \begin{tabular}[c]{@{}l@{}}XGBoost, Artificial \\ Neural  Network (ANN), \\ Symbolic Regression (SR)\end{tabular} & 64 & 41.0 & 0.78 \\ \hline
P-58 & \cite{xia2019dual} & 2020 & PD & Sensor & vGRF & CNN, LSTM & 166 & 98.0 & 0.89 \\ \hline
P-59 & \cite{perez2020identification} & 2020 & PD & Sensor & IMUs+Force sensitive resistor & \begin{tabular}[c]{@{}l@{}}Adaptive Unsupervised  \\ Learning\end{tabular} & 12 & 98.3 & 0.33 \\ \hline
P-60 & \cite{som2020unsupervised} & 2020 & PD & Sensor & IMUs & MLP, SVM & 34 & 68.9 & 0.67 \\ \hline
P-61 & \cite{rehman2020turning} & 2020 & PD & Sensor & IMUs & PLS-DA & 93 & 97.9 & 0.56 \\ \hline
P-62 & \cite{nair2020predicting} & 2020 & PD & Sensor & IMUs & LR & 16 & 93.6 & 0.44 \\ \hline
P-63 & \cite{moon2020classification} & 2020 & PD & Sensor & vGRF & NN, SVM, KNN, DT, RF & 567 & 49.0-61.0 (F1 score) & 0.67 \\ \hline
P-64 & \cite{gong2020novel} & 2020 & PD & Vision & Camera & Mask R-CNN & 301 & 97.3 & 0.67 \\ \hline
\end{tabular}
}
}
}
}
\end{table*}

\begin{table*}[]
\centering
\caption{Continued table of Table \ref{PD_2} for AI-assisted PD diagnosis based on gait.}
\label{PD_3}
\scalebox{0.565}{
\renewcommand\arraystretch{1.1}{
\setlength{\tabcolsep}{2.5mm}{
\hc{
\begin{tabular}{cclccllccc}
\hline
\textbf{ID} & \textbf{Study} & \multicolumn{1}{c}{\textbf{Year}} & \textbf{NDs Types} & \textbf{Data Type} & \textbf{Device/Data} & \textbf{AI Model} & \textbf{Sample Size} & \textbf{Accuracy (\%)} & \textbf{Quality ($S_{Q}$)} 
\\ \hline
P-65 & \cite{balaji2020supervised} & 2020 & PD & Sensor & vGRF & SVM, DT & 166 & 69.9-99.4 & 0.78 \\ \hline
P-66 & \cite{butt2020biomechanical} & 2020 & PD & Sensor & IMUs & Bidirectional-LSTM & 114 & 82.4 & 0.67 \\ \hline
P-67 & \cite{zhang2020deep} & 2020 & PD & Sensor & IMUs & CNN & 2804 & 0.86 (AUC) & 0.67 \\ \hline
P-68 & \cite{yurdakul2020detection} & 2020 & PD & Sensor & vGRF & ANN & 166 & 98.3 & 0.89 \\ \hline
P-69 & \cite{balaji2021automatic} & 2021 & PD & Sensor & vGRF & LSTM & 64 & 98.6 & 0.89 \\ \hline
P-70 & \cite{pham2021classification} & 2021 & PD & Sensor & vGRF & Bi-LSTM & 116 & 91.2-100.0 & 0.89 \\ \hline
P-71 & \cite{balaji2021data} & 2021 & PD & Sensor & vGRF & KNN, NB, SVM & 165 & 88.5-98.8 & 0.78 \\ \hline
P-72 & \cite{williamson2021detecting} & 2021 & PD & Sensor & IMUs & GMM & 380 & 0.69 (AUC) & 0.56 \\ \hline
P-73 & \cite{mirelman2021detecting} & 2021 & PD & Sensor & IMUs & \begin{tabular}[c]{@{}l@{}}Random  under-sampling \\ boosting (RUSBoost) \\ classification\end{tabular} & 432 & 0.76-0.90 (AUC) & 0.56 \\ \hline
P-74 & \cite{ma2023explainable} & 2021 & PD & Sensor & Force sensitive resistor & CNN, XGBoost & 166 & 97.3-98.4 & 0.78 \\ \hline
P-75 & \cite{alharthi2020gait} & 2021 & PD & Sensor & vGRF & CNN & 166 & 92.7-98.3 & 1.00 \\ \hline
P-76 & \cite{yuan2021accelerating} & 2021 & PD & Others & Electronic medical record & LR, RNN & 28216 & 0.874 (AUC) & 0.78 \\ \hline
P-77 & \cite{park2021classification} & 2021 & PD & Vision & \begin{tabular}[c]{@{}l@{}}Infrared camera+\\ Reflective marker\end{tabular} & \begin{tabular}[c]{@{}l@{}}LR, KNN, NB, \\ LDA, SVM, RF\end{tabular} & 111 & 98.1 & 0.56 \\ \hline
P-78 & \cite{filtjens2021modelling} & 2021 & PD & Vision & \begin{tabular}[c]{@{}l@{}}Infrared camera+\\ Reflective marker\end{tabular} & CNN & 42 & 98.7 & 0.67 \\ \hline
P-79 & \cite{shalin2021prediction} & 2021 & PD & Sensor & vGRF & LSTM & 11 & 84.5 & 0.56 \\ \hline
P-80 & \cite{guayacan2021visualising} & 2021 & PD & Vision & Camera & CNN & 22 & 94.9 & 0.33
\\ \hline
P-81 & \cite{ghaderyan2021inter} & 2021 & PD & Sensor & vGRF & \begin{tabular}[c]{@{}l@{}}Sparse  non-negative \\ least-squares (NNLS)\end{tabular} & 166 & 97.2 & 0.67 \\ \hline
P-82 & \cite{nguyen2021classification} & 2021 & PD & Sensor & vGRF & NN & 166 & 84.8 & 1.00 \\ \hline
P-83 & \cite{zheng2021svm} & 2021 & PD & Sensor & IMUs & SVM & 300 & 96.7 & 0.56 \\ \hline
P-84 & \cite{ahamed2021spatiotemporal} & 2021 & PD & Sensor & vGRF & CNN & 166 & 97.0 & 0.67 \\ \hline
P-85 & \cite{salimi2023type} & 2022 & PD & Sensor & vGRF & KNN, Fuzzy neural network & 165 & 97.6 & 1.00 \\ \hline
P-86 & \cite{carvajal2022classification} & 2022 & PD & Sensor & IMUs & CNN, GRU & 134 & 83.7-92.7 & 0.67 \\ \hline
P-87 & \cite{mehra2022design} & 2022 & PD & Sensor & vGRF & RF & 165 & 94.5-98.8 & 0.78 \\ \hline
P-88 & \cite{li2022detecting} & 2022 & PD & Sensor & vGRF & LR, SVM, DT, KNN & 306 & 81.0-85.0 & 0.56 \\ \hline
P-89 & \cite{ogata2022foot} & 2022 & PD & Sensor & IMUs & SVM & 115 & 92.7-96.3 & 0.56 \\ \hline
P-90 & \cite{rehman2022investigating} & 2022 & PD & Sensor & IMUs & SVM, RF & 99 & 64.1-72.7 & 0.56 \\ \hline
P-91 & \cite{trabassi2022machine} & 2022 & PD & Sensor & IMUs & SVM, DT, RF, KNN, MLP & 128 & 74.0-86.0 & 0.56 \\ \hline
P-92 & \cite{munoz2022machine} & 2022 & PD & Vision & Camera (Kinect) & LR, SVM, DT, NB, RF & 60 & 81.8-84.5 & 0.56 \\ \hline
P-93 & \cite{yang2022pd} & 2022 & PD & Sensor & IMUs & PD-ResNet & 457 & 95.5 & 1.00 \\ \hline
P-94 & \cite{guayacan2022quantification} & 2022 & PD & Sensor & Force sensitive resistor & SVM, RF & 22 & 99.6 & 0.44 \\ \hline
P-95 & \cite{goni2022smartphone} & 2022 & PD & Sensor & IMUs (Smart phone) & RF & 1397 & 73.0 & 0.56 \\ \hline
P-96 & \cite{doi:10.1080/20476965.2022.2125838} & 2022 & PD & Vision & Camera & \begin{tabular}[c]{@{}l@{}}Feature-weighted  Minimum \\ Distance Classifier Model\end{tabular} & 28 & 93.6 & 0.67 \\ \hline
P-97 & \cite{acsurouglu2022deep} & 2022 & PD & Sensor & vGRF & CNN, RF & 166 & 99.5 & 1.00 \\ \hline
P-98 & \cite{olmos2022riemannian} & 2022 & PD & Vision & Camera & \begin{tabular}[c]{@{}l@{}}Symmetric positive  definite \\ (SPD) Riemannian network\end{tabular} & 22 & 99.4 & 0.56 \\ \hline

\end{tabular}
}
}
}
}
\end{table*}

\begin{table*}[]
\centering
\caption{Continued table of Table \ref{PD_3} for AI-assisted PD diagnosis based on gait.}
\label{PD_4}
\scalebox{0.565}{
\renewcommand\arraystretch{1.1}{
\setlength{\tabcolsep}{1.25mm}{
\hc{
\begin{tabular}{cclccllccc}
\hline
\textbf{ID} & \textbf{Study}                           & \multicolumn{1}{c}{\textbf{Year}} & \textbf{NDs Types} & \textbf{Data Type} & \textbf{Device/Data}                                                    & \textbf{AI Model}                                                                                                                 & \textbf{Sample Size} & \textbf{Accuracy (\%)} & \textbf{Quality ($S_{Q}$)} \\ \hline
P-99 & \cite{ingelse2022personalised} & 2022 & PD & Sensor & IMUs & CNN,   NN & 17 & 91.0-92.0 & 0.44 \\ \hline
P-100 & \cite{filtjens2022automated} & 2022 & PD & Sensor & Retro-reflective markers & \begin{tabular}[c]{@{}l@{}}Spatial-temporal  graph convolutional\\  network (MS-GCN)\end{tabular} & 42 & 83.0 & 0.56 \\ \hline
P-101 & \cite{carvajal2022effective} & 2022 & PD & Sensor & IMUs & XGBoost, SVM, RF & 134 & 81.0-88.3 & 0.56 \\ \hline
P-102 & \cite{chatzaki2022can} & 2022 & PD & Sensor+Vision & Camera+Force sensitive resistor & AdaBoost, Extra Trees, RF & 44 & 88.0 & 0.56 \\ \hline
P-103 & \cite{ferreira2022machine} & 2022 & PD & Sensor+Vision & Camera+Force sensitive resistor & \begin{tabular}[c]{@{}l@{}}NB, SVM, DT, RF,\\  LR, MLP\end{tabular} & 126 & 76.9-84.6 & 0.56 \\ \hline
P-104 & \cite{brand2022gait} & 2022 & PD & Sensor & IMUs & DCNN, LSTM & 30 & 83.3 & 0.78 \\ \hline
P-105 & \cite{atri2022deep} & 2022 & PD & Sensor & IMUs (Smart phone) & \begin{tabular}[c]{@{}l@{}}CNN, LR, RF, \\ Gradient-Boosted Trees\end{tabular} & 11 & 99.0-100.0 & 0.33 \\ \hline
P-106 & \cite{zhao2021multimodal} & 2022 & PD & Sensor+Vision & Camera (Kinect)+vGRF+IMUs & \begin{tabular}[c]{@{}l@{}}Correlative Memory \\ Neural Network (CorrMNN)\end{tabular} & 93 & 99.5-100.0 & 1.00 \\ \hline
P-107 & \cite{mendoza2022local} & 2022 & PD & Vision & Camera & CNN, SVM & 22 & 96.6 & 0.33 \\ \hline
P-108 & \cite{nino2023parkinsonian} & 2022 & PD & Vision & Camera & \begin{tabular}[c]{@{}l@{}}Gaussian Naive Bayes, \\ LR, RF, SVM\end{tabular} & 22 & 99.0 & 0.44 \\ \hline
P-109 & \cite{zeng2022slowfast} & 2022 & PD & Vision & Camera (Smart phone) & SlowFast GCN network & 68 & 77.1-87.5 & 0.89 \\ \hline
P-110 & \cite{li2022abnormal} & 2022 & PD & Sensor & Force sensitive resistor+IMUs & RF, LR, Gradient Boosting & 22 & 97.3 & 0.44 \\ \hline
P-111 & \cite{wang2022gait} & 2022 & PD & Sensor & vGRF & SVM & 166 & 98.2 & 0.78 \\ \hline
P-112 & \cite{goh2022gait} & 2022 & PD & Sensor & vGRF & SVM, ANN & 48318 & 93.0 & 0.56 \\ \hline
P-113 & \cite{kour2022vision} & 2022 & PD & Sensor+Vision & \begin{tabular}[c]{@{}l@{}}Camera (DSLR)+\\ Passive marker\end{tabular} & KNN & 96 & 90.1 & 0.56 \\ \hline
P-114 & \cite{beigi2022classification} & 2022 & PD & Sensor & IMUs & MLP, KNN, RF, SVM & 30 & 100.0 & 0.56 \\ \hline
P-115 & \cite{baby2017parkinsons} & 2022 & PD & Sensor & vGRF & ANN & 166 & 64.4-86.8 & 0.67 \\ \hline
P-116 & \cite{divyashree2022ai} & 2023 & PD & Sensor & vGRF & 2D CNN & 166 & 98.0 & 0.89 \\ \hline
P-117 & \cite{beigi2023freezing} & 2023 & PD & Sensor & IMUs & \begin{tabular}[c]{@{}l@{}}KNN, DT, SVM, \\ ANN, RF\end{tabular} & 30 & 94.0-96.0 & 0.56 \\ \hline
P-118 & \cite{uchitomi2023classification} & 2023 & PD & Sensor & IMUs & CNN & 90 & 82.8-86.4 & 0.56 \\ \hline
P-119 & \cite{cai2023specific} & 2023 & PD & Sensor & IMUs & PCA, SVM & 280 & 88.0-90.0 & 0.56 \\ \hline
P-120 & \cite{sun2023novel} & 2023 & PD & Sensor & Force sensitive resistor & \begin{tabular}[c]{@{}l@{}}Radial basis function \\ neural network (RBFNN)\end{tabular} & 216 & 87.6-100.0 & 0.89 \\ \hline
P-121 & \cite{zhang2023wm} & 2023 & PD & Vision & Camera (Smart phone) & \begin{tabular}[c]{@{}l@{}}Weighted adjacency matrix based \\ Spatiotemporal Graph \\ Convolution Network (WM-STGCN)\end{tabular} & 50 & 87.1 & 0.89 \\ \hline
P-122 & \cite{dotov2023coordination} & 2023 & PD & Sensor & IMUs & SVM & 78 & 82.0 & 0.56 \\ \hline
P-123 & \cite{dong2023static} & 2023 & PD & Sensor & IMUs & \begin{tabular}[c]{@{}l@{}}Static-Dynamic temporal \\ networks\end{tabular} & 166 & 96.7 & 1.00 \\ \hline
P-124 & \cite{han2023automatic} & 2023 & PD & Sensor & IMUs & Non-linear model & 53 & 84.9 & 0.56 \\ \hline
P-125 & \cite{johnson2024wearable} & 2024 & PD & Sensor & Wearable devices & Random forest & 132 & 92.3 & 0.88 \\ \hline
P-126 & \cite{tian2024cross} & 2023 & PD & Sensor & Kinect & \begin{tabular}[c]{@{}l@{}} Cross-spatiotemporal graph\\ convolution network (CST-GCN)\end{tabular} & 148 & 84.9 & 1.00
 \\ \hline
\end{tabular}
}
}
}
}
\end{table*}

\FloatBarrier

\begin{table*}[t]
\centering
\caption{Existing studies (C-1 to C-21) that include combinatorial diagnosis of different NDs (PD, ALS, HD) based on gait. We summarize input data, main AI model(s), experimental sample size, accuracy (range), and quality score of each study.
Quality scores that evaluate the novelty of AI models, comprehensiveness of method comparison, and sufficiency of experimental samples are computed (see Sec. \ref{quality_criterion}).}
\label{combine_table}
\scalebox{0.565}{
\renewcommand\arraystretch{1.1}{
\setlength{\tabcolsep}{1mm}{
\hc{
\begin{tabular}{ccllcllccc}
\hline
\textbf{ID} & \textbf{Study} & \multicolumn{1}{c}{\textbf{Year}} & \multicolumn{1}{c}{\textbf{NDs Types}} & \textbf{Data Type} & \textbf{Device/Data} & \textbf{AI Model} & \textbf{Sample Size} & \textbf{Accuracy (\%)} & \textbf{Quality ($S_{Q}$)} \\ \hline
C-1 & \cite{daliri2012automatic} & 2012 & PD, ALS, HD & Sensor & Force  sensitive resistor & SVM & 64 & 90.6 & 0.56 \\ \hline
C-2 & \cite{dutta2013hybrid} & 2013 & PD, ALS, HD & Sensor & vGRF & Elman’s recurrent neural network (ERNN) & 63 & 90.6 & 0.78 \\ \hline
C-3 & \cite{xia2015classification} & 2015 & PD, ALS, HD & Sensor & Force sensitive resistor & SVM, RF, MLP, KNN & 64 & 96.6-100.0 & 0.56 \\ \hline
C-4 & \cite{pratiher2017towards} & 2017 & PD, ALS, HD & Sensor & vGRF & SVM, EC, DT, KNN & 62 & 79.3 & 0.67 \\ \hline
C-5 & \cite{najafabadian2018neurodegenerative} & 2017 & PD, ALS, HD & Sensor & vGRF & Adaboost & 64 & 87.3-92.3 & 0.78 \\ \hline
C-6 & \cite{athisakthi2017statistical} & 2018 & PD, ALS, HD & Sensor & vGRF & RF, SVM & 64 & 90.6-96.9 & 0.56 \\ \hline
C-7 & \cite{klomsae2018string} & 2018 & PD, ALS, HD & Sensor & Force sensitive resistor & fusszy KNN & 64 & 98.4 & 0.56 \\ \hline
C-8 & \cite{yan2020classification} & 2020 & PD, ALS, HD & Sensor & Force sensitive resistor & Random forest & 64 & 79.3-91.7 & 0.78 \\ \hline
C-9 & \cite{begum2020recognition} & 2020 & PD, ALS, HD & Vision & Camera & Discriminant  Analysis, RF, Multi-SVM, KNN & 64 & 92.2 & 0.56 \\ \hline
C-10 & \cite{nam2020development} & 2020 & PD, ALS, HD & Sensor & Force sensitive resistor & SVM, KNN & 64 & 99.8-100.0 & 0.56 \\ \hline
C-11 & \cite{lin2020evaluation} & 2020 & PD, ALS, HD & Sensor & vGRF & CNN & 64 & 98.4-100.0 & 0.67 \\ \hline
C-12 & \cite{beyrami2020robust} & 2020 & PD, ALS, HD & Sensor & vGRF & \begin{tabular}[c]{@{}l@{}}Sparse non-negative least squares (NNLS), SVM,\\ multilayer feed forward neural network (MLFN)\end{tabular} & 64 & 99.6-100.0 & 0.89 \\ \hline
C-13 & \cite{yilmaz2020evaluation} & 2020 & PD, ALS, HD & Sensor & Force sensitive resistor & KNN & 64 & 90.5 & 0.56 \\ \hline
C-14 & \cite{setiawan2021identification} & 2021 & PD, ALS, HD & Sensor & vGRF & CNN, SVM & 64 & 97.4-100 & 0.78 \\ \hline
C-15 & \cite{zhao2021multimodal} & 2022 & PD, ALS, HD & Sensor+Vision & \begin{tabular}[c]{@{}l@{}}Camera+IMUs\\ +Force sensitive resistor\end{tabular} & Correlative memory neural network (CorrMNN) & 64 & 99.7-100.0 & 0.89 \\ \hline
C-16 & \cite{setiawan2022development} & 2022 & PD, ALS, HD & Sensor & vGRF & AlextNet CNN & 63 & 96.4 & 0.78 \\ \hline
C-17 & \cite{tobar2022petri} & 2022 & PD, ALS, HD & Sensor & vGRF & RF,  classification trees & 64 & 69.6-95.1 & 0.78 \\ \hline
C-18 & \cite{torres2023using} & 2023 & PD, ALS, HD & Sensor & Force sensitive resistor & \begin{tabular}[c]{@{}l@{}}Learning Algorithm for Multivariate \\ Data Aalysis (LAMDA), Neural networks.\end{tabular} & 60 & 98.3 & 0.56 \\ \hline
C-19 & \cite{zhou2023deep} & 2023 & PD, ALS, HD & Sensor & Force sensitive resistor & LSTM,  GRU, ResNet, FCN, etc & 64 & 75.4-83.3 & 0.56 \\ \hline
C-20 & \cite{erdacs2024cnn} & 2024 & PD, ALS, HD & Sensor & Force sensitive resistor & CNN & 64 & 93.56-97.65 & 0.77 \\ \hline
C-21 & \cite{li2023multimodal} & 2024 & PD, ALS, HD & Sensor & Force sensitive resistor & CNN-BiLSTM  & 64 & 95.43-1 & 1.00 \\ \hline
\end{tabular}
}
}
}
}
\end{table*}

\begin{table*}[t]
\centering
\caption{Existing studies of AI-assisted diagnosis of AD (A-1 to A-6), ALS (S-1 to S-3), HD (H-1 to H-7) or MS (M-1 to M-6) based on gait.  We summarize input data, main AI model(s), experimental sample size, accuracy (range), and quality score of each study.
Quality scores that evaluate the novelty of AI models, comprehensiveness of method comparison, and sufficiency of experimental samples are computed (see Sec. \ref{quality_criterion}).}
\label{AD_ALS_HD_MS_table}
\scalebox{0.575}{
\renewcommand\arraystretch{1.1}{
\setlength{\tabcolsep}{1mm}{
\hc{
\begin{tabular}{cllccllccc}
\hline
\textbf{ID} & \textbf{Study} & \multicolumn{1}{c}{\textbf{Year}} & \textbf{NDs Type} & \textbf{Data Type} & \textbf{Device/Data} & \textbf{AI Model} & \textbf{Sample Size} & \textbf{Accuracy (\%)} & \textbf{Quality ($S_{Q}$)} \\ \hline
A-1 & \cite{costa2016application} & \multicolumn{1}{c}{2016} & AD & Sensor & IMUs & SVM,  MLP, RBNs, DBSs & 72 & 91.0-96.6 & 0.67 \\ \hline
A-2 & \cite{you2020alzheimer} & \multicolumn{1}{c}{2020} & AD & Sensor+Vision & Camera+EEG & SVM,  RF, CNN, ST-GCN, etc & 87 & 91.1-98.6 & 0.89 \\ \hline
A-3 & \cite{ghoraani2021detection} & \multicolumn{1}{c}{2020} & AD & Sensor & Pressure sentsitive mat & SVM & 78 & 78.0 & 0.56 \\ \hline
A-4 & \cite{seifallahi2022alzheimer} & \multicolumn{1}{c}{2022} & AD & Sensor+Vision & Kinect depth sensor + Camera & SVM & 85 & 97.8 & 0.56 \\ \hline
A-5 & \cite{jeon2023early} & \multicolumn{1}{c}{2023} & AD & Sensor & IMUs & SVM, KNN, RF, MLP, etc & 145 & 72.2 & 0.67 \\ \hline
A-6 & \cite{bringas2024cladsi} & \multicolumn{1}{c}{2024} & AD & Sensor & Accelerometer Sensor (Smartphone) & CLADSI & 35 & 84-87 & 0.77 \\ \hline
S-1 & \cite{xia2016novel} & 2016 & ALS & Sensor & Force sensitive resistor & SVM & 28 & 92.9 & 0.44 \\ \hline
S-2 & \cite{felix2020effective} & 2020 & ALS & Sensor & Force sensitive resistor & SVM, KNN. DT & 29 & 86.2-96.6 & 0.44 \\ \hline
S-3 & \cite{felix2021automatic} & 2021 & ALS & Sensor & Force sensitive resistor & SVM, KNN, NB, LDA, etc & 29 & 89.7 & 0.67 \\ \hline
H-1 & \cite{mannini2016machine} & 2016 & HD & Sensor & IMUs & HMM, SVM & 42 & 88.2 & 0.56 \\ \hline
H-2 & \cite{acosta2018meta} & 2018 & HD & Sensor & IMUs & DT, RF, Logitboost, Multiboost & 14 & 78.6-92.9 & 0.67 \\ \hline
H-3 & \cite{zhang2019deep} & 2019 & HD & Sensor & Walkway system & SVM, 3D CNN & 12 & 82.0-86.9 & 0.44 \\ \hline
H-4 & \cite{felix2019automatic} & 2019 & HD & Sensor & force sensitive resistor & SVM, KNN, NB, LDA, DT & 36 & 58.3-94.4 & 0.56 \\ \hline
H-5 & \cite{paula2019diagnosing} & 2019 & HD & Sensor & force sensitive resistor & SVM, KNN, NB, LDA, DT & 36 & 100.0 & 0.78 \\ \hline
H-6 & \cite{huang2019machine} & 2019 & HD & Sensor & vGRF & SVM, NB, DT, RF, LR & 35 & 85.7-97.1 & 0.56 \\ \hline
H-7 & \cite{trojaniello2015assessment} & 2020 & HD & Sensor & \begin{tabular}[c]{@{}l@{}}Magnetics and inertial \\ measurement units\end{tabular} & SVM & 42 & 90.5 & 0.56 \\ \hline
M-1 & \cite{patil2019clinical} & 2019 & MS & Sensor & Force sensitive sensor & SVM, MLP, KNN, ELM & 945 & 89.8 & 0.67 \\ \hline
M-2 & \cite{kaur2020predicting} & 2021 & MS & Sensor & Treadmill & SVM, RF, AdaBoost, MLP, etc & 40 & 94.3 & 0.56 \\ \hline
M-3 & \cite{trentzsch2021using} & 2021 & MS & Sensor & Force sensitive sensor & SVM, Gaussian Naive Bayes, DT, KNN, etc & 60 & 74.5 & 0.56 \\ \hline
M-4 & \cite{creagh2021interpretable} & 2021 & MS & Sensor & IMUs & TL DCNN HAR models & 97 & 77.6-91.1 & 0.89 \\ \hline
M-5 & \cite{creagh2020smartphone} & 2021 & MS & Sensor & IMUs & SVM, LR, RF & 97 & 79.3-85.1 & 0.56 \\ \hline
M-6 & \cite{hu2022machine} & 2022 & MS & Sensor & Walkway system & SVM, LR, XGBoost & 88 & 77.0-81.0 & 0.56 \\ \hline
\end{tabular}
}
}
}
}
\end{table*}


\subsection{Combinatorial Diagnosis of Different NDs}\label{sec_Combine}
\hc{As shown in Table \ref{combine_table} and \ref{num_types}, 
there are 21 studies that simultaneously focus on diagnosis of PD, ALS, and HD based on human gait, and 11, 7, 3 studies use C-ML, C-DL, A-DL models, respectively. We provide a brief content summary and discussion for representative studies below.}

Daliri et al. \cite{daliri2012automatic} (Table \ref{combine_table}, ID = C-1) used force sensitive sensors to collect stride intervals and footfall contact times. Features were later derived from these time series data and filtered using a genetic algorithm for diagnosing purpose. The best accuracy results for HC $vs.$ HD, HC $vs.$ PD, HC $vs.$ ALS were achieved using the features selected and SVM models with the corresponding rate of 90.3\%, 89.3\%, and 96.8\%, respectively. The accuracy for the diagnosis of three NDs altogether was 90.6\%.
Dutta et al. \cite{dutta2013hybrid} (ID = C-2) used a cross-correlation method to extract features with recurrent neural networks to classify HC and people with NDs, achieving highly competitive performance with a 90.6\% accuracy.
The study by Xia et al. \cite{xia2015classification} (ID = C-3) exploited 9 statistical features (mean, minimum, etc) of gait rhythm signals ($i.e.$, gait cycle parameters) to perform NDs diagnosis using different classic machine learning models (SVM, RF, MLP, KNN), and achieved a 96.6-100\% accuracy. 
Pratiher et al. \cite{pratiher2017towards} (ID = C-4) proposed a novel analytical methodology for distinguishing healthy individuals and NDs patients based on gait patterns. The method employed intrinsic mode functions (IMFs), and extracted discriminative features from the area measure and relative change in centroid position of the polygon formed by the Convex Hull of these analytic IMFs. 
Najafabadian et al. \cite{najafabadian2018neurodegenerative} (ID = C-5) presented a nonlinear gait signal analysis approach to classify NDs by extracting chaotic and fractal features from gait data and reached a classification accuracy of 92.34\%.
Athisakthi \cite{athisakthi2017statistical} (ID = C-6) introduced a technique called Statistical Energy Values and Peak Analysis (SEP) for detecting NDs from signals obtained through force-sensitive resistors. The accuracy, sensitivity, and specificity values were best achieved using the Random Forest Classifier with reported accuracy at 96.9\%, 96.8\%, and 96.7\%, respectively. 
In the work by Klomsae et al. \cite{klomsae2018string} (ID = C-7), the authors transformed left foot stride and stride interval time series data into a string using symbolic aggregate approximation (SAX). The converted string was then utilized to find a string prototype for each disease using the proposed algorithm called sgUPFCMed. Finally, they implemented a fuzzy KNN model to find the best match for a test data sample to classify different diseases and HCs.
Yan et al. \cite{yan2020classification} (ID = C-8) introduced a framework called Topological Motion Analysis (TMA) which embeds gait fluctuation time series into a phase space. In detail, persistent homology was used to extract topological signatures of barcodes, and a Random Forest classifier was employed for analysis. This study demonstrated the effectiveness of TMA in classifying diseases such as ALS, HD, and PD compared to HC, achieving high accuracies of 79.31\%, 91.67\%, and 87.10\%, respectively.
Begum et al. \cite{begum2020recognition} (ID = C-9) utilized two feature extraction approaches, $i.e.$, Recurrence Quantification Analysis and Fast Walsh-Hadamard Transform, to extract gait features and then fed them into different ML models such as RF, SVM, and KNN to classify NDs and HC.
Nam et al. \cite{nam2020development} (ID = C-10) leveraged differential transformation of gait force (GF) signals and multiscale sample entropy (MSE) to classify different neurodegenerative diseases using KNN and SVM. This study also showed that the accuracy could be improved by applying the method SMOTE to balance the amount of data in each class.
Lin et al. \cite{lin2020evaluation} (ID = C-11) proposed a CNN model to differentiate gait of NDs patients and HC based on vGRF signal data (10-second signal) and achieved a 95.95-100\% accuracy.
Beyrami et al. \cite{beyrami2020robust} (ID = C-12) introduced a non-invasive, cost-effective diagnostic tool for analysing gait signals obtained from vGRF sensors. The best classification results were obtained using SVM model and the achieved average accuracy rates were 100\% for ALS, 99.78\% for PD, and 99.90\% for HD.
Yilmaz et al. \cite{yilmaz2020evaluation} (ID = C-13) explored the usefulness of state space vector, which is combined with KNN, to learn features from gait data for NDs detection. Besides, it also involved statistical features' extraction from gait signals. It also showed the state space approach's potential for analyzing other physiological signals.
Setiawan et al. \cite{setiawan2021identification} (ID = C-14) builded a preprocessing process, a feature transformation process (PCA), and a classification process (CNN and SVM) to leverage five-minute time-frequency spectrogram of vGRF to classify NDs.
A hybrid model that combined multiple sensor data was proposed in Zhao et al. \cite{zhao2021multimodal} (ID = C-15) to learn gait differences between ALS, HD, PD, and healthy individuals. The model included a spatial feature extractor (SFE) and a correlative memory neural network (CorrMNN) for capturing representative features and temporal information, which outperformed many state-of-the-art techniques with a 99.7-100.0\% accuracy. 
Setiawan et al. \cite{setiawan2022development} (ID = C-16) proposed to use wavelet coherence to extract features from gait force signal and to represent those extracted features by using spectrogram image. Those features were then processed by CNN to classify HCs and NDs.
Tobar et al. \cite{tobar2022petri} (ID = C-17) devised Petri net models of human gait with Rs and classification trees (CTs) to classify different NDs(e.g., PD, HD, ALS) from Petri net features of vGRFs, and achieved a 91.7-95.1\% accuracy.
Spatio-temporal gait data were used in the work by Torres et al. \cite{torres2023using} (ID = C-18) to measure the energy cost and power spectral density in gait pathologies of different NDs, while fuzzy c-means techniques, learning algorithm for multivariate data analysis (LAMDA), and neural networks are exploited to classify between NDs and the control group. 
The authors compared various deep learning models including LSTM, GRU, InceptionTime, ResNet, FCN, TST, and PatchTST in their ability to differentiate healthy gaits from PD, ALS, and HD gaits (Zhou et al. \cite{zhou2023deep} (ID = C-19)).

\subsection{Huntington's Disease (HD)}\label{sec_HD}
As shown in Table \ref{AD_ALS_HD_MS_table}, there are 7 studies that include AI-assisted HD diagnosis based on human gait (not including 21 combinatorial studies in Table \ref{combine_table}). The majority of these studies (6 studies) leverage C-ML models, while other studies (1 study) use C-DL models.

Mannini et al. \cite{mannini2016machine} (Table \ref{AD_ALS_HD_MS_table}, ID = H-1) proposed a method to classify gait between HD and health elderly based on IMU data using HMM derived features and SVM classifier, and achieved an overall accuracy of 90.5\% by combining HMM-based , time, and frequency domain features.
Eleven gait features were extracted from accelerometer data to classify HD patients using different machine learning models (\textit{e.g.,} RF, DT, Logiboost, Multibooost) achieving up to 92.86\% accuracy. Two attribute selection algorithms were proposed to choose the most representative gait patterns of subjects (Acosta et al. \cite{acosta2018meta} (ID = H-2)).
Zhang et al. \cite{zhang2019deep} (ID = H-3) used pressure data collected during formation of individual footsteps to classify HD by 3D CNN. The experiments showed that using the basic deep learning backbone VGG16 and similar modules could achieve a classification accuracy of 89\%.
Different machine learning models (KNN, SVM, NB, DT, LDA) were utilized in the work by Felix et al. \cite{felix2019automatic} (ID = H-4) to identify HD from foot force based gait dynamics parameters and achieved up to 100\% accuracy with SVM and DT when considering the right foot stance interval.
Paula et al. \cite{paula2019diagnosing} (ID = H-5) leveraged classic machine learning models (SVM, DT, KNN, NB, LDA) to identify HD from the extracted gait features such as right stance interval.
Huang et al. \cite{huang2019machine} (ID = H-6) utilized five machine learning models (SVM, DT, NV, RF, LR) to recognize HD from gait dynamics features and achieved 85.71-97.14\% accuracy.
Trojaniello \cite{trojaniello2015assessment} (ID = H-7) used the IMU data and SVM model to classify HD gait and healthy gait of elderly, achieving a 90.5\% accuracy.

\subsection{Multiple Sclerosis (MS)}\label{sec_MS}
As reported in Table \ref{AD_ALS_HD_MS_table}, 6 studies that focus on MS diagnosis based on human gait, and there are 4, 1, and 1 using C-ML, C-DL, and A-DL models, respectively. 

Patil et al. \cite{patil2019clinical} (Table \ref{AD_ALS_HD_MS_table}, ID = M-1), used Extreme Learning Machine (ELM), SVM, MLP, and KNN to classify multiple gait disorders including MS, stroke, and cerebral palsy from gait force data, achieving an accuracy result of 89.8\%. 
Kaur et al. \cite{kaur2020predicting} (ID = M-2) proposed two normalization methods to normalized the subject's derived gait characteristics, which were fed into ML models such as SVM, RF, Adaboost, and MLP to classify MS and HC.
Trentzsch et al. \cite{trentzsch2021using} (ID = M-3) investigated the most efficient gait parameters that could aid in the detection of subtle gait changes in individuals with MS. Aiming at this goal, they compared three diagnostic gait systems and employed six different machine learning algorithms. They contributed to the field by providing insights into the use of technology and machine learning in the management of MS.
The study of Creagh et al. \cite{creagh2021interpretable} (ID = M-4) devised a CNN-based transfer learning framework to identify MS patients from wearable smartphone sensor data and achieved up to 91.1\% accuracy. The interpretations of results also suggested that cadence-based measures, gait speed, and ambulation-related signal perturbations were distinct features for classification.
The work by Creagh et al. \cite{creagh2020smartphone} (ID = M-5) leveraged smart-phone and smart-watch to collect gait data and compute related features. Then, the features were fed as input to the ML models (\textit{e.g.,} SVM, LR, RF) to classify HC and MS.
Hu et al. \cite{hu2022machine} (ID = M-6) leveraged gait parameters extracted from raw data of a walkway system to identify HD patients using different classic ML models (\textit{e.g.,} SVM, LR, XGBoost) and achieved up to 81\% accuracy. The performance was shown to be further improved to 88\% when augmenting standard parameters with other custom parameters and normalized subject characteristics.

\subsection{Alzheimer's Disease (AD)}\label{sec_AD}
\hc{As presented in Table \ref{AD_ALS_HD_MS_table}, there are 6 studies that focus on AD diagnosis based on human gait, in which 3, 1, 2 studies exploit C-ML, C-DL, and A-DL models.  }

In the work by Costa et al. \cite{costa2016application} (ID = A-1),
the authors explored the use of ML classifiers including SVM, MLP, RBNs, and DBSs for diagnosing AD based on postural control kinematics. Four classifiers were compared, and the accuracy of AD diagnosis ranges from 91\% to 96.6\% when combining postural kinematics and Montreal Cognitive Assessment (MoCA) variable. This study suggested that machine learning models could aid in computer-aided diagnosis of AD using postural control kinematics.
You et al. \cite{you2020alzheimer} (ID = A-2) proposed a cascade neural network approach that combined gait and EEG data for more accurate and efficient classification of AD. The model utilized attention-based ST-GCN to extract discriminative features from gait data. Experimental results showed superior performance compared to other AD diagnosis methods, with the lower body and right upper limb identified as important for early AD diagnosis.
The study of Ghoraani et al. \cite{ghoraani2021detection} (ID = A-3) collected walking data and derived gait features from participants by using both the "single-tasking" and "dual-tasking' test. The features were then fed as input to the SVM model to classify HC and AD, achieving a 78.0\% accuracy.
In the work by Seifallahi et al. \cite{seifallahi2022alzheimer} (ID = A-4), the authors explored the feasibility of using the Timed Up and Go (TUG) test, a simple balance and walking assessment, as a tool for detecting AD from HC. Joint position data was collected from subjects performing the TUG test using a Kinect camera. Through signal processing and statistical analyses, significant features were identified and used with a support vector machine classifier, achieving a high accuracy (97.75\%).
Different ML models (SVM, KNN, RF, MLP, NB, MLP) by Jeon et al. \cite{jeon2023early} (ID = A-5) were assembled to diagnose AD using gait parameters extracted from IMU sensors signals, and achieved about 72.2\% accuracy. This study totally proposed seven walking experiment paradigms and collected data using seven wearable devices.

\subsection{Amyotrophic Lateral Sclerosis (ALS)}\label{sec_ALS}
As reported in Table \ref{AD_ALS_HD_MS_table}, there are 3 studies that focus on AI-assisted ALS diagnosis based on human gait (excluding 21 combinatorial studies in Table \ref{combine_table}), and all studies employ C-ML models. 

In the work by Xia et al. \cite{xia2016novel} (Table \ref{AD_ALS_HD_MS_table}, ID = S-1), they focused on analyzing gait variability in patients with ALS and extracted two key features, standard deviation statistics and permutation entropy. These features were inputted into a support vector machine classifier to classify ALS patients and healthy controls. The results demonstrated high accuracy (92.86\%) in distinguishing ALS patients from controls, indicating the potential of gait variability analysis as a diagnostic or monitoring tool for ALS.
Felix et al. \cite{felix2020effective} (ID = S-2) leveraged one-minute gait series and parameters to classify between ALS patients and healthy subjects using machine learning models (SVM, KNN, DT) and achieved up to 96.6\% accuracy with DT.
Felix et al. \cite{felix2021automatic} (ID = S-3), in their another work, extracted metrics of fluctuation magnitude and fluctuation dynamics from gait time series, and leveraged five classic ML models (SVM, KNN, NB, LDA, DT) to classify ALS and healthy subjects. Both SVM and KNN achieved the best performance with a 89.7\% accuracy on the GaitNDD database.

\hc{
\section{Comparison with Existing  Surveys}
\label{sec_comparison}
To highlight the distinct contributions of our survey, we provide a brief comparison with the latest representative surveys or reviews on similar topics \cite{gupta2023new,keserwani2024comparative,abumalloh2024parkinson} in this section.
\begin{enumerate}
   \item Comparison with the review by Keserwani et al.  \cite{keserwani2024comparative}: First, it only surveyed the PD and its related AI algorithms, while our survey simultaneously focused on AI techniques for the diagnosis of five typical NDs, including PD, AD, ALS, HD, and MS. Our survey is more comprehensive than this review by including a larger number of representative studies (32 more studies) and cited papers (85 more papers).
Second, the focused main modality and its taxonomy are significantly different. The authors concentrated on the medical data such as neuroimaging, and regarded gait as a \textit{single} modality without further categorizing its data types. In practice, gait can also be captured from different modalities such as RGB videos and force signals ($e.g.,$ vGRF), which requires a finer data classification. In our survey, with a focus on gait-based diagnosis, we present a more \textit{systematic} taxonomy of gait data in terms of different sensor modalities, vision modalities, their combined modalities, and others (see Sec. \ref{sec_gait_taxonomy}). Moreover, this review only briefly mentioned several challenges without detailing their potential solutions and future directions, while our survey provided a more comprehensive discussion on different challenges with an elaboration on the potential future solution for each challenge. We further offer a novel research vision on the application of 3D skeleton data (see Sec. \ref{sec_challenges_and_directions}).

\item Comparison with the review by Gupta et al. \cite{gupta2023new}: (1) Firstly, this review only concerned with Parkinson’s Disease (PD) related research, while our review covered the 5 most prevalent neurodegenerative diseases (NDs), including PD, Alzheimer’s disease (AD), Amyotrophic Lateral Sclerosis (ALS), Huntington’s Disease (HD), and Multiple Sclerosis (MS). (2) Secondly, the focus of the data in this review and ours are distinctly different as ours focused entirely on works that harness different types of gait data. For instance, sensor-based (\textit{e.g.,} Inertial Measurement Unit (IMU), force-sensitive resistors), vision-based (\textit{e.g.,} 3D motion capture system), and multi-modal (\textit{e.g.,} Microsoft Kinect) gait data. However, the compared review \cite{gupta2023new} focused on works that explored various PD data modalities, such as biomarkers, speech recording, handwriting data. (3) In addition, the authors reviewed those works that target at different purposes such as PD diagnosis, treatment, management, identification of novel biomarkers in the progression of disease, and using AI to advance neurosurgical process and drug discovery, while our focus was solely on NDs' diagnosis. (4) The review also discussed the potential role of the metaverse, the Internet of Things, and electronic health records in the effective management of PD to improve the quality of life. However, it lacks a discussion regarding current biggest challenges in current field of interest such as data scarcity, multi-modal integration, and the design of AI models which are detailed out in our review. What's more, we also proposed harnessing novel 3D skeleton-based gait data to further advance the development of the field by addressing the state-of-the-art approaches related.

\item  Comparison with the review \cite{abumalloh2024parkinson}: (1) Compared to review~\cite{abumalloh2024parkinson} that focuses solely on Parkinson’s Disease (PD) and deep learning (DL) techniques through bibliometric analysis and method review, our study is broader and deeper in scope, methodology, and analysis. First, while the survey is limited to PD, we cover five major neurodegenerative diseases (NDs), including Alzheimer’s Disease (AD), amyotrophic lateral sclerosis (ALS), Huntington’s Disease (HD), and Multiple Sclerosis (MS), providing a more comprehensive perspective. (2) In terms of diagnostic methods, the review~\cite{abumalloh2024parkinson} emphasizes various PD-related indicators (e.g., biomarkers, handwriting, and speech analysis), whereas we focus on gait data as a key motor symptom, introducing a systematic taxonomy of gait data types (e.g., sensor modalities, vision modalities, and their combinations). (3) While the review~\cite{abumalloh2024parkinson} utilizes VOSviewer for bibliometric analysis, we not only review more studies (164) but also propose an innovative quality evaluation criterion to quantify the quality of existing research, adding depth to our analysis. (4) The review~\cite{abumalloh2024parkinson} highlights research gaps in incremental learning and big data analysis without offering detailed solutions, whereas we comprehensively discuss challenges such as data scarcity, multi-modal integration, and AI model design, proposing potential solutions and envisioning the application of 3D skeleton data and the development of efficient AI models. Through broader disease coverage, a stronger focus on gait analysis, and a detailed discussion of challenges and solutions, our survey provides systematic and forward-looking guidance for AI research in ND diagnosis.


\end{enumerate}


\section{Representative Datasets and Methods}
\label{sec_representative_datasets}
\subsection{Public Datasets} 
According to our survey, many studies manually collect their own gait data without publishing them as a public benchmark. Here we introduce several most commonly used gait datasets.

\begin{itemize}
\item \textit{Gait in Neurodegenerative Disease Database} \cite{goldberger2000physiobank}. This database includes force-based records from patients with Parkinson's disease (15 persons), Huntington's disease (20 persons), and amyotrophic lateral sclerosis (13 persons). Additionally, records from 16 healthy control subjects are provided. The raw data were collected using force-sensitive resistors, which produce an output roughly proportional to the force exerted under the foot. From these signals, stride-to-stride measurements of footfall contact times were derived.

\item \textit{Gait in Parkinson's Disease} \cite{goldberger2000physiobank}. It includes gait measurements from 93 patients with idiopathic Parkinson's disease (mean age: 66.3 years; 63\% male) and 73 healthy controls (mean age: 66.3 years; 55\% male). Vertical ground reaction force data were recorded as participants walked at their usual, self-selected pace for approximately 2 minutes on level ground. Each foot had 8 sensors that measured force over time. The output from each of these 16 sensors was digitized at a rate of 100 samples per second. Additionally, two signals reflecting the summed outputs of the 8 sensors per foot are included.

\item \textit{SDUGait} \cite{wang2016gait}. This dataset includes 52 subjects (28 males and 24 females) with an average age of 22. Each subject has 20 sequences, which contain at least 6 fixed walking directions and 2 arbitrary directions, totaling 1,040 sequences. It was collected using the second-generation Kinect (Kinect V2), which not only provides a broader viewing angle but also produces higher-resolution depth images with 25 body joints.


\item Yogev \textit{et al.} (also termed Ga dataset) \cite{yogev2005dual}. This dataset collects vertical ground reaction force data of subjects by putting 8 sensors on their feet when subjects walk on level ground. It contains 29 PD patients (9 females, 20 males) with an average age of 61.6 years, an average height of 1.67 meters, and an average weight of 73.1 kilograms. 18 healthy controls (8 females, 10 males) are involved with an average age of 57.9 years, an average height of 1.68 meters, and an average weight of 74.2 kilograms.

\item Frenkel-Toledo \textit{et al.} (also termed Ju dataset) \cite{frenkel2005treadmill}. This dataset collects vertical ground reaction force data by 8 sensors when the  subjects  walk  on a treadmill. It contains 35 PD patients (13 females, 22 males) and 29 healthy controls (11 females, 18 males).

\item Hausdorff \textit{et al.} (also termed Si dataset) \cite{hausdorff2007rhythmic}.  This dataset collects vertical ground reaction force data when the subjects move at a comfortable place with Rhythmic Auditory Stimulation (RAS). It contains 29 PD patients (13 females, 16 males) and 26 healthy controls (14 females, 12 males).

\item \textit{Ga, Ju, Si Groups} \cite{frenkel2005treadmill,hausdorff2007rhythmic,yogev2005dual}. This dataset is a combination of the above three small datasets Ga \cite{yogev2005dual}, Si \cite{frenkel2005treadmill}, and Ju \cite{hausdorff2007rhythmic}, and is commonly used in gait-based PD classification and diagnosis tasks.

\end{itemize}
All the above datasets are widely used as public benchmarks to evaluate the performance of models.
We further discuss the challenges of existing datasets in Sec. \ref{data_issues}.

\subsection{State-of-The-Art Methods and Performance}
We utilize the proposed quality evaluation criterion (see Sec. \ref{quality_criterion}) to select the most representative state-of-the-art methods from studies with the highest quality:

\textbf{Representative State-of-The-Art Methods for Gait-Based PD Diagnosis:} 
\begin{itemize}
    \item In \cite{wang2016recognizing} (ID = P-15, $S_{Q}=1.00$), a novel hybrid model based on BP neural networks is proposed to classify PD and HC, which outperforms four baseline methods on a public dataset with an average accuracy of 87.9\%.

    \item \cite{jane2016q} (ID = P-17, $S_{Q}=1.00$) introduces a Q-backpropagated Time Delay Neural Network (Q-BTDNN), which harvests the temporal gait data collected from wearable devices, combined with the Q-backpropagation (Q-BP) algorithm for predicting gait disturbances in PD.

    \item \cite{alharthi2020gait} (ID = P-75, $S_{Q}=1.00$) uses deep CNN to classify gait deterioration in PD patients based on spatio-temporal vGRF signals and achieves 95.5\% accuracy in average. The study suggests that the heel strike and body balance are the most indicative gait elements in PD classification.

    \item \cite{nguyen2021classification} (ID = P-82, $S_{Q}=1.00$) develops a novel NN model to classify PD and HC based on the vGRF dataset collected from a public dataset and achieves an accuracy result of 84.8\%.

    \item \cite{salimi2023type} (ID = P-85, $S_{Q}=1.00$) proposes an interval type-2 fuzzy neural network with novel quasi-Levenberg-Marquardt (qLM) learning approach to identify PD patients from the gait cycle based vGRF signals. Based on the 10 clinical features extracted from 14 gait features, the model achieves 97.6\% accuracy with a smaller size of parameters and a good interpretability ($i.e.$, rules based on fuzzy logic). 

    \item In \cite{yang2022pd} (ID = P-93, $S_{Q}=1.00$), the authors propose a novel architecture PD-ResNet based on ResNet \cite{he2016deep} and use it to differentiate PD and HC. Its result outperforms many existing state-of-the-art models with 95.5\% accuracy. 

    \item In \cite{acsurouglu2022deep} (ID = P-97, $S_{Q}=1.00$), a novel model with CNN and RF is devised to perform PD detection tasks from time and frequency features of vGRF, achieving 99.5\% accuracy.

    \item In \cite{zhao2021multimodal} (ID = P-106, $S_{Q}=1.00$) proposes a correlative memory neural network (CorrMNN) scheme to extract spatio-temporal features to discriminate NDs' gait from normal gait based on vGRF, and achieves up to 100\% accuracy.

    \item In \cite{dong2023static} (ID = P-123, $S_{Q}=1.00$) devises a static-dynamic temporal network that consists of one-dimensional, two-dimensional convolutional networks, and attention mechanisms to classify PD patients based on foot force (vGRF time series) and achieves 96.7\% accuracy.

    \xj{
    \item Johnson et al. (ID = P-125, $S_{Q}=0.88$) addressed early-stage PD monitoring and diagnostic challenges using consumer-grade wearable devices and sensor data from the WATCH-PD study. Data were gathered over a year from early-stage PD patients (n = 82) and age-matched controls (n = 50). Leveraging disease- and behavior-relevant feature engineering with multivariate machine learning, the authors developed a PD screening tool based on a random forest model. Achieving 92.3\% accuracy, 90.0\% sensitivity, and 100\% specificity (AUC = 0.92), this approach underscores the potential of remotely monitored wearable devices for population-wide PD screening and monitoring.

    \item The necessity for objective and efficient gait assessment in PD using the Movement Disorder Society Unified Parkinson’s Disease Rating Scale~(MDS-UPDRS) is proposed in~\cite{tian2024cross} (ID = P-126, $S_{Q}=1$). To overcome the limitations of existing PD gait skeleton datasets, they collect a sizable dataset including 102 PD patients, 30 healthy older adults, and 16 young adults using multi-view Azure Kinect sensors. In addition, this paper develops a cross-spatiotemporal graph convolution network (CST-GCN) that captures complex gait features by modeling cross-spacetime dependencies between joints to assess age-related effects on PD gait.
    
    }

\end{itemize}

\textbf{Representative State-of-The-Art Method for Gait-Based AD Diagnosis:} 
\begin{itemize}

\item \cite{you2020alzheimer} (ID = A-2, $S_{Q}=0.89$) proposes an innovative cascade neural network approach that combines gait and EEG data for more accurate and efficient classification of AD. The model utilizes attention-based ST-GCN to extract discriminative features from gait data. Experimental results show superior performance compared to other AD diagnosis methods, with the lower body and right upper limb identified as important for early AD diagnosis.

\xj{
\item CLADSI~\cite{bringas2024cladsi} (ID = A-6, $S_{Q}=0.77$) addresses the challenge of continuous monitoring and stage identification in Alzheimer’s disease (AD) using motion sensor data. Data from 35 AD patients, monitored over a week via accelerometers in a daycare center, was used to develop a method enabling CNN to learn from a continuous stream of sensor data without retaining prior data. This approach allows CNN to self-configure and adapt as new data is received. The CLADSI achieved accuracy rates of 86.94\%, 86.48\%, and 84.37\% for 2, 3, and 4-stage classification tasks, respectively, demonstrating potential for continual learning applications in long-term patient monitoring without human intervention.

}

\end{itemize}

\textbf{Representative State-of-The-Art Method for Gait-Based ALS Diagnosis:} 
\begin{itemize}

\item In \cite{felix2021automatic} (ID = S-3, $Q_{S}=0.67$) extracts metrics of fluctuation magnitude and fluctuation dynamics from gait time series, and leverages five classic machine learning models (SVM, KNN, NB, LDA, DT) to classify ALS and healthy subjects. Both SVM and KNN achieves the best performance with 89.7\% accuracy on the GaitNDD database.

\end{itemize}

\textbf{Representative State-of-The-Art Method for Gait-Based HD Diagnosis:} 
\begin{itemize}

\item In \cite{paula2019diagnosing} (ID = H-5, $S_{Q}=0.78$) leverages classic machine learning models (SVM, DT, KNN, NB, LDA) to identify HD from the extracted gait features such as right stance interval.

\end{itemize}

\textbf{Representative State-of-The-Art Method for Gait-Based MS Diagnosis:} 
\begin{itemize}

\item The study of \cite{creagh2021interpretable} (ID = M-4, $S_{Q}=0.89$) devises a novel CNN-based transfer learning framework to identify MS patients from wearable smartphone sensor data and achieves up to 91.1\% accuracy. The interpretations of results also suggest that cadence-based measures, gait speed, and ambulation-related signal perturbations are distinct features for classification.

\end{itemize}

\textbf{Representative State-of-The-Art Methods for Gait-Based Combinatorial Diagnosis of Different NDs (PD, ALS, HD):} 
\begin{itemize}

\item In \cite{beyrami2020robust} (ID = C-12, $S_{Q}=0.89$) introduces a non-invasive, cost-effective diagnostic tool for analyzing gait signals obtained from vGRF sensors. The best classification results are obtained using SVM model and the achieved average accuracy rates are 100\% for ALS, 99.78\% for PD, and 99.90\% for HD.

\item A hybrid model that combines multiple sensor data is proposed in \cite{zhao2021multimodal} (ID = C-15, $S_{Q}=0.89$) to learn gait differences between ALS, HD, PD, and healthy individuals. The model includes a spatial feature extractor (SFE) and a correlative memory neural network (CorrMNN) for capturing representative features and temporal information, which outperforms many state-of-the-art techniques with 99.7-100.0\% accuracy. 

\xj{
\item A reliable diagnostic model is proposed in~\cite{erdacs2024cnn} (ID = C-20, $S_{Q}=0.77$) for neurodegenerative diseases using gait data transformed into QR codes and classified with CNN. This model, trained on QR-encoded gait data from patients (PD: n = 15, HD: n = 20, ALS: n = 13) and healthy controls (n = 16), achieved high accuracy in distinguishing neurodegenerative conditions: 94.86\% for NDD vs. control, 95.81\% for PD vs. control, 93.56\% for HD vs. control, 97.65\% for ALS vs. control, and 84.65\% for PD vs. HD vs. ALS vs. control. These results highlight the potential of this approach as a complementary diagnostic tool, especially for individuals exhibiting different levels of motor impairment.

\item A multimodal gait-abnormality-recognition framework named CNN-BiLSTM is proposed in~\cite{li2023multimodal} (ID = C-21, $S_{Q}=1$) to enhance neurological disease detection, addressing limitations in single-sensor approaches. This framework processes data from multiple sensors, utilizing an adaptive sliding window and converting time series into time-frequency plots for unified multi-sensor representation. In testing, the framework achieved 98.89\% accuracy in Parkinson’s severity classification and high recognition rates for ALS (100\%), PD (96.97\%), and HD (95.43\%). The adaptability and efficiency of CNN-BiLSTM make it promising for daily gait monitoring and personalized rehabilitation planning.

}

\end{itemize}

As these studies with the highest quality scores possess higher novelty, more comprehensive comparison, and more sufficient experimental samples that are crucial for creating advanced models, the AI models in these studies can be viewed as state-of-the-art representatives for the diagnosis of different NDs. Notably, both classic machine learning models (6 studies, 30.0\%) and deep learning models (14 studies, 70.0\%) are valuable for NDs diagnosis tasks and can achieve highly competitive performance. 
However, there are still many challenges (e.g., generalizability) confronted by these AI models, and we further discuss them in Sec. \ref{challenges_AI_models}.

\ml{\textbf{Representative State-of-The-Art Multi-Modal Approaches for Gait-Based Diagnosis of NDs} 
\begin{itemize}
    \item These two following articles both employed Microsoft Kinect for data collection.
    First article \cite{buongiorno2019low} extracted postural and kinematic features (gait, finger tapping, and foot tapping) for PD classification. A total of nine features for gait analysis were used to train an ANN classifier to perform PD classification with 89.4\% accuracy. 
    Second study \cite{tupa2015motion} presented a digital signal processing methods that to extract gait features such as leg length, stride length, and gait velocity. Results showed accurate classification using neural networks, indicating the potential use of Kinect sensors for gait analysis and disorder recognition. The study discussed the possibility of using Kinect as an inexpensive alternative to complex camera systems.
    \item This work used a marker-based IR camera and a force-sensitive platform to derive gait parameters such as spatiotemporal, kinematic, and kinetic features. It then utilized two machine learning model, SVM and ANN, to classify PD and HC by using those features and achieved a high accuracy \cite{tahir2012parkinson}.
    \item This work \cite{chatzaki2022can} combined conventional camera, IMUs, and a force platform for data collection. It also constitutes an experimental protocol for the assessment of PD motor signs of lower extremities, and leverages classic machine learning models to classify PD patients from 18 gait features extracted from gait data and achieved an 88\% accuracy.
    \item This article \cite{zhao2021multimodal} utilized not only Kinect, but also vGRF and IMU sensors to collect data and proposed a correlative memory neural network scheme to extract spatio-temporal features and discriminate NDDs' gait from normal gait and achieved an up to 100\% accuracy. The proposed model can also be used to perform gait recognition which outperformed existing state-of-the-art methods.
\end{itemize}
}

}
\section{Current Challenges and Future Directions}
\label{sec_challenges_and_directions}
In this section, we first identify the key challenges in AI-assisted NDs diagnosis based on human gait. Then, with regard to these challenges, we correspondingly propose potential solutions and promising future directions. Finally, we provide a research vision on 3D skeleton based gait representations and AI models for more efficient NDs diagnosis.

\subsection{Limited NDs Types and Studies}
As shown in Sec. \ref{ND_type_analysis}, most studies (87.2\%) in this area focus on PD gait and related diagnosis, while the number of other NDs studies is very limited. For example, although AD is the most influential neurodegenerative disease around the world (see Fig. \ref{overview_pop}), only a few works (5 studies) focus on the AI-assisted diagnosis of AD from gait. On the other hand, apart from the five most representative NDs focused in this survey, there are also other rare NDs such as prion disease and motor neuron disease, which deserve more attention and effort to explore AI solutions for their gait-based diagnosis. 

In the future, it is necessary to conduct a deeper investigation on abnormal gaits of more NDs, especially the rarely-explored NDs, and devise corresponding AI models to automatically assist their clinical diagnosis. Another important direction is to devise a general gait learning model that can identify common and different gait abnormality across various NDs, so as to improve the generalization capacity of gait analysis models to serve for more other diseases.

\subsection{Scarcity and Imbalance of Data}
\label{data_issues}
The medical data including gait data collected for NDs diagnosis are highly limited \cite{elazab2024alzheimer}, and in the selected 164 articles only several datasets are publicly available: The \textit{Gait in Neurodegenerative Disease Database} and \textit{Gait in Parkinson's Disease} from Physionet database \cite{goldberger2000physiobank} are two most frequently-used used datasets;
Other datasets include \textit{SDUGait} \cite{wang2016gait},
\textit{Floodlight PoC dataset} \cite{midaglia2019adherence}, 
\textit{UK Biobank Dataset} \cite{doherty2017large}, 
\textit{Ga, Ju, Si Groups} \cite{frenkel2005treadmill,hausdorff2007rhythmic,yogev2005dual}.
However, these datasets are limited in both sample sizes and NDs types. Most of them contain less than 100 subjects (see Sec. \ref{sample_size_analysis}), and the number of NDs types is typically less than 3, while PD gait data are much more than other NDs in terms of quantity and quality ($i.e.,$ unbalanced data size of different diseases).
This could limit the learning performance of AI models especially deep learning models, which inherently require large-scale and class-balanced data  \cite{fang2024your} to achieve more reliable prediction. 

To address these challenges, more high-quality gait data of different NDs should be collected. First, it is feasible to cooperate with larger hospitals, medical institutes, or communities to involve more patients and health controls into the collection of gait and related data. Second, the number of patients in each disease and the data size of each patient should be controlled within a reasonable range, $e.g.,$ enlarging the sample size of each disease to more than hundreds and keeping the total data size of each disease as similar as possible. Such balanced data could benefit the model to learn higher inter-class difference with better robustness against intra-class diversity \cite{johnson2019survey}. 
To address the challenges of data scarcity and imbalance in this area and facilitate related research, we will collect and open a large-scale multi-modal gait dataset for NDs diagnosis in the future.

\subsection{Lack of Multi-Source/Modal Gait Data}
As reported in Sec. \ref{data_type_analysis}, most studies in this area typically collect either sensor-based or vision-based gait data with a single device. For sensor-based data, force sensitive resistor or IMU is the commonly-used device, while they are rarely combined in the same study for gait data collection. Nevertheless, as AI-assisted disease diagnosis is a vital task with high demand of model accuracy and reliability, utilizing only single-source or single-modality data might be insufficient to fully capture latent disease features from different dimensions. Taking PD as an example, the vision-based appearance features ($e.g.,$ unnatural facial expression) and sensor-based motion features ($e.g.,$ walking tremor) can be complementary to provide a full picture for PD diagnosis. 

Therefore, an important future direction is to collect gait data from multiple sources and modalities  \cite{elazab2024alzheimer}. First, it is feasible to combine multiple devices as different sources of same-type data ($e.g.,$ multi-view data with differently-positioned cameras), where all devices/sources can compensate each other to better reconstruct or depict the whole pattern with less information loss \cite{liang2020behavioral}. Second, combining different modalities such as RGB images, depth images, 3D skeletons, and force data is valuable as they can provide gait information from different dimensions ($e.g.,$ appearances, poses) to better predict NDs \cite{zhao2021multimodal}. 

\subsection{Challenges in AI Model Design}
\label{challenges_AI_models}
According to the analysis of existing AI methods in gait-based NDs diagnosis, there are several challenges in designing more reliable AI models as follows.

\textbf{(1) Accuracy:} Existing studies report the model accuracy typically ranging from 60\% to around 95\%, with very few studies possessing accuracy less than 60\%. Many studies using C-ML models report model performance with very high accuracy. For the low-accuracy methods, they cannot reach the requirement to assist in medical diagnosis, as the frequent wrong prediction might induce serious medical risk. However, for the methods with very high accuracy, careful evaluations with other indicators such as generalizability and interpretability are required. It is worth noting that these models with nearly 100\% accuracy possibly overfit the training data while the the testing data may possess highly similar data distribution to training data ($i.e.,$ small domain shift \cite{sun2016return}). In this case, these AI models are hard to get satisfactory performance on out-of-distribution data such as new real-world gait data.  

\textbf{(2) Generalizability:}
Despite high accuracy of quite a few AI methods in this area, many of them are trained, validated, and tested on a single dataset or self-collected gait data, which usually contain limited data sizes, views, scenes or conditions. Therefore, these methods may only perform well on the scenarios that are similar to that of the training data, while they cannot generalize to more challenging data \cite{bar2020impact}. For example, if the training data contain only gait data of elderly patients, the trained AI model could possess lower accuracy when predicting NDs of young adults using corresponding gait data. In summary, the AI model trained on a single dataset or datasets with limited scenarios ($e.g.,$ single setting) typically possesses low generalization ability in real-world application.

\textbf{(3) Efficiency:}
As shown in Sec. \ref{model_type_analysis} and Table \ref{num_types}, 
the majority of existing studies (104 papers, 63.4\%) utilize C-ML models, which typically enjoy much fewer model parameters and less computational complexity than deep learning models. In general, C-ML models is more efficient than DL models (C-DL and A-DL) under the same accuracy, which makes C-ML more favorable to be deployed in real-world applications. However, compared with conventional ML models, the emerging DL models have stronger ability in mining effective patterns from large-scale medical data, which are more useful and popular in the big data era. For DL models, it is necessary to achieve a good trade-off between model efficiency ($e.g.$, model parameter size, computational complexity) and model performance ($e.g.$, accuracy, speed) \cite{menghani2023efficient}, so as to better assist in disease diagnosis.

\textbf{(4) Interpretability:}
In high-stake fields like medical diagnosis, it is important for AI models to provide human-friendly reliable explanations to facilitate the medical decision \cite{tjoa2020survey,payrovnaziri2020explainable,zhang2022applications}. In the surveyed area, there are only a few C-ML model based studies that have identified the important body parts or gait features in classification, while most DL models do not devise an effective mechanism such as feature disentangling and importance visualization \cite{zhang2018visual} to provide interpretability. This may increase the risk of wrong diagnosis and discourage the users such as doctors from adopting the predicted result. 

\hc{
\textbf{(5) Fairness of Comparison:}
Since a unified model evaluation protocol ($e.g.,$ metrics) is still unexplored in this area, the performance comparison between different studies/models might be unfair. For example, some studies adopt the area under the Receiver Operating Characteristic (ROC) curve for evaluation, while some others take accuracy as metric. Moreover, some studies evaluate their models on different testing sets of the same dataset, or use different datasets to train models, which they cannot be fairly compared even using the same evaluation metric. In our survey, we report the model accuracy as the main metric, while their comparison should be conducted under using the same metric, dataset, and testing set.

To address the aforementioned challenges, it is essential to improve not only the accuracy of an AI model but also its generalizability, efficiency, and interpretability. A potential future direction is to exploit more advanced deep learning models and larger-scale gait datasets with different scenario settings to learn both domain-specific ($e.g.,$ AD-specific) and domain-general ($i.e.,$ NDs-shared) gait features, so as to improve the model accuracy and enable it to be transferred/generalized to other scenarios. For efficiency, it is feasible to employ lighter network architecture ($e.g.,$ MobileNets \cite{howard2017mobilenets}) or model compression techniques ($e.g.,$ knowledge distillation \cite{gou2021knowledge}, network pruning \cite{liu2018rethinking}) to reduce the model size and computational complexity but retaining similar model accuracy. Another promising direction is to adopt smaller gait data such as the emerging 3D skeleton data, which can concisely represent human poses with key body joints and typically require lightweight AI models to learn discriminative gait patterns \cite{rao2021self}.
To improve model interpretability, different types of human-friendly explanation including text description and feature visualization can be considered. We can integrate model-specific explanation mechanisms into different AI architectures, $e.g.,$ visual class activation maps (CAM) \cite{zhou2016learning} for CNN, knowledge graphs \cite{ji2021survey} for GNN. Inspired by the recent success of ChatGPT, it can be used to provide insightful description for the importance of features if we transform gait features into text format. Moreover, ChatGPT can also act as an agent to lead and improve the AI model learning \cite{haupt2023ai}, in which we can add prompts to generate interpretable results. As there exist key body parts and poses for NDs diagnosis, we can try to disentangle the gait abnormality related semantics from normal motion semantics ($e.g.,$ daily actions) to identify the most important or unique component of NDs patients' features. Simultaneously improving the accuracy, generalizability, efficiency, and interpretability, it is hopefully to obtain a more reliable and practical AI model for gait-based NDs diagnosis. Meanwhile, a unified evaluation protocol, which synergizes different benchmark datasets, validation settings, and evaluation metrics, should be further devised to fairly compare different AI models to facilitate the development of this area.
}

\begin{figure*}[t]
    \centering
    \scalebox{0.65}{
    \includegraphics{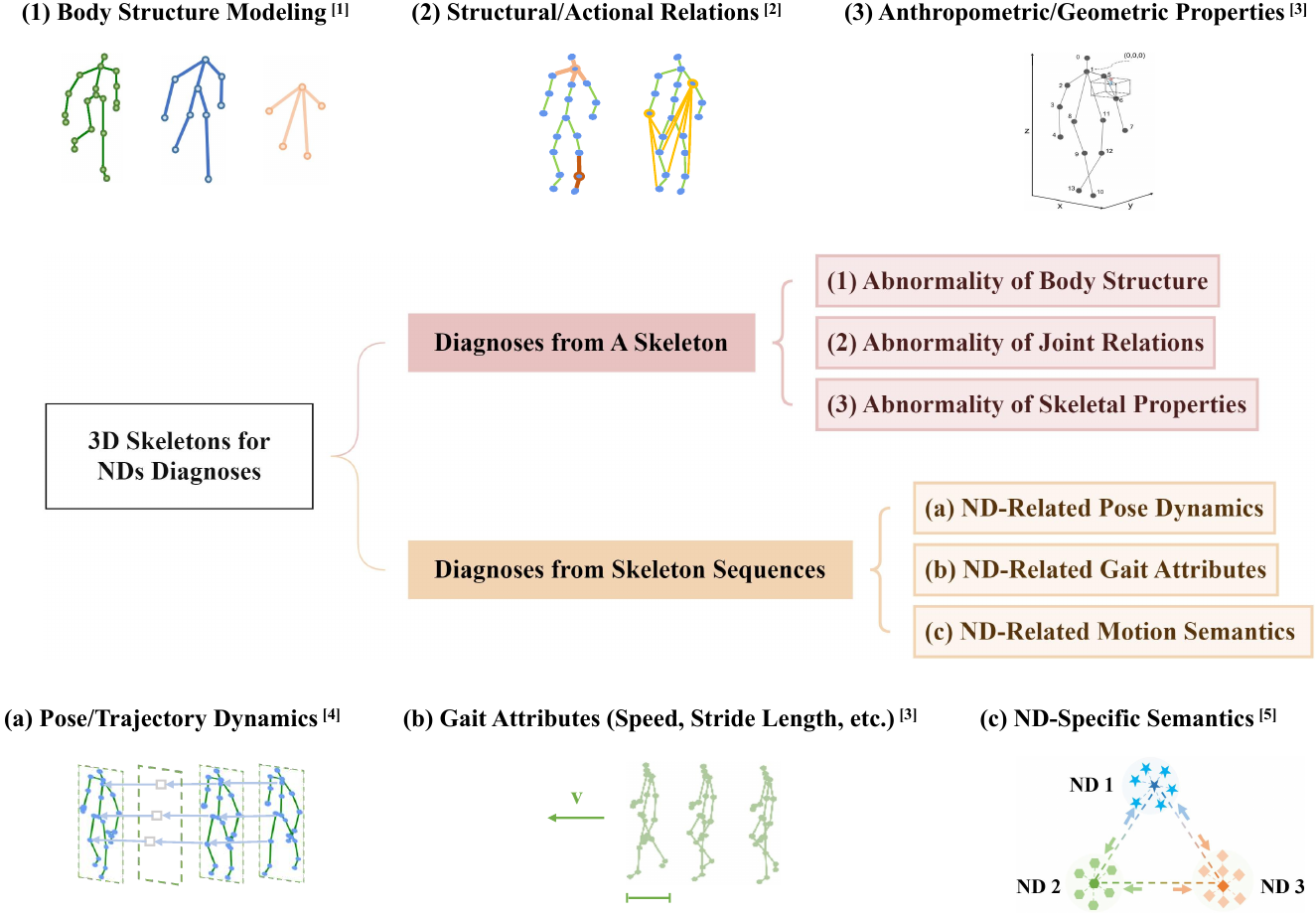}
    }
    \caption{Overview of 3D skeleton based NDs diagnosis framework. We visualize different directions of AI model design in terms of skeleton-level ((1)-(3)) and sequence-level NDs diagnosis ((a)-(c)), and also provide representative AI methods of this direction as an example solution (${[1]}$: \cite{rao2023hierarchical}; ${[2]}$: \cite{rao2021sm}; ${[3]}$: \cite{andersson2015person}; ${[4]}$: \cite{rao2023transg}); ${[5]}$: \cite{rao2022simmc})}
    \label{skeleton_framework}
\end{figure*}

\subsection{Research Vision: Application of 3D Skeleton Data}
\label{sec_research_vision}
Recently driven by economical, non-obtrusive and accurate skeleton-tracking devices like Kinect \cite{shotton2011real-time}, 3D skeleton data has been a popular and generic data modality for many gait-related tasks such as gait recognition and person re-identification \cite{pala2019enhanced,liao2020model,andersson2015person,munaro20143d,munaro2014one,barbosa2012re,rao2020self,rao2021self,rao2021sm,rao2021multi}. A 3D skeleton is defined as 3D coordinates of key human joints (typically 20 or 25 key joints \cite{shotton2011real-time}) of a person, while 3D skeletons are defined as all joints' temporal series conveying motion dynamics of the person.
In this concise way, 3D skeletons can provide both body structure and pose information for gait analysis. Encouraged by the promising results of 3D skeletons in many fields, we propose the research vision of exploiting 3D skeleton data to characterize NDs-related gait patterns to enhance the diagnosis performance. In this section, we first discuss the unique advantages of 3D skeleton data for gait representation and learning, which can potentially address challenges in AI-based NDs diagnosis in terms of efficiency and scalability (see Sec. \ref{sec_advantages_skeletons}).
Then, we propose a generic 3D skeleton based framework to design AI models specifically for NDs diagnosis (see Sec. \ref{sec_skeleton_framework}).

\subsubsection{Skeletons as Efficient and Scalable Gait Representations}
\label{sec_advantages_skeletons}
Unlike traditional methods that rely on visual appearance features ($e.g.,$ silhouettes) or force signals ($e.g.,$ vGRF) to capture gait, 3D skeletons can simultaneously model human body structure, poses, and gait patterns using \textit{only} 3D positions of key body joints. They can not only offer detailed spatial 3D coordinates and their temporal dynamics for statistical gait analysis, but also provide intuitive body representations ($e.g.,$ pose visualization) that help illustrate variations in gait. Moreover, the 3D skeletons can be conveniently captured by a single contactless device such as Kinect.

Specifically, 3D skeleton data possess the following advantages to improve the efficiency of gait learning: (a) Smaller data inputs compared with appearance-based methods that typically require large-size ($e.g.,$ high-resolution) image/video data; (b) Lighter AI models than conventional image-based models to process and learn skeletons \cite{rao2022simmc}; (c) Lower resource/device requirement compared with other sensor-based data that require multiple devices such as IMUs or expensive systems such as walkway systems; (d) Higher convenience of collection with unobtrusive and contactless detection compared with other sensor methods that usually require wearing devices ($e.g.,$ sensors) or body markers. These advantages enable them to be potentially applied to capturing and learning gait with higher efficiency in terms of real-world deployment and model design.

On the other hand, 3D skeletons can serve as scalable gait representations for different application scenarios: (a) For medical diagnosis and other privacy-sensitive areas, 3D skeletons can replace traditional vision data such as RGB images to \textit{exclude} all visual appearance information ($e.g.,$ faces of patients), so as to provide better privacy protection for downstream tasks such as gait classification;
(b) Compared with other gait representations that rely on RGB or depth images, 3D skeleton based representations possess more robust performance under variations of scales, views or other external factors \cite{han2017space}, which enables them to be flexibly used in more scenarios;
(c) As 3D skeletons are general modality for many tasks including action recognition \cite{rao2021augmented}, gait recognition \cite{liao2020model}, and person re-identification \cite{rao2021self}, all these tasks could be flexibly transferred or combined to provide better abnormal gait classification for NDs diagnosis. For example, the most distinctive pose regions in action recognition tasks can be utilized to locate the gait abnormality of NDs patients, while the identity-specific gait patterns in person re-identification tasks can be exploited to recognize NDs patients and help decouple disease-specific gait patterns; (d) There are many successful practices to combine 3D skeleton data and other modalities such as RGB images, depth images, and 3D clouds \cite{pala2019enhanced,munaro20143d} to improve feature effectiveness for various tasks. It is feasible to combine 3D skeleton data with the above data modalities or other data such as IMUs signals and force signals to further boost the performance of gait learning and NDs diagnosis. It is worth noting that the most commonly used skeleton-tracking device, Kinect, can simultaneously capture RGB images, depth images, and 3D skeletons, which provides great benefit for multi-modal representations of gait.

\subsubsection{3D Skeleton based NDs Diagnosis Framework}
\label{sec_skeleton_framework}
As presented in Fig. \ref{skeleton_framework}, by using 3D skeletons as gait representations, NDs can be diagnosed from two main aspects, a single skeleton and a skeleton sequence. For skeleton-level NDs diagnosis, the abnormalities in body structure, joint relations, and skeletal properties are key focus of AI models: (1) We can exploit multi-level or hierarchical body structure modeling \cite{rao2023hierarchical} to help capture more discriminative features between NDs patients and healthy subjects; (2) The structural and actional relations between body joints or parts usually characterize unique walking patterns of a person \cite{murray1964walking}, so we can combine them to learn both local and global motion correlations using AI models ($e.g.,$ graph attention networks \cite{feng2024multi}) to better catch valuable patterns \cite{rao2021sm} ($e.g.,$ abnormal gait patterns); (3) There exist crucial skeletal properties including anthropometric and geometric attributes ($e.g.,$ joint distances and angles) that can help recognize different actions \cite{rao2021augmented} or identify different persons \cite{andersson2015person}. These key skeletal properties incorporating domain-specific knowledge could be potentially used to identify different NDs.

For skeleton sequence level NDs diagnosis, learning NDs-related pose dynamics, gait attributes, and latent motion semantics is the key to AI model design (see Fig. \ref{skeleton_framework}): (a) For NDs patients, their poses might be unnatural or twisted, which could be reflected on their motion dynamics such as body-joint trajectory. Therefore, it is feasible to capture pose or trajectory dynamics from 3D skeletons \cite{rao2023transg} and extract the NDs-related parts for diagnosis; (b) Considering that NDs patients typically possess abnormal gait attributes such as lower speed and uncertain stride length, we can compute these gait features from 3D skeletons \cite{andersson2015person} as important discriminators of different diseases; (c) Advanced AI models such as deep learning models can learn latent high-level semantics in high-dimensional feature space to help better classify abnormal gaits. For example, we may devise self-supervised tasks such as skeleton sequence reconstruction, sorting \cite{rao2021self}, forecasting \cite{feng2022relation}, or contrasting \cite{rao2022simmc} to learn high-level semantics in terms of motion continuity and consistency, which can be used to detect gait abnormality by comparing with normal samples for NDs diagnosis. We can also leverage unsupervised clustering algorithms to find latent gait semantics ($e.g.,$ gait prototypes \cite{rao2022simmc}) related to different NDs for diagnosis.

\section{Conclusion}
\label{conclusion}
In this paper, we provide an overview of the recent advancements of AI models for diagnosing NDs through human gait. Our survey showcases a general process of gait-based NDs diagnosis using AI with focus on five representative NDs: PD, AD, HD, ALS, MS, and systematically reviews all existing studies with an elaboration on their used gait data, AI models, and overall performance. 
A novel quality evaluation criterion is also devised to quantitatively assess the quality of surveyed studies.
We outline and discuss the current challenges, potential solutions, and promising future directions in this field. Furthermore, we propose a novel research vision on the utilization of emerging 3D skeleton data for human gait representation and AI model development for more efficient NDs diagnosis.

As an important interdisciplinary topic, utilizing AI technologies for gait-based NDs diagnosis requires expertise in gait analysis, medical science, and AI. Therefore, it may necessitate a significant amount of time and effort to conduct an in-depth understanding and exploration of this topic. 
By systematically reviewing existing technologies, revealing their key challenges, and identifying future research directions, our survey aims to provide a quick view of this topic to researchers especially those with knowledge in gait, medicine, or AI alone.
We hope that this survey can provide valuable insights for researchers from diverse backgrounds and facilitate the future innovation of this  interdisciplinary realm.

\section*{Acknowledgements}
This research is supported by the National Research Foundation, Singapore under its AI Singapore Programme (AISG Award No: AISG2-PhD/2022-01-034[T]).

  \bibliographystyle{elsarticle-num}
 \bibliography{reference}

\clearpage
\newpage


\section*{Supplementary material}

# A Survey of Artificial Intelligence in Gait-Based Neurodegenerative Disease Diagnosis – Appendix

Haocong Rao[a,b], Minlin Zeng[a,b], Xuejiao Zhao[a,b], Chunyan Miao[a,b,*]

[a]*Joint NTU-UBC Research Centre of Excellence in Active Living for the Elderly (LILY), Nanyang Technological University, Singapore*
[b]*College of Computing and Data Science, Nanyang Technological University, 50 Nanyang Avenue, 639798, Singapore*

**Appendix Outline**

The overview for this appendix is presented as follows.

- In Sec. 1, we detail the survey strategy and process, including the inclusion and exclusion criteria (see Sec. 1.1), query phrases (see Sec. 1.2), screening and data abstraction (see Sec. 1.3).

- In Sec. 2, we provide a full description for potential disease-related gaits.

- In Sec. 3, we conduct a complementary elaboration on existing gait data types and their collecting technologies.

- In Sec. 4, we offer the lists of all included papers and representative excluded papers.

## 1. Survey Strategy and Process

### 1.1. Inclusion and Exclusion Criteria

For the definition of NDs and related concepts, we follow the standard in "International Classification of Diseases 11[th] Revision (ICD-11)" released by the World Health Organization (WHO)[1]. We search and gather papers information from three well-established literature sources, namely *PubMed*, *Web of Science (WoS)*, and *Google Scholar*.

We provide the detailed inclusion and exclusion criteria in Table 1. From the papers surveyed, we include those containing keywords associated with: "pathological gaits", "neurodegenerative

---

*Corresponding author
Email addresses:* haocong001@ntu.edu.sg (Haocong Rao), minlin001@ntu.edu.sg (Minlin Zeng), xjzhao@ntu.edu.sg (Xuejiao Zhao), ascymiao@ntu.edu.sg (Chunyan Miao)
[1]International Classification of Diseases 11[th] Revision (ICD-11)

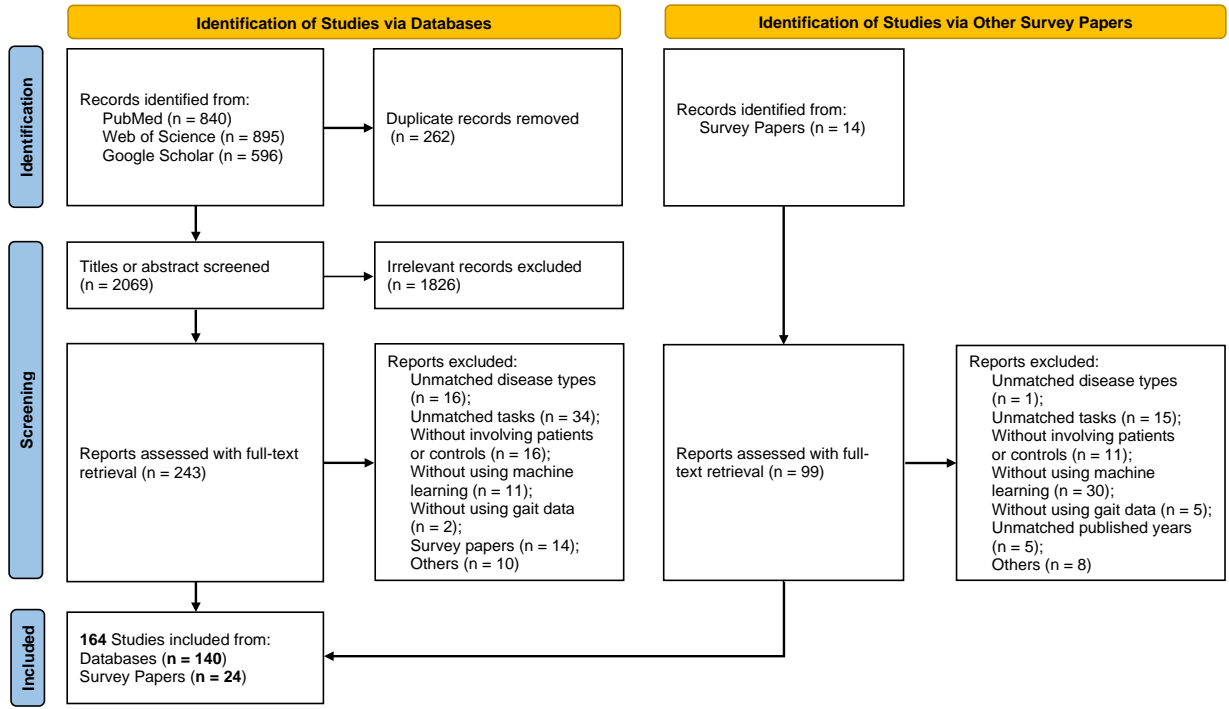

Figure 1: Overview of the literature selection process based on Preferred Reporting Items for Systematic Reviews and Meta-Analyses (PRISMA) flow diagrams.

disease diagnosis", and "artificial intelligence". Meanwhile, we exclude papers that focus on other topics such as "therapy", "treatment", "intervention", and "prevention". To provide a comprehensive review of recent advances in this field, we restrict the publication time range of the papers to between Jan 1, 2012, and September 1, 2023. The language and the studied species of the papers are restricted to English and human, respectively. The example search keywords of PubMed is presented in Table 2, and we provide the concrete query phrases for PubMed and WoS searching in Sec. 1.2.

Table 1: Inclusion and exclusion criteria for the survey.

| No. | Category of Criteria | Inclusion Criteria | Exclusion Criteria |
| --- | --- | --- | --- |
| 1 | Types of Diseases | Studies focused on five most common neurodegenerative diseases (NDs) (AD, PD, ALS, HD, MS) associated with gaits | Studies focused on other diseases instead of five NDs, or NDs unrelated to gaits |

| | | | |
|---|---|---|---|
| 2 | Task of the Study | Studies focused on the diagnosis (including prediction, classification or identification) of gait-related NDs | Studies focused on the classification and identification of gait-related information rather than the diagnosis of diseases. Some excluded examples as shown as follows: gait recognition/gait identification; (focused on identities, ages, genders, etc.); gait prediction (focused on motion prediction); gait simulation/estimation/detection; gait collection techniques; gait abnormality assessment (without health controls); gait phase/cycle classification/identification; gait-based fall detection; gait-based fatigue detection; Freezing of gait (FoG) detection (focused on FoG in Parkinson's patients without healthy controls); development/treatments/therapy for diseases; disease severity assessment; disease management/intervention/prevention; disease related operation |
| 3 | Artificial Intelligence (AI) Techniques | 1. Conventional Machine Learning (C-ML) Models: Decision Trees, Support Vector Machines, Linear Regression, Naive Bayes, K-Nearest Neighbors, Random Forest, 3-layer multilayer perceptrons, etc; 2. Conventional Deep Learning (C-DL) Models: Convolutional Neural Networks, Recurrent Neural Networks, vanilla long short-term memory, ResNet, etc; 3. Advanced Deep Learning (A-DL) Models: Graph Convolutional Networks, Transformers, Generative Adversarial Networks, etc. | Not use artificial intelligence approaches, or only use pure statistical methods, such as: Rule-based systems, Linear mixed-effect model, ANOVA |
| 4 | Experimental Sample Size | The size of samples (including health controls and patients) is not less than 10 | Sample size less than 10, or only contain patients without health controls |
| 5 | Performance Metrics | Studies reporting accuracy-related performance metrics (e.g., precision, AUC) | Did not report accuracy-related performance metrics, or only report hypothesis testing results (e.g.,significance) |

| 6 | Article Type | Published peer-reviewed articles: Original research or Structured reviews of the literature reported in accordance with PRISMA guidelines | Unpublished articles, articles without formal peer reviews, or not published in the English language |
|---|---|---|---|
| 7 | Species of Samples | Human | Different animals except human |
| 8 | Publication Year | Articles published between 2012 to 2023 | Articles published before 2012 |

*1.2. Query Phrases*

**PubMed Query Phrases Using Advanced Search:** ("pathological gaits" OR "gait impairment" OR "abnormal gait" OR "diplegic gait" OR "hemiplegic gait" OR "neuropathic gait" OR "Parkinsonian gait" OR "gait abnormality" OR "gait disorders" OR "gait diagnosis" OR "Gait / classification"[Mesh] OR "gait classification" OR "gait abnormality classification" OR "pathological gait classification" OR "gait disease diagnosis" OR "gait disease prediction" OR "gait disease classification" OR "gait disease identification" OR "gait disease assessment") AND ("deep learning"[Mesh] OR "artificial intelligence"[Mesh] OR "machine learning"[Mesh] OR "artificial intelligence" OR "deep learning" OR "machine learning" OR "Support Vector Machine" OR "Transformer" OR "Graph Neural Network" OR "Bayesian" OR "Decision Trees" OR "Fuzzy Logic" OR "Gradient Boosting" OR "k-means Clustering" OR "Nearest Neighbors" OR "Neural Networks" OR "random forests" OR "reinforcement learning") NOT ((Therapy[Title]) OR (Treatment[Title]) OR (Intervention[Title]) OR (Prevention[Title])))

**WoS Query Phrases Using Advanced Search:** TI = (((pathological gaits) OR (gait impairment) OR (abnormal gait) OR (diplegic gait) OR (hemiplegic gait) OR (neuropathic gait) OR (Parkinsonian gait) OR (gait abnormality) OR (gait disorders) OR (gait diagnosis) OR (gait classification) OR (gait abnormality classification) OR (pathological gait classification) OR (gait disease diagnosis) OR (gait disease prediction) OR (gait disease classification) OR (gait disease identification) OR (gait disease assessment)) AND ((deep learning) OR (artificial intelligence) OR (machine learning))) OR AB = (((pathological gaits) OR (gait impairment) OR (abnormal gait) OR (diplegic gait) OR (hemiplegic gait) OR (neuropathic gait) OR (Parkinsonian gait) OR (gait abnormality) OR (gait disorders) OR (gait diagnosis) OR (gait classification) OR (gait abnormality classification) OR (pathological gait classification) OR (gait disease diagnosis) OR (gait disease prediction) OR (gait disease classification) OR (gait disease identification) OR (gait disease assessment)) AND ((deep learning) OR (artificial intelligence) OR (machine learning))) NOT TI = ((therapy) OR (treatment) OR (intervention) OR (prevention)) NOT SO=CLINICAL REHABILITATION NOT WC= Rehabilitation AND PY=(2012-2023)

With the aforementioned search, inclusion, and exclusion strategies, we first select the articles via screening their titles and abstracts, and then perform full-text assessment on the selected articles to ensure that they match the requirement of this review. It is worth noting that we adopt different inclusion and exclusion guidelines and several quality evaluation constraints as shown in Appendix B. For example, the included papers are limited to those using ML/DL models for NDs

Table 2: Search strategy and main inclusion/exclusion criteria used in the literature search from PubMed. We manually filter and screen the searched articles according the disease and time range domains. The concrete query phrases for PubMed and WoS databases are shown in Sec. 1.1.

| Domain | Search Keywords |
| --- | --- |
| Gait | ("pathological gaits" OR "gait impairment" OR "abnormal gait" OR "diplegic gait" OR "hemiplegic gait" OR "neuropathic gait" OR "Parkinsonian gait" OR "gait abnormality" OR "gait disorders" OR "gait diagnosis" OR "Gait/classification"[Mesh] OR "gait classification" OR "gait abnormality classification" OR "pathological gait classification" OR "gait disease diagnosis" OR "gait disease prediction" OR "gait disease classification" OR "gait disease identification" OR "gait disease assessment") AND |
| Technology | ("deep learning"[Mesh] OR "artificial intelligence"[Mesh] OR "machine learning"[Mesh] OR "artificial intelligence" OR "deep learning" OR "machine learning" OR "Support Vector Machine" OR "Transformer" OR "Graph Neural Network" OR "Bayesian" OR "Decision Trees" OR "Fuzzy Logic" OR "Gradient Boosting" OR "k-means Clustering" OR "Nearest Neighbors" OR "Neural Networks" OR "random forests" OR "reinforcement learning") |
| Excluded Type | NOT ((Therapy[Title]) OR (Treatment[Title]) OR (Intervention[Title]) OR (Prevention[Title]))) |
| Disease | (Manually filter) Include "Parkinson's disease" OR "Alzheimer's disease" OR "Amyotrophic lateral sclerosis" OR "Huntington's disease" |
| Time range | (Manually filter) Published from January 1, 2012 to September 1, 2023 |

diagnosis, excluding those using only rule-based or mathematical analysis method. All included papers conduct experiments with both healthy controls and NDs patients. Papers are excluded if the sample size is less than 10. For included papers, we report the experimental results of the best-performing AI models, using commonly-used evaluation metrics (*e.g.*, accuracy, area under the curve (AUC)).

*1.3. Screening and Data Abstraction*

The systematic literature selection process is conducted according to the criteria of the Preferred Reporting Items for Systematic Reviews and Meta-Analyses (PRISMA), as shown in Fig. 1. The three phases of PRISMA can be summarized as follows:

- **Identification:** Firstly, different authors conduct the preliminary literature search from PubMed, WoS, and Google Scholar independently, and identify all potentially-matched articles (2069 articles) after discarding duplicate papers.

- **Screening:** Then, we select relevant papers via screening their titles and abstracts according to the pre-defined inclusion and exclusion criteria. The selected articles (243 articles), whose titles and abstracts matched the inclusion criteria, are further reviewed through a full-text assessment.

- **Inclusion:** Finally, we review the full text of the screened articles and exclude the unmatched articles. We record the key contents of all included articles (164 articles), including title, authors, year, country, organization, journal/book, name of disease, data type, AI model, number of samples, etc., which are summarized in our paper.

## 2. Potential Disease-Related Gaits

- **Myopathic gait** is sometimes called a waddling gait and characterized by a side-to-side movement. It is often a result of the weakness in the pelvic area. Myopathic gait may be a symptom of muscular dystrophy, muscle disease, or spinal muscle weakness.

- **Ataxic gait** is known for staggering movements during walking. A person with such gait tends to wobble from side to side and is unable to walk in a straight line. Additionally, their balance while standing could be affected as well, leading to a swaying motion. Ataxic gait can be a symptom of alcohol intoxication or a sign of brain injury.

- **Hemiplegic gait** is an abnormal gait that affects one side of the body, *i.e.*, one side of the arm does not move and the leg on the same side needs to be dragged in a semi-circle to be brought forward during walking. A hemiplegic gait is often the result of a stroke.

- **Diplegic gait** affects both sides of the body and is characterized by stiff, slightly flexed hips and knees, with ankles turned inward. The gait often involves a scissor-like movement of the legs and tiptoe walking, resulting in shorter steps. A diplegic gait can be a result of cerebral palsy, stroke, or head trauma.

- **Sensory gait** occurs when a person is unable to receive proprioceptive information, which is necessary to discern the position of their legs. As a result, individuals often lift their feet higher than usual and then slam them down hard onto the ground to sense their location. This gait can be seen in disorders with damaged dorsal columns (*e.g.*, B12 deficiency or tabes dorsalis) or in diseases affecting the peripheral nerves (*e.g.*, uncontrolled diabetes).

## 3. Gait Data Taxonomy

We present the taxonomy of all collected gait data types by categorizing them into four main classes based on their collection modalities. We elaborate on the common techniques used in each modality, along with their corresponding merits and demerits. At the end, we summarize by discussing the benefits and limitations of using these four different modalities in various applications.

- **Sensor Modality**: Advancement in sensor technologies allows its application in various tasks such as action recognition, event detection, kinematic and kinetic parameter estimation, gait classification, *ect*. The sensors explored in the collected papers include force sensors[1–4], inertial sensors, some hybrid sensors comprise of different kinds of sensors[5–8] as well as other less prevalent sensors. Depending on the specific sensor type, they offer various advantages, including portability, low cost, and miniaturization.[9].

  1. **Force Sensor**: There are mainly two types of force or pressure capturing approaches, in the filed of NDs classification based on gait, that are used to collect the foot-ground interaction data during locomotion. The first is to use non-wearable floor-mounted devices (*e.g*, force plates, pressure platform) with either load cells or force sensitive resistors (FSRs) where one measures the ground reaction force (GRF) and moments in three dimensions: vertical, anteroposterior (forward-backward), and mediolateral (side-to-side), and the other obtains detailed plantar pressure profiles, respectively[9]. In a laboratory setting, either load cells or FSRs would be sufficient for precise data collection when integrated into some stationary floor-mounted devices. However, since those devices are often bulky, operating-costly, and expensive, it's not often clinically

applicable. Aiming at solving these constrains, researchers develop the second approach by merging the load cells and/or FSRs into some small wearable devices (*e.g.*, insole) [10, 11]. Although these wearable sensors have less accuracy and reliability compare with the floor-mounted ones, due to its portable features and cheap installment, its application in the field has became a popular choice and risen significantly in the past 2 decades.

2. **Inertial Sensor**: Also termed as Inertial Measurement Unit (IMU), is an amalgamation of a triaxial accelerometer, gyroscope, or magnetometer, each providing critical data on three-dimensional acceleration, angular velocity, and the ambient magnetic field's magnitude and direction, respectively[9, 12]. It is often attached to limb segments (*e.g*, foot, shank, hip)[13–15] or embedded in wearable devices (*e.g.*, shoes, smartphone, smartwatch) to capture dynamic motion of the subject[16–21]. These sensors or devices, depending on their placement locations, record the repeating pattern of gait signals generated by the locomotion of the subjects. These signals will then be interpreted with various algorithms for the intended tasks, including gait event detection, kinetic or kinematic parameters estimation, and gait classification.

3. **Multi-Sensor**: These sensors typically combine various types of sensors for a more comprehensive data collection. Among the surveyed papers, studies such as [22–24] utilize both IMU sensors and FSRs for gait data collection, while [25] employs both EEG sensors and FSRs. These combinations harvest the strengths of each sensor type for a more comprehensive understanding of gait characteristic hence help enhancing the reliability and accuracy of gait event detection and classification.

4. **Other Sensors**: Sensors included in this category are less prevalent and exploited among collected papers. However, these sensors still play an important role in providing a more comprehensive understanding of the human gait. For instance, *EEG (electroencephalogram) Sensor* is a type of sensor used to measure the electrical activity of the brain using small, flat metal discs attached to the scalp. Its signals contain opulent information regarding brain's functional process[26]. In the context of gait disorder detection, EEG data can be used to analyze brain activity patterns that may correlate with specific gait abnormalities[25, 27]. Researchers have been using EEG data to aid the diagnosis, early detection, and classification of NDs (*e.g.*, AD) in several studies[28, 29]. Besides, *EMG (Electromyography) Sensor* is used to measure the electrical activity produced by muscles during contraction and rest[30]. These muscle cells activity is controlled by the nerve cells called motor neurons. Since NDs patients (*e.g.*, PD) suffer from the degeneration of nerve cells, they manifest significantly different muscle electrical pattern with various disease severity as well as from healthy subjects[12]. Therefore, their EMG data can be utilized to analyze muscle activation patterns during walking and hence aid in the diagnostic of various NDs such as PD[30].

- **Vision Modality**: The vision modality employs optoelectronic motion capture (Mocap) systems to capture the locomotion of subjects during walking. It's used to mimic the clinical examination which purely depends on the physicians' naked eyes. However, due to the con-

7

strains of human eyes, using a more robust camera-based system for gait diagnosis is more practical. The vision modality uses various types of cameras (*e.g.*, Analog, Smartphone[31, 32], Digital Single Lens Reflex (DSLR) [33], and Infrared (IR) Camera[34]) to estimate human gait. Depends on the precision requirement of the video data, the vision modality could be further divided into two sub-categories namely marker-based and marker-less Mocap systems[35].

1. **Marker-Based**: This approach is developed in an early stage when the motion capture (Mocap) system are not advanced enough for precise motion capturing. Therefore, these Mocap systems (*e.g.*, Vicon, IR Cameras, etc) use reflective markers which can be placed on the body landmarks and used as a guide for the camera to precisely capture target's movement. The model of human body is constructed after for feature extraction. The advantages of using the marker-based approach are that is highly accurate and provides exact location of the body landmark. These advantages hence make this modality as a gold standard in the field for collecting the most precise gait data. However, this method requires attaching markers to the body, leading to additional expenses and the potential to alter the gait, high setup costs, a lack of clinical friendliness. [36–38].

2. **Marker-Free**: Due to the limitation of the marker-based approach mentioned above and the development of the camera technology, the marker-free approach comes into appearance in the mid-ninety. This approach capture the gait video of the target without attaching nay markers on the subject's body. Then the vision related statistical information and other biomechanical features (*e.g.*, GEIs (Gait Energy Images)[39] or Optical Flow[40]) are derived from the video clips. Moreover, The application of smartphone technology, although still in its infancy, has also given a new insight towards this field since it's has all the benefits of this modality (*e.g.*, highly compact, portable, no requirement of lab environment, becoming cheaper, fast advancement, etc). However, since it's still in its early developing stage, it also suffers from disadvantages such as lower accuracy, etc.

- **Sensor+Vision Modality**: This modality could also be referred to as multi-modal modality. In comparison with single-modal approaches, although the latter ones have their own merits and show some promising results, combining sensor and vision modality unleashes a bigger potential, such as providing richer data and more comprehensive features, capturing more subtle changes in gait, improving recognition performance[27, 41, 42]. In recent years, the application of Microsoft Kinect system has gain a tremendous amount of attention from many authors in the field due to its utilization of both depth sensor, which contains an infrared laser projector and an infrared camera, and RGB camera. This integration enables a more accurate estimation and calculations for the gait parameters while not sacrificing the advantages mentioned in the marker-less vision modality. More specifically, using only the conventional camera to estimate skeleton coordinates usually produce coarse results. However, when employing the depth sensor, the spatiotemporal features of subjects can be derived from the precise coordinates of skeleton joints obtained using it. This additional spa-

tial information, which is not available in the marker-less vision approach, provides a more comprehensive description of the gait. In addition, the precision and efficacy of using Kinect has been verified in many studies which compares the result with the data obtained using gold standard (Marker-Based vision) approach. Nevertheless, one of the short comings of using the depth sensor is that its precision reduces as the subject walks further away from the camera, resulting in a width and length-limited data collection space[43–46]. Besides the RGB-D camera approach, there are also other combinations of these two modalities. For example, You et al. combined vision techniques with EEG sensors in their study [27]. Similarly, Tahir et al. fused a marker-based IF camera and a force-sensitive platform [47], while Chatzaki et al. utilized a conventional camera, IMU, and a force platform [42]. Additionally, Zhao et al. merge some vGRF and IMU sensors with Kinect[41].

- **Other Modalities**: Trabassi et al. use electronic medical record (EMR) which contains various patient-related features and medical history data that are indicative of early PD symptoms or risk factors as well as the first diagnosis of gait or tremor disorders to create prediction models.[48]

In a nutshell, despite having so many benefits of using sensor technology to directly obtain NDs gait, it still suffers from certain limitations such as high cost, time, power, and personnel for the non-wearable stationary sensors and wearing discomfort, drifting effect, and potential gait altering problem for the wearable ones. These limitations thus make it unsuitable for obtaining rich gait parameters in a precise and clinic-friendly manner. On the other hand, the vision based modality involved using highly-precise Mocap systems and is considered as the gold standard for data collection. However, it too has some flaws such as costly to operate, necessity of lab environment, requirement of high proficiency, *ect*. However, the combined sensor and vision approach utilizes both modalities' merits without sacrificing its performance too much. It also provides a more comprehensive view of the human gait data by considering both the spatiotemporal data and the overall skeleton joints correlations of the whole body. Thanks to these advantages, we further highlighted the benefit of using skeleton data to aid in NDs diagnosis in the research vision section.

## 4. Supplementary lists of Included and Excluded Papers

Table 3: List of included survey papers (N = 14).

| No. | First Author | Title | Organization | Journal/Conference | Year | Country | Database | DOI |
|---|---|---|---|---|---|---|---|---|
| 1 | Navleen Kour | [33] | Shri Mata Vaishno Devi University | EXPERT SYSTEMS | 2022 | India | Web of Science | 10.1111/exsy.12955 |
| 2 | Preeti Khera | [49] | Academy of Scientific & Innovative Research (AcSIR) | ENGINEERING SCIENCE AND TECHNOLOGY-AN INTERNATIONAL JOURNAL-JESTECH | 2022 | India | Web of Science | 10.1016/j.jestch.2021.05.009 |
| 3 | Luay Fraiwan | [50] | Abu Dhabi University | PLOS ONE | 2021 | Abu Dhabi | Web of Science | 10.1371/journal.pone.0252380 |

9

| No. | First Author | Title | Organization | Journal/Conference | Year | Country | Database | DOI |
|---|---|---|---|---|---|---|---|---|
| 4 | NAVLEEN KOUR | [35] | Shri Mata Vaishno Devi University | IEEE ACCESS | 2019 | India | Web of Science | 10.1109/ACCESS.2019.2949744 |
| 5 | Navleen Kour | [12] | Shri Mata Vaishno Devi University | MULTIMEDIA TOOLS AND APPLICATIONS | 2022 | India | Web of Science | 10.1007/s11042-022-13398-7 |
| 6 | di Biase L | [51] | Università Campus Bio-Medico di Roma | Sensors (Basel) | 2020 | Italy | PubMed | 10.3390/s20123529 |
| 7 | Pardoel S | [52] | University of Waterloo | Sensors (Basel) | 2019 | Canada | PubMed | 10.3390/s19235141 |
| 8 | Gupta R | [53] | Delhi Technological, University | Ageing Res Rev | 2023 | USA | PubMed | 10.1016/j.arr.2023.102013 |
| 9 | Figueiredo J | [54] | University of Minho | Med Eng Phys | 2018 | Portugal | Google Scholar | 10.1016/j.medengphy.2017.12.006 |
| 10 | Vienne A | [55] | Université Paris Descartes | Front Psychol | 2017 | France | Google Scholar | 10.3389/fpsyg.2017.00817 |
| 11 | Das R | [9] | North-Eastern Hill University | Frontier in Neuroscience | 2022 | India | Google Scholar | 10.3389/fnins.2022.859298 |
| 12 | Salchow-Hömmen | [56] | Charite University Berlin | Frontiers in Human Neuroscience | 2022 | German | Google Scholar | 10.3389/fnhum.2022.768575 |
| 13 | Hui Wen Loh | [57] | Singapore University of Social Sciences (SUSS) | SENSORS | 2021 | Singapore | Web of Science | 10.3390/s21217034 |
| 14 | Zainab Ayaz | [58] | University of Peshawar | NEURAL COMPUTING & APPLICATIONS | 2021 | Pakistan | Web of Science | 10.1007/s00521-021-06626-y |

Table 4: List of all included papers (N=164).

| No. | First Author | Title | Organization | Journal/Conference | Year | Country | Database | DOI | Disease |
|---|---|---|---|---|---|---|---|---|---|
| 1 | Abir Alharbi | [59] | King Saud University | INTERNATIONAL JOURNAL OF COMPUTER MATHEMATICS | 2020 | Saudi Arabia | Web of Science | 10.1080/00207160.2019.1607842 | PD |
| 2 | Domenico Buongiorno | [43] | Politecnico di Bari | BMC MEDICAL INFORMATICS AND DECISION MAKING | 2019 | Italy | Web of Science | 10.1186/s12911-019-0987-5 | PD |
| 3 | Andrea Mannini | [15] | Scuola Superiore Sant'Anna | SENSORS | 2016 | Italy | Web of Science | 10.3390/s16010134 | HD (and Post Stroke) |
| 4 | Armin Salimi-Badr | [60] | Shahid Beheshti University | APPLIED INTELLIGENCE | 2022 | Iran | Web of Science | 10.1007/s10489-022-04276-8 | PD |
| 5 | Juliana Paula Felix | [61] | Universidade Federal de Goias | 2019 IEEE 31ST INTERNATIONAL CONFERENCE ON TOOLS WITH ARTIFICIAL INTELLIGENCE (ICTAI 2019) | 2019 | Brazil | Web of Science | 10.1109/ICTAI.2019.00243 | HD |
| 6 | Juliana Paula Felix | [62] | Universidade Federal de Goias | 2020 IEEE INTERNATIONAL CONFERENCE ON BIOINFORMATICS AND BIOMEDICINE | 2020 | Brazil | Web of Science | 10.1109/BIBM49941.2020.9313308 | ALS |
| 7 | Carlos Fernandes | [63] | Universidade do Minho | PROCEEDINGS 2018 IEEE INTERNATIONAL CONFERENCE ON BIOINFORMATICS AND BIOMEDICINE (BIBM) | 2018 | Portugal | Web of Science | 10.1109/BIBM.2018.8621466 | PD |
| 8 | Balaji, E | [64] | PSG College Technology | APPLIED SOFT COMPUTING | 2021 | India | Web of Science | 10.1016/j.asoc.2021.107463 | PD |

10

| # | Author | Ref | Affiliation | Venue | Year | Country | Database | DOI | Disease |
|---|---|---|---|---|---|---|---|---|---|
| 9 | Felix, Juliana Paula | [65] | Universidade Federal de Goias | 2021 IEEE 45TH ANNUAL COMPUTERS, SOFTWARE, AND APPLICATIONS CONFERENCE (COMPSAC 2021) | 2021 | Brazil | Web of Science | 10.1109/COMPSAC51774.2021.00295 | ALS |
| 10 | Tuan D. Pham | [66] | Prince Mohammad Bin Fahd University | INTERNATIONAL CONFERENCE ON ELECTRICAL, COMPUTER AND ENERGY TECHNOLOGIES (ICECET 2021) | 2021 | Saudi Arabia | Web of Science | 10.1109/ICECET52533.2021.9698524 | PD |
| 11 | Yi Xia | [1] | Anhui University | BIOMEDICAL SIGNAL PROCESSING AND CONTROL | 2015 | China | Web of Science | 10.1016/j.bspc.2015.02.002 | ALS, PD, HD |
| 12 | Ferdous Wahid | [67] | University of Melbourne | IEEE JOURNAL OF BIOMEDICAL AND HEALTH INFORMATICS | 2015 | Australia | Web of Science | 10.1109/JBHI.2015.2450232 | PD |
| 13 | Helber Andrés Carvajal-Castaño | [13] | Universidad de Antioquia | ELECTRONICS | 2022 | Colombia | Web of Science | 10.3390/electronics11172684 | PD |
| 14 | Rehman, Rana Zia Ur | [68] | Newcastle University | SENSORS | 2019 | UK | Web of Science | 10.3390/s19245363 | PD |
| 15 | Khoury, Nicolas | [69] | Universite Paris-Est-Creteil-Val-de-Marne (UPEC) | SENSORS | 2019 | France | Web of Science | 10.3390/s19020242 | PD |
| 16 | Balaji E | [70] | PSG College Technology | MEDICAL ENGINEERING & PHYSICS | 2021 | India | Web of Science | 10.1016/j.medengphy.2021.03.005 | PD |
| 17 | Imanne El Maachi | [71] | Universite de Montreal | EXPERT SYSTEMS WITH APPLICATIONS | 2020 | Canada | Web of Science | 10.1016/j.eswa.2019.113075 | PD |
| 18 | Navita Mehra | [72] | Maharshi Dayanand University | INTERNATIONAL JOURNAL OF ADVANCED COMPUTER SCIENCE AND APPLICATIONS | 2022 | India | Web of Science | | PD |
| 19 | James R. Williamson | [73] | Lincoln Laboratory; Massachusetts Institute of Technology (MIT) | SENSORS | 2021 | USA | Web of Science | 10.3390/s21062047 | PD |
| 20 | Alex Li | [74] | Stanford University | DIAGNOSTICS | 2022 | USA | Web of Science | 10.3390/diagnostics12102404 | PD |
| 21 | Mirelman, Anat | [75] | Teva Pharmaceutical Industries | MOVEMENT DISORDERS | 2021 | Israel | Web of Science | 10.1002/mds.28631 | PD |
| 22 | Nguyen, Quoc Duy Nam | [76] | National Cheng Kung University | ENTROPY | 2020 | Taiwan, China | Web of Science | 10.3390/e22121340 | PD, HD, ALS |
| 23 | Juliana Paula Felix | [61] | Universidade Federal de Goias | ADVANCES IN VISUAL COMPUTING, ISVC 2019, PT II | 2019 | Brazil | Web of Science | 10.1007/978-3-030-33723-0_41 | HD |
| 24 | Che-Wei Lin | [77] | National Cheng Kung University | SENSORS | 2020 | Taiwan, China | Web of Science | 10.3390/s20143857 | PD, HD, ALS |
| 25 | Yi-Wei Ma | [78] | National Taiwan University of Science & Technology | SOFT COMPUTING | 2021 | Taiwan, China | Web of Science | 10.1007/s00500-021-06170-w | PD |
| 26 | Taiki Ogata | [79] | Tokyo Institute of Technology | FRONTIERS IN PHYSIOLOGY | 2022 | Japan | Web of Science | 10.3389/fphys.2022.726677 | PD |
| 27 | Christian Urcuqui | [80] | Universidad ICESI | 2018 14TH INTERNATIONAL CONFERENCE ON SEMANTICS, KNOWLEDGE AND GRIDS (SKG) | 2018 | Colombia | Web of Science | 10.1109/SKG.2018.00029 | PD |
| 28 | Shyam V. Perumal | [8] | State University System of Florida | ICT EXPRESS | 2016 | USA | Web of Science | 10.1016/j.icte.2016.10.005 | PD |

| # | Author | Ref | Institution | Journal/Conference | Year | Country | Database | DOI | Disease |
|---|---|---|---|---|---|---|---|---|---|
| 29 | Enas Abdulhay | [81] | Jordan University of Science & Technology | FUTURE GENERATION COMPUTER SYSTEMS-THE INTERNATIONAL JOURNAL OF ESCIENCE | 2018 | Jordan | Web of Science | 10.1016/j.future.2018.02.009 | PD |
| 30 | Abdullah S. Alharthi | [59] | University of Manchester | IEEE SENSORS JOURNAL | 2021 | UK | Web of Science | 10.1109/JSEN.2020.3018262 | PD |
| 31 | Juan C. Pérez-Ibarra | [22] | Universidade de Sao Paulo | IEEE SENSORS JOURNAL | 2020 | USA | Web of Science | 10.1109/JSEN.2020.3011627 | PD |
| 32 | Febryan Setiawan | [82] | National Cheng Kung University | BRAIN SCIENCES | 2021 | Taiwan, China | Web of Science | 10.3390/brainsci11070902 | PD, HD, ALS |
| 33 | Carlotta Caramia | [14] | Roma Tre University | IEEE JOURNAL OF BIOMEDICAL AND HEALTH INFORMATICS | 2018 | Italy | Web of Science | 10.1109/JBHI.2018.2865218 | PD |
| 34 | Andrew P. Creagh | [83] | University of Oxford | SCIENTIFIC REPORTS | 2021 | UK | Web of Science | 10.1038/s41598-021-92776-x | MS |
| 35 | Rana Zia Ur Rehman | [84] | Newcastle University | FRONTIERS IN AGING NEUROSCIENCE | 2022 | UK | Web of Science | 10.3389/fnagi.2022.808518 | PD |
| 36 | Juan Felipe Reyes | [45] | Universidad ICESI | 2019 IEEE COLOMBIAN CONFERENCE ON COMMUNICATIONS AND COMPUTING (COLCOM 2019) | 2019 | Colombia | Web of Science | | PD |
| 37 | Rami Alkhatib | [85] | Rafik Hariri University | IEEE SENSORS LETTERS | 2020 | Lebanon | Web of Science | 10.1109/LSENS.2020.2994938 | PD |
| 38 | Elham Rastegari | [86] | University of Nebraska System | PROCEEDINGS OF THE 52ND ANNUAL HAWAII INTERNATIONAL CONFERENCE ON SYSTEM SCIENCES | 2019 | USA | Web of Science | | PD |
| 39 | Dante Trabassi | [87] | Sapienza University Rome | SENSORS | 2022 | Italy | Web of Science | 10.3390/s22103700 | PD |
| 40 | Xiuyu Huang | [88] | University of Sydney | NEURAL INFORMATION PROCESSING (ICONIP 2019), PT IV | 2019 | Australia | Web of Science | 10.1007/978-3-030-36808-1_66 | HD |
| 41 | Wenting Hu | [6] | Memorial University Newfoundland | BIOMEDICAL ENGINEERING ONLINE | 2022 | Canada | Web of Science | 10.1186/s12938-022-00992-x | MS |
| 42 | Munoz-Ospina, Beatriz | [46] | Fundacion Valle del Lili | FRONTIERS IN HUMAN NEUROSCIENCE | 2022 | Colombia | Web of Science | 10.3389/fnhum.2022.826376 | PD |
| 43 | Hamza Abujrida | [89] | Worcester Polytechnic Institute | BIOMEDICAL PHYSICS & ENGINEERING EXPRESS | 2020 | USA | Web of Science | 10.1088/2057-1976/ab39a8 | PD |
| 44 | Francisco Diego Acosta-Escalante | [90] | Universidad Juarez Autonoma de Tabasco | IEEE ACCESS | 2018 | Mexico | Web of Science | 10.1109/ACCESS.2018.2840327 | HD |
| 45 | Yajing Guo | [25] | Shanghai University | PROCEEDINGS OF 2019 IEEE 8TH JOINT INTERNATIONAL INFORMATION TECHNOLOGY AND ARTIFICIAL INTELLIGENCE CONFERENCE (ITAIC 2019) | 2019 | China | Web of Science | 10.1109/ITAIC.2019.8785586 | PD |
| 46 | Fabio Cuzzolin | [91] | Oxford Brookes University | GAIT & POSTURE | 2017 | UK | Web of Science | 10.1016/j.gaitpost.2017.02.012 | PD |
| 47 | Camilo Vasquez-Correa | [92] | Universidad de Antioquia | IEEE JOURNAL OF BIOMEDICAL AND HEALTH INFORMATICS | 2019 | Colombia | Web of Science | 10.1109/JBHI.2018.2866873 | PD |
| 48 | Alexandra-Georgiana Andrei | [93] | Polytechnic University of Bucharest | 2019 E-HEALTH AND BIOENGINEERING CONFERENCE (EHB) | 2019 | Romania | Web of Science | 10.1109/EHB47216.2019.8969942 | PD |
| 49 | Srivardhini Veeraragavan | [94] | Monash University | FRONTIERS IN PHYSIOLOGY | 2020 | Malaysia | Web of Science | 10.3389/fphys.2020.587057 | PD |

| # | Author | Ref | Institution | Journal | Year | Country | Source | DOI | Disease |
|---|---|---|---|---|---|---|---|---|---|
| 50 | Yang xiaoli | [95] | Guangdong University of Technology | IEEE JOURNAL OF TRANSLATIONAL ENGINEERING IN HEALTH AND MEDICINE | 2022 | China | Web of Science | 10.1109/JTEHM.2022.3180933 | PD |
| 51 | Rachneet Kaur | [7] | University of Illinois | IEEE TRANSACTIONS ON BIOMEDICAL ENGINEERING | 2021 | USA | Web of Science | 10.1109/TBME.2020.3048142 | MS |
| 52 | Luis C. Guayacan | [96] | Universidad Industrial de Santander | JOURNAL OF MEDICAL AND BIOLOGICAL ENGINEERING | 2022 | Colombia | Web of Science | 10.1007/s40846-022-00691-x | PD |
| 53 | S.A.Vajiha Begum | [97] | Mother Teresa Women's University | PROCEEDINGS OF THE INTERNATIONAL CONFERENCE ON INTELLIGENT COMPUTING AND CONTROL SYSTEMS (ICICCS 2020) | 2020 | India | Web of Science | 10.1109/ICICCS48265.2020.9120920 | PD, ALS, HD |
| 54 | Rana Zia Ur Rehman | [98] | Newcastle University | SCIENTIFIC REPORTS | 2019 | UK | Web of Science | 10.1038/s41598-019-53656-7 | PD |
| 55 | A.P. Creagh | [83] | University of Oxford | IEEE JOURNAL OF BIOMEDICAL AND HEALTH INFORMATICS | 2021 | UK | Web of Science | 10.1109/JBHI.2020.2998187 | MS |
| 56 | MARÍA GOÑI | [20] | Helmholtz Association | IEEE ACCESS | 2022 | Germany | Web of Science | 10.1109/ACCESS.2022.3156659 | PD |
| 57 | Hamza Abujrida | [21] | Worcester Polytechnic Institute | 2017 IEEE-NIH HEALTHCARE INNOVATIONS AND POINT OF CARE TECHNOLOGIES (HI-POCT) | 2017 | China | Web of Science | 10.1109/HIC.2017.8227621 | PD |
| 58 | Trentzsch, Katrin | [99] | Technische Universitat Dresden | BRAIN SCIENCES | 2021 | Germany | Web of Science | 10.3390/brainsci11081049 | MS |
| 59 | Md Nafiul Alam | [5] | University of North Dakota Grand Forks | PLOS ONE | 2017 | USA | Web of Science | 10.1371/journal.pone.0175951 | PD |
| 60 | Sathya Bama, B | [100] | Sathyabama Institute of Science & Technology | HEALTH SYSTEMS | 2022 | India | Web of Science | 10.1080/20476965.2022.2125838 | PD |
| 61 | Younghoon Jeon | [101] | Korea National University of Transportation | IEEE SENSORS JOURNAL | 2023 | Korea | Web of Science | 10.1109/JSEN.2023.3259034 | AD |
| 62 | P Divyashree | [102] | Indian Institute of Information Technology | IEEE TRANSACTIONS ON COMPUTATIONAL SOCIAL SYSTEMS | 2023 | India | Web of Science | 10.1109/TCSS.2022.3224046 | PD |
| 63 | Tunç Aşuroğlu | [103] | Tampere University | HEALTH AND TECHNOLOGY | 2022 | Finland | Web of Science | 10.1007/s12553-022-00698-z | PD |
| 64 | Olmos J | [104] | Biomedical Imaging, Vision and Learning Laboratory | Annu Int Conf IEEE Eng Med Biol Soc | 2022 | Colombia | PubMed | 10.1109/EMBC48229.2022.9871206 | PD |
| 65 | Yuan W | [48] | Harvard Medical School | BMC Neurol | 2021 | USA | PubMed | 10.1186/s12883-021-02226-4 | PD |
| 66 | Zhao A | [41] | Qingdao University | IEEE Trans Cybern | 2022 | China | PubMed | 10.1109/TCYB.2021.3056104 | ALS, HD, and PD |
| 67 | You Z | [27] | Shenzhen People's Hospital | Front Public Health | 2020 | China | PubMed | 10.3389/fpubh.2020.584387 | ALS |
| 68 | Zhang S | [105] | The University of Sydney | Stud Health Technol Inform | 2019 | Australia | PubMed | 10.3233/SHTI190267 | HD |
| 69 | Hughes JA | [106] | St. Francis Xavier University | IEEE J Biomed Health Inform | 2020 | Canada | PubMed | 10.1109/JBHI.2019.2961808 | PD |
| 70 | Ingelse L | [107] | Universidade de Lisboa | Sensors (Basel) | 2022 | Portugal | PubMed | 10.3390/s22113980 | PS, PD |
| 71 | Xia Y | [108] | Anhui University | IEEE Trans Neural Syst Rehabil Eng | 2020 | China | PubMed | 10.1109/TNS.2019.2946194 | PD |
| 72 | Filtjens B | [109] | KU Leuven | J Neuroeng Rehabil | 2022 | Belgium | PubMed | 10.1186/s12984-022-01025-3 | PD |

13

| # | Author | Ref | Institution | Journal | Year | Country | Source | DOI | Disease |
|---|---|---|---|---|---|---|---|---|---|
| 73 | Mileti I | [23] | Sapienza University of Rome | Sensors (Basel) | 2018 | Italy | PubMed | 10.3390/s18030919 | PD |
| 74 | Zeng W | [110] | Longyan University | Neurosci Lett | 2016 | China | PubMed | 10.1016/j.neulet.2016.09.043 | PD |
| 75 | Carvajal-Castaño HA | [111] | Universidad de Antioquia | Hum Mov Sci | 2022 | Colombia | PubMed | 10.1016/j.humov.2021.102891 | PD |
| 76 | Park H | [34] | Dong-A University | J Neuroeng Rehabil | 2021 | Korea | PubMed | 10.1186/s12984-021-00975-4 | PD |
| 77 | Rovini E | [112] | Viale Rinaldo Piaggio | Ann Biomed Eng | 2018 | Italy | PubMed | 10.1007/s10439-018-2104-9 | PD (and Idiopathic Hyposmia (IH)) |
| 78 | Seifallahi M | [113] | Florida Atlantic University | IEEE Trans Neural Syst Rehabil Eng | 2022 | USA | PubMed | 10.1109/TNSRE.2022.3181252 | AD |
| 79 | Pham TD | [114] | Linkoping University | IEEE Trans Biomed Eng | 2018 | Sweden | PubMed | 10.1109/TBME.2017.2779884 | PD |
| 80 | Filtjens B | [36] | KU Leuven | BMC Med Inform Decis Mak | 2021 | Belgium | PubMed | 10.1186/s12911-021-01699-0 | PD |
| 81 | Ťupa O | [44] | University of Chemistry and Technology in Prague | Biomed Eng Online | 2015 | Czech Republic | PubMed | 10.1186/s12938-015-0092-7 | PD |
| 82 | Shalin G | [115] | University of Waterloo | J Neuroeng Rehabil | 2021 | Canada | PubMed | 10.1186/s12984-021-00958-5 | PD |
| 83 | Costa L | [116] | University of Minho | Comput Intell Neurosci | 2016 | Portugal | PubMed | 10.1155/2016/3891253 | AD |
| 84 | Perez-Ibarra JC | [22] | University of São Paulo | IEEE Trans Neural Syst Rehabil Eng | 2020 | Brazil | PubMed | 10.1109/TNSRE.2020.3039999 | PD |
| 85 | Ricciardi C | [117] | University Hospital of Naples | Comput Methods Programs Biomed | 2019 | Italy | PubMed | 10.1016/j.cmpb.2019.105033 | PD |
| 86 | Som A | [118] | Arizona State University | Annu Int Conf IEEE Eng Med Biol Soc | 2020 | USA | PubMed | 10.1109/EMBC44109.2020.9176572 | PD |
| 87 | Khorasani A | [119] | Iran University of Science and Technology | J Med Syst | 2014 | Iran | PubMed | 10.1007/s10916-014-0147-5 | PD |
| 88 | Rehman RZU | [120] | Newcastle University | Sensors (Basel) | 2020 | UK | PubMed | 10.3390/s20185377 | PD |
| 89 | Aich S | [121] | Inje Universit | Sensors (Basel) | 2018 | Korea | PubMed | 10.3390/s18103287 | PD |
| 90 | Zeng W | [122] | Longyan University | Neural Netw | 2019 | China | PubMed | 10.1016/j.neunet.2018.12.012 | PD |
| 91 | Joshi D | [123] | Indian Institute of Technology | Comput Methods Programs Biomed | 2017 | Indian | PubMed | 10.1016/j.cmpb.2017.04.007 | PD |
| 92 | Nair P | [124] | IITB-Monash Research Academy and IIT Bombay | Annu Int Conf IEEE Eng Med Biol Soc | 2020 | India | PubMed | 10.1109/EMBC44109.2020.9175343 | PD (Drug induced) |
| 93 | Xia Y | [125] | Anhui University | Med Biol Eng Comput | 2016 | China | PubMed | 10.1007/s11517-015-1413-5 | ALS |
| 94 | Kugler P | [30] | Friedrich Alexander University of Erlangen-Nuremberg | Annu Int Conf IEEE Eng Med Biol Soc | 2013 | Germany | PubMed | 10.1109/EMBC.2013.6610865 | PD |
| 95 | Omid Mohamad Beigi | [126] | Brock University | Biosystems | 2023 | Canada | PubMed | 10.1016/j.biosystems.2023.105006 | PD |
| 96 | Hirotaka Uchitomi | [127] | Tokyo Institute of Technology | Scientific Reports | 2023 | Japan | PubMed | 10.1038/s41598-023-39862-4 | PD |
| 97 | Guoen Cai | [128] | Fujian Medical University Union Hospital | The Journals of Gerontology, Series A: Biological Sciences and Medical Sciences, | 2023 | China | PubMed | 10.1093/gerona/glad101 | PD |
| 98 | Yubo Sun | [129] | Nankai University | Mathematical Biosciences and Engineering | 2023 | China | PubMed | 10.3934/mbe.2023601 | PD |
| 99 | Zhang J | [31] | Sungkyunkwan University | Sensors (Basel) | 2023 | Korea | PubMed | 10.3390/s23104980 | PD |

14

| # | Author | Ref | Affiliation | Journal/Conference | Year | Country | Source | DOI | Disease |
|---|---|---|---|---|---|---|---|---|---|
| 100 | Dobromir Dotova | [130] | McMaster University | Journal of Motor Behavior | 2023 | Canada | PubMed | 10.1080/00222895.2023.2217100 | PD |
| 101 | Chenhui Dong | [131] | Changzhou University | IEEE TRANSACTIONS ON NEURAL SYSTEMS AND REHABILITATION ENGINEERING | 2023 | China | PubMed | 10.1109/TNSRE.2023.3269569 | PD |
| 102 | Chatzaki C | [42] | Hellenic Mediterranean University | Sensors (Basel) | 2022 | Greece | PubMed | 10.3390/s22249937 | PD |
| 103 | Marta Isabel A. S. N. Ferreira | [38] | Universidade do Porto | Gait Posture | 2022 | Portugal | PubMed | 10.1016/j.gaitpost.2022.08.014 | PD |
| 104 | Yonatan E. Brand | [132] | Tel Aviv Sourasky Medical Center | Sensors (Basel) | 2022 | Israel | PubMed | 10.3390/s22187094 | PD |
| 105 | Roozbeh Atri | [18] | Cohen Veterans Bioscience | Sensors (Basel) | 2022 | USA | PubMed | 10.3390/s22186831 | PD |
| 106 | Cristian Tobar | [133] | Campus de Tulcán | Biomedical Physics & Engineering Express | 2022 | Colombia | PubMed | 10.1088/2057-1976/ac8c9a | PD, ALS, HD |
| 107 | Aite Zhao, | [41] | Qingdao University | IEEE Transactions on Cybernetics | 2022 | China | PubMed | 10.1109/TCYB.2021.3056104 | PD |
| 108 | Y. Yan | [134] | Shenzhen Institutes of Advanced Technology, Chinese Academy of Sciences China | IEEE Access | 2020 | China | Google Scholar | 10.1109/ACCESS.2020.2996667 | ALS, HD, PD |
| 109 | Mannini, A | [15] | University of Sassari | Sensor 2016 | 2020 | Italy | Google Scholar | 10.1186/s12984-020-00728-9 | HD (and post-stroke) |
| 110 | Y. Mittra | [135] | Guru Gobind Singh Indraprastha University | 2018 International Conference on Automation and Computational Engineering (ICACE) | 2018 | India | Google Scholar | 10.1109/ICACE.2018.8687022 | PD |
| 111 | Moon S | [136] | Department of Physical Therapy, Ithaca College | J Neuroeng Rehabil | 2020 | USA | Google Scholar | 10.1186/s12984-020-00756-5 | PD (and essential tremor) |
| 112 | Tianben Wang | [137] | Northwestern Polytechnical University | ACM Trans. | 2016 | China | Google Scholar | 10.1145/2890511 | PD |
| 113 | Luis C. Guayacan | [40] | Universidad Industrial de Santander | Journal of Biomedical Informatics 123 (2021) 103935 | 2021 | Colombia | Google Scholar | 10.1016/j.jbi.2021.103935 | PD |
| 114 | Klomsae A | [138] | Chiang Mai University | Comput Intell Neurosci | 2018 | Thailand | Google Scholar | 10.1155/2018/1869565 | PD, HD, and ALS |
| 115 | Behnaz Ghoraani | [139] | Florida Atlantic University | Biomedical Signal Processing and Control 64 (2021) 102249 | 2020 | USA | Google Scholar | 10.1016/j.bspc.2020.102249 | AD (and mild cognitive impairment (MCI)) |
| 116 | FEBRYAN SETIAWAN | [140] | National Cheng Kung University | IEEE Access | 2022 | Chinese Taiwan | Google Scholar | 10.1109/ACCESS.2022.3158961 | ALS, HD, and PD |
| 117 | Mendoza Oscar | [141] | National Institute of Technology, Srinagar | Multimedia Tools and Applications | 2022 | Colombia | Google Scholar | 10.1007/s11042-022-12280-w | PD |
| 118 | L. Gong | [39] | University of Lincoln | 2020 IEEE Intl Conf on Dependable, Autonomic and Secure Computing, Intl Conf on Pervasive Intelligence and Computing, Intl Conf on Cloud and Big Data Computing, Intl Conf on Cyber Science and Technology Congress (DASC/ PiCom/ CBDCom/ CyberSciTech) | 2020 | UK | Google Scholar | 10.1109/DASC-PICom-CBDCom-CyberSciTech49142.2020.00045 | PD |
| 119 | Peyvand Ghaderyan | [142] | Sahand University of Technology | Measurement 177 (2021) 109249 | 2021 | Iran | Google Scholar | 10.1016/j.measurement.2021.109249 | PD |

| # | Author | Ref | Institution | Venue | Year | Country | Source | DOI | Disease |
|---|---|---|---|---|---|---|---|---|---|
| 120 | Patil Prithvi | [2] | National Institute of Technology, Srinagar | 2019 1st international conference on advances in science, engineering and robotics technology (ICASERT) | 2019 | India | Google Scholar | 10.1109/ICASERT.2019.8934463 | MS (and stroke, cerebral palsy (children)) |
| 121 | YILMAZ DERYA | [143] | Başkent University | EJONS International Journal on Mathematic, Engineering and Natural Sciences | 2020 | Turkey | Google Scholar | 10.38063/ejons.255 | PD, HD, or ALS |
| 122 | Khang Nguyen | [76] | Institute of Science and Information Technology | Research in Intelligent and Computing in Engineering. Advances in Intelligent Systems and Computing, vol 1254 | 2021 | Vietnam | Google Scholar | 10.1007/978-981-15-7527-3_56 | PD |
| 123 | Dutta Saibal | [144] | Heritage Institute of Technology | Advances in Heuristic Signal Processing and Applications | 2013 | India | Google Scholar | 10.1007/978-3-642-37880-5_12 | PD, HD, ALS |
| 124 | Klucken J | [17] | University Hospital Erlangen | PLoS ONE 8(2): e56956. | 2013 | Germany | Google Scholar | 10.1371/journal.pone.0056956 | PD |
| 125 | Tahir Nooritawati Md | [47] | Universiti teknologi MARA | 2012 Asain Network for Scientific Information | 2012 | Malaysia | Google Scholar | 10.3923/jas.2012 | PD |
| 126 | Manap Hany Hazfiza | [145] | Universiti teknologi MARA | 2013 European Modelling Symposium | 2013 | Malaysia | Google Scholar | 10.1109/EMS.2013.36 | PD |
| 127 | Niño Santiago | [146] | Universidad Industrial de Santander | Pattern Analysis and Applications | 2022 | Colombia | Google Scholar | 10.1007/s10044-022-01115-x | PD |
| 128 | Y. Zheng | [147] | Guangdong University of Technology | 2021 15th International Symposium on Medical Information and Communication Technology (ISMICT) | 2021 | China | Google Scholar | 10.1109/ISMICT51748.2021.9434916 | PD |
| 129 | Zeng Qingyi | [32] | Nankai University | 2022 12th International Conference on CYBER Technology in Automation, Control, and Intelligent Systems (CYBER) | 2022 | China | Google Scholar | 10.1109/CYBER55403.2022.9907308 | PD |
| 130 | Ahamed Musthaq | [148] | The Open University of Sri Lanka | arXiv preprint arXiv:2102.00628 | 2021 | Nugegoda | Google Scholar | 10.48550/arXiv.2102.00628 | PD |
| 131 | Balaji E | [149] | PSG College of Technology | Applied Soft Computing | 2020 | India | Google Scholar | 10.1016/j.asoc.2020.106494 | PD |
| 132 | Pratiher Sawon | [150] | Indian Institute of Technology | Automated Visual Inspection and Machine Vision II | 2017 | India | Google Scholar | 10.1117/12.2278894 | PD, HD and ALS |
| 133 | Li, Yan | [24] | University of Science and Technology of China | 2022 15th International Congress on Image and Signal Processing, BioMedical Engineering and Informatics (CISP-BMEI) | 2022 | China | Google Scholar | 10.1109/CISP-BMEI56279.2022.9980005 | PD |
| 134 | Wang, Qinghui | [151] | Longyan University | Cognitive Neurodynamics | 2022 | China | Google Scholar | 10.1007/s11571-022-09925-9 | PD |
| 135 | Torres, Andrés Mauricio Cárdenas | [152] | University of San Buenaventura | Ingenierías USBMed | 2023 | Colombia | Google Scholar | 10.1007/s11571-022-09925-9 | PD, ALS, HD |
| 136 | Goh, Choon-Hian | [153] | Universiti Tunku Abdul Rahman | 2022 IEEE-EMBS Conference on Biomedical Engineering and Sciences (IECBES) | 2022 | Malaysia | Google Scholar | 10.1109/IECBES54088.2022.10079640 | PD |
| 137 | Yi han | [154] | Zhejiang University | Sensors | 2023 | China | Google Scholar | 10.3390/s23042104 | PD |
| 138 | Kour, Navleen | [33] | Shri Mata Vaishno Devi University | Expert Systems | 2022 | India | Google Scholar | 10.1111/exsy.12955 | PD (and knee osteoarthritis (KOA)) |

| # | Author | Ref | Affiliation | Venue | Year | Country | Source | DOI | Disease |
|---|---|---|---|---|---|---|---|---|---|
| 139 | Zhou, Zeyang | [155] | University of Technology Sydney | Proceedings of the 2023 International Conference on Robotics, Control and Vision Engineering | 2023 | Australia | Google Scholar | 10.1145/3608143.3608154 | PD, ALS, HD |
| 140 | Beigi, Omid Mohamad | [156] | Brock University | 2022 IEEE Conference on Computational Intelligence in Bioinformatics and Computational Biology (CIBCB) | 2022 | Canada | Google Scholar | 10.1109/CIBCB55180.2022.9863050 | PD |
| 141 | Xu Chen | [157] | Hefei University of Technology | Applications in Health, Assistance, and Entertainment: 4th International Conference | 2018 | China | Google Scholar | 10.1007/978-3-319-92037-5_20 | PD |
| 142 | Yuyao Zhang | [158] | Intelligent Polymer Research Institute | 2013 International Conference on Digital Image Computing: Techniques and Applications (DICTA) | 2013 | Australia | Google Scholar | 10.1109/DICTA.2013.6691510 | PD |
| 143 | Rami Alkhatib | [159] | Université Jean-Monnet | Journal of Computer and Communications | 2015 | France | Google Scholar | 10.4236/jcc.2015.33003 | PD |
| 144 | Ömer Faruk Ertuğrul | [160] | Batman University | Expert Systems With Applications | 2016 | Turkey | Google Scholar | 10.1016/j.eswa.2016.03.018 | PD |
| 145 | Y. Nancy Jane | [4] | Anna University | Journal of Biomedical Informatics | 2016 | India | Google Scholar | 10.1016/j.jbi.2016.01.014 | PD |
| 146 | Yunfeng Wu | [161] | Xiamen University | Biomedical Signal Processing and Control | 2017 | China | Google Scholar | 10.1016/j.bspc.2016.08.022 | PD |
| 147 | Seyede Marziyeh Ghoreshi Beyrami | [162] | Sahand University of Technology | Measurement | 2020 | Iran | Google Scholar | 10.1016/j.measurement.2020.107579 | PD, HD, ALS |
| 148 | A.Athisakthi | [163] | Mother Teresa Women's University | International Journal Of Modern Engineering Research (IJMER) | 2018 | India | Google Scholar | 10.1109/WCCCT.2016.66 | PD, HD, ALS |
| 149 | Bashir Najafabadian | [164] | Islamic Azad University | 26th Iranian Conference on Electrical Engineering (ICEE2018) | 2017 | Iran | Google Scholar | 10.1109/ICEE.2018.8472503 | PD, HD, ALS |
| 150 | Satyabrata Aich | [37] | Inje University | International Journal of Engineering & Technology | 2018 | South Korea | Google Scholar | 10.14419/ijet.v7i3.29.18547 | PD |
| 151 | Anna Krajushkina | [165] | Tallinn University of Technology | 2018 IEEE International Conference on Systems, Man, and Cybernetics | 2018 | Estonia | Google Scholar | 10.1109/SMC.2018.00630 | PD |
| 152 | Tunç Aşuroğlu | [166] | Başkent University | Biocybernetics and Biomedical Engineering | 2018 | Turkey | Google Scholar | 10.1016/j.bbe.2018.06.002 | PD |
| 153 | Rana Hossam Elden | [167] | Helwan University | 2018 IEEE 4th Middle East Conference on Biomedical Engineering (MECBME | 2018 | Egypt | Google Scholar | 10.1109/MECBME.2018.8402417 | PD |
| 154 | Siddharth Arora | [16] | Aston University | 2014 IEEE International Conference on Acoustic, Speech and Signal Processing (ICASSP) | 2014 | UK | Google Scholar | 10.1109/ICASSP.2014.6854280 | PD |
| 155 | A Procházk | [168] | Institute of Chemical Technology in Prague | Digital Signal Processing | 2015 | Czech Republic | Google Scholar | 10.1016/j.dsp.2015.05.011 | PD |
| 156 | Milica Djurić-Jovičić | [169] | University of Belgrade | Neurological Research | 2017 | Serbia | Google Scholar | 10.1080/01616412.2017.1348690 | PD |
| 157 | Tuan D. Pham | [114] | Linkoping University | IEEE Transactions on Biomedical Engineering | 2017 | Sweden | Google Scholar | 10.1109/TBME.2017.2779884 | PD |
| 158 | Tripoliti, Evanthia E | [170] | University of Ioannina | Computer methods and programs in biomedicine | 2013 | Greece | Google Scholar | 10.1016/j.cmpb.2012.10.016 | PD |

| 159 | Mohammad Reza Daliri | [3] | Iran University of Science and Technology | Measurement | 2012 | Iran | Google Scholar | 10.1016/j.measurement.2012.04.013 | PD, HD, ALS |
| 160 | Zhao, Aite | [171] | Ocean University of China | Neurocomputing | 2018 | China | Google Scholar | 10.1016/j.neucom.2018.03.032 | PD |
| 161 | Butt, Abdul Haleem | [172] | University of Florence | 2020 42nd Annual International Conference of the IEEE Engineering in Medicine & Biology Society (EMBC) | 2020 | Italy | Google Scholar | 10.1109/EMBC44109.2020.9176051 | PD |
| 162 | Zhang, Hanrui | [173] | University of Michigan Medical School | Patterns | 2020 | USA | Google Scholar | 10.1016/j.patter.2020.100042 | PD |
| 163 | Yurdakul, Oğul Can | [174] | Middle East Technical University | Biomedical Signal Processing and Control | 2020 | Turkey | Google Scholar | 10.1016/j.bspc.2020.102070 | PD |
| 164 | M. Sneha Baby | [175] | Rajiv Gandhi Institute of Technology | 2017 International Conference on Circuits Power and Computing Technologies [ICCPCT] | 2022 | India | Google Scholar | 10.1109/ICCPCT.2017.8074230 | PD |

Table 5: List of Representative Excluded Papers and Reasons.

| No. | Title | Excluded Reason |
| --- | --- | --- |
| 1 | Accuracy of the Microsoft Kinect sensor for measuring movement in people with Parkinson's disease. | *Without using machine learning |
| 2 | Gait analysis comparing Parkinson's disease with healthy elderly subjects | *Without using machine learning |
| 3 | Arm swing magnitude and asymmetry during gait in the early stages of Parkinson's disease. | *Without using machine learning |
| 4 | A vision-based analysis system for gait recognition in patients with Parkinson's disease. | *others |
| 5 | Using Kinect to classify Parkinson's disease stages related to severity of gait impairment | *Without involving patients or controls |
| 6 | Postural control deficit during sit-to-walk in patients with Parkinson's disease and freezing of gait | *Without using machine learning |
| 7 | Decomposition of complex movements into primitives for Parkinson's disease assessment | *Without involving patients or controls |
| 8 | A validation study of freezing of gait (FoG) detection and machinelearning-based FoG prediction using estimated gait characteristics with a wearable accelerometer, | *Without involving patients or controls |
| 9 | Using transfer learning for classification of gait pathologies, | *Unmatched tasks |
| 10 | Classification of pathologies using a vision based feature extraction | *Without involving patients or controls |
| 11 | A vision-based system for movement analysis in medical applications: The example of Parkinson disease | *Without involving patients or controls |
| 12 | Principal component analysis of gait in Parkinson's disease: Relevance of gait velocity | *Without using machine learning |
| 13 | A vision-based regression model to evaluate Parkinsonian gait from monocular image sequences | *Without using machine learning |
| 14 | Quantification and recognition of Parkinsonian gait from monocular video imaging using kernel-based principal component analysis, | *Others |
| 15 | Video analysis of human gait and posture to determine neurological disorders | *Others |
| 16 | A novel single-sensor-based method for the detection of gaitcycle breakdown and freezing of gait in Parkinson's disease | *Unmatched tasks |
| 17 | Feature-Set-Engineering for Detecting Freezing of Gait in Parkinson's Disease using Deep Recurrent Neural Networks | *Unmatched tasks |
| 18 | A non-invasive medical device for parkinson's patients with episodes of freezing of gait | *Others |

| | | |
|---|---|---|
| 19 | Home monitoring of motor fluctuations in Parkinson's disease patients | *Without involving patients or controls |
| 20 | Gait analysis with wearables predicts conversion to parkinson disease. | *Without involving patients or controls |
| 21 | Automatic detection system for freezing of gait in Parkinson's Disease based on the clustering algorithm | *Unmatched tasks |
| 22 | Deep learning for freezing of gait detection in Parkinson's disease patients in their homes using a waistworn inertial measurement unit | *Unmatched tasks |
| 23 | Determining the optimal features in freezing of gait detection through a single waist accelerometer in home environments | *Unmatched tasks |
| 24 | Smart gait-aid glasses for Parkinson's disease patients. | *Unmatched tasks |
| 25 | Freezing of gait detection in parkinson's disease: a subject-independent detector using anomaly scores | *Unmatched tasks |
| 26 | A smartphone-based architecture to detect and quantify freezing of gait in Parkinson's disease. | *Unmatched tasks |
| 27 | Development and clinical validation of inertial sensor-based gait-clustering methods in Parkinson's disease. | *Without involving patients or controls |
| 28 | Gait anomaly detection of subjects with Parkinson's disease using a deep time series-based approach | *Others |
| 29 | The reliability of gait variability measures for individuals with Parkinson's disease and healthy older adults–the effect of gait speed. | *Without using machine learning |
| 30 | Effects of exercise on gait and motor imagery in people with Parkinson disease and freezing of gait | *Without using machine learning |
| 31 | The coefficient of friction in Parkinson's disease gait. | *Without using machine learning |
| 32 | Postural sensory correlates of freezing of gait in Parkinson's disease | *Unmatched tasks |
| 33 | Detection and quantification of freezing of gait and falls in Parkinson's disease patients using a body-worn sensor | *Unmatched tasks |
| 34 | Freezing of Gait detection in Parkinson's disease using accelerometer based smart clothes | *Unmatched tasks |
| 35 | Electromyography gait test for Parkinson disease recognition using artificial neural network classification in Indonesia. | *Without involving patients or controls |
| 36 | Home-based monitoring of falls using wearable sensors in Parkinson's disease. | *Without using machine learning |
| 37 | An in–laboratory validity and reliability tested system for quantifying hand–arm tremor in motions | *Without using machine learning |
| 38 | Fallers with Parkinson's disease exhibit restrictive trunk control during walking | *Others |
| 39 | Characterization of gait abnormalities in Parkinson's disease using a wireless inertial sensor system | *Without involving patients or controls |
| 40 | Diagnosing health problems from gait patterns of elderly | *Others |
| 41 | Biometric and mobile gait analysis for early diagnosis and therapy monitoring in Parkinson's disease | *Others |
| 42 | Accelerometry-based gait analysis and its application to parkinson's disease assessment-Part 2: A new measure for quantifying walking behavior. | *Without using machine learning |
| 43 | Automatic recognition of Parkinson's disease using surface electromyography during standardized gait tests | *Without using gait data |
| 44 | Detecting freezing-of-gait during unscripted and unconstrained activity | *Others |
| 45 | Objective detection of subtle freezing of gait episodes in Parkinson's disease | *Others |
| 46 | On assessing motor disorders in parkinson's disease | *Others |
| 47 | Smart Gait-Aid Glasses for Parkinson's Disease Patients | *Unmatched tasks |
| 48 | Gait and balance analysis for patients with Alzheimer's disease using an inertial-sensor-based wearable instrument | *Without using machine learning |
| 49 | A validated smartphone-based assessment of gait and gait variability in Parkinson's disease | *Without using machine learning |
| 50 | Comparative gait analysis in progressive supranuclear palsy and Parkinson's disease | *Without using machine learning |
| 51 | Ambulatory gait behavior in patients with dementia: a comparison with Parkinson's disease | *Without involving patients or controls |
| 52 | Contribution of a trunk accelerometer system to the characterization of gait in patients with mild-to-moderate Parkinson's disease | *Without using machine learning |

| | | |
|---|---|---|
| 53 | Disability and fatigue can be objectively measured in multiple sclerosis | *Without using machine learning |
| 54 | Free-living gait characteristics in ageing and Parkinson's disease: impact of environment and ambulatory bout length | *Without using machine learning |
| 55 | Insights into gait disorders: walking variability using phase plot analysis, Huntington's disease | *Without using machine learning |
| 56 | Quantitative evaluation of gait ataxia by accelerometers | *Without using machine learning |
| 57 | The parkinsonian gait spatiotemporal parameters quantified by a single inertial sensor before and after automated mechanical peripheral stimulation treatment | *Without using machine learning |
| 58 | Uncontrolled head oscillations in people with Parkinson's disease may reflect an inability to respond to perturbations while walking | *Without using machine learning |
| 59 | Validity and reliability of an IMU-based method to detect APAs prior to gait initiation | *Without using machine learning |
| 60 | Clinical assessment of standing and gait in ataxic | *Unmatched disease type |
| 61 | Validation of an Accelerometer to Quantify a Comprehensive Battery of Gait Characteristics in Healthy Older Adults and Parkinson's Disease: Toward Clinical and at Home Use | *Without using machine learning |
| 62 | Wearable Sensors in Huntington Disease A Pilot Study | *Without using machine learning |
| 63 | A comprehensive assessment of gait accelerometry signals in time, frequency and time-frequency domains | *Without using machine learning |
| 64 | A mobile Kalman-filter based solution for the real-time estimation of spatio-temporal gait parameters | *Without using machine learning |
| 65 | A novel approach to reducing number of sensing units for wearable gait analysis systems | *Without using machine learning |
| 66 | Accurate and reliable gait cycle detection in Parkinson's disease | *Without using machine learning |
| 67 | An inertial sensor based balance and gait analysis system | *Without using machine learning |
| 68 | Levodopa Is a Double-Edged Sword for Balance and Gait in People With Parkinson's Disease | *Unmatched tasks |
| 69 | Prolonged walking with a wearable system providing intelligent auditory input in people with Parkinson's disease | *Unmatched tasks |



\end{document}